\documentclass[journal]{IEEEtran}
\IEEEoverridecommandlockouts
\usepackage{cite}
\usepackage{amsmath,amssymb,amsfonts}
\usepackage{float}
\usepackage{graphicx}
\usepackage{textcomp}
\usepackage{amsmath, amssymb, amsthm}

\usepackage{xcolor}
\usepackage{algorithm}
\usepackage{algpseudocode}
\usepackage{graphicx}
\usepackage{subcaption}
\usepackage[justification=centering]{caption}
\usepackage{orcidlink}
\newcommand{\comment}[1]{}
\usepackage{dblfloatfix} 
\usepackage{threeparttable}
\usepackage{multirow} 

\newcommand{\orcidicon}[1]{\href{https://orcid.org/#1}{\textcolor{black}{\aiOrcid}}}
\ifCLASSINFOpdf
\else
\fi
\begin{document}

%
\title{Semantic-Aware Gaussian Process Calibration with Structured Layerwise Kernels for Deep Neural Networks}
%
%
%

\author{Kyung-Hwan Lee \orcidlink{0000-0002-4832-0420}
        and~Kyung-Tae Kim,~\IEEEmembership{Member, IEEE}

\thanks{Kyung-Hwan Lee is with the Next Generation Defence Multidisciplinary
Technology Research Center, Pohang University of Science and Technology,
Pohang 790-784, South Korea (e-mail: kyunghwanlee@postech.ac.kr) \\
\hspace*{1.5em}Kyung-Tae Kim is with the Department of Electrical Engineering, Pohang University of Science and Technology, Pohang 790-784, South Korea (e-mail: kkt@postech.ac.kr) \\
Corresponding Author: Kyung-Tae Kim
}}

%
%

\markboth{Journal of \LaTeX\ Class Files,~Vol.~13, No.~9, March~2025}%
{Shell \MakeLowercase{\textit{et al.}}: Bare Demo of IEEEtran.cls for Journals}
%



\maketitle

\begin{abstract}
Calibrating the confidence of neural network classifiers is essential for quantifying the reliability of their predictions during inference. However, conventional Gaussian Process (GP) calibration methods often fail to capture the internal hierarchical structure of deep neural networks, limiting both interpretability and effectiveness for assessing predictive reliability. We propose a Semantic-Aware Layer-wise Gaussian Process (SAL-GP) framework that mirrors the layered architecture of the target neural network. Instead of applying a single global GP correction, SAL-GP employs a multi-layer GP model, where each layer’s feature representation is mapped to a local calibration correction. These layerwise GPs are coupled through a structured multi-layer kernel, enabling joint marginalization across all layers. This design allows SAL-GP to capture both local semantic dependencies and global calibration coherence, while consistently propagating predictive uncertainty through the network. The resulting framework enhances interpretability aligned with the network architecture and enables principled evaluation of confidence consistency and uncertainty quantification in deep models.

\end{abstract}


%
\IEEEpeerreviewmaketitle

\section{Introduction}
\IEEEPARstart{R}{ecent} advancements in state-of-the-art deep learning classifiers have significantly improved target classification accuracy across various applications \cite{b1, b2, b3, b4}. This high accuracy has driven the widespread adoption of deep neural networks in real-world decision-making pipelines \cite{b5, b6, b7, b8, b9, b10, b11}. However, despite the common perception of their strong performance, modern deep neural classifiers can unexpectedly misclassify targets and produce unreliable confidence estimates during the testing phase \cite{b12, b13, b14}.

To effectively manage risks in scenarios where incorrect decisions by neural networks may have critical consequences, it is essential to ensure well-calibrated confidence scores, where the assigned probabilities accurately reflect the likelihood of correctness for a predicted class \cite{b15, b16}. By providing reliable confidence estimates for each class, calibration enhances model interpretability, enabling users to assess classifier reliability and make informed decisions based on classification results \cite{b17, b18, b19}.  For instance, confidence probabilities can be integrated into auxiliary probabilistic models to refine classification predictions, ultimately improving decision-making robustness. Therefore, confidence calibration in neural networks, which involves adjusting predicted probability distributions to better represent true correctness likelihoods, has been of great interest for ensuring that classification outputs accurately reflect model uncertainty \cite{b20, b21, b22, b23, b24, b25, b26, b27, b28, b29, b30}. 

Previous post-processing calibration methods include Bayesian Binning into Quantiles (BBQ), Isotonic Regression (IR), and Platt Scaling (PS), all of which were originally developed for binary classification but have since been extended to multi-class settings \cite{b13, b31, b32, b33, b34, b35, b36}. Among these, \cite{b14} demonstrated that Temperature Scaling (TS)—a multi-class extension of PS—is a particularly simple and effective calibration method for modern neural networks. However, these post-hoc methods do not generally provide explicit uncertainty estimates for their calibration performance.

Bayesian probabilistic frameworks that augment neural networks provide an alternative approach to calibration by enabling direct uncertainty quantification. For example, Bayesian Neural Networks (BNNs) output predictive distributions over class probabilities, inherently representing both mean predictions and associated uncertainties \cite{b37, b38, b39, b40, b41, b42, b43}. These Bayesian calibration methods are particularly advantageous when data is limited, as they regularize predictions and prevent overconfidence by integrating prior information and accounting for parameter uncertainty. With sufficiently large datasets, BNNs continue to provide well-calibrated confidence estimates, though their predictive means may converge to those of standard neural networks \cite{b44}. Despite these benefits, BNNs are often computationally intensive and may require more complex model architectures \cite{b21, b23, b25}. Furthermore, their predictive accuracy can sometimes lag behind that of conventional neural networks, especially when implemented with simpler architectural choices \cite{b67, b68, b69}.

In addition to providing uncertainty quantification without excessive computational or accuracy trade-offs, calibration techniques should also demonstrate robustness and reliability across a range of experimental conditions, including variations in datasets and neural network architectures. This requirement is particularly important due to domain shift between training and test datasets, which often arises from limited diversity in training data and inevitable differences in observational conditions between the training and deployment phases \cite{b14, b69, b70}. 

Another strategy for mitigating overconfidence in neural networks is label smoothing, a regularization technique that modifies the loss function during training by redistributing confidence probabilities across all classes \cite{b51, b52}. By softening the target label distribution, label smoothing discourages the model from assigning excessive probability to any single class, thereby reducing overconfident predictions. Unlike post-hoc calibration methods, which adjust model outputs at inference time, label smoothing is applied during training and does not explicitly address calibration under domain shift. As a result, it does not adapt confidence estimates to account for distributional discrepancies between training and test data, and may fail to produce well-calibrated probabilities in the presence of unseen or shifted test conditions \cite{b14, b52}.

To address these limitations, Gaussian process (GP)-based calibration methods have been introduced as alternatives to conventional post-hoc techniques, aiming to provide both predictive uncertainty estimates and reliable calibration for neural network outputs \cite{b27, b29, b30, b58, b59}. Most previous GP-based approaches adopt a single-layer mapping, which raises concerns regarding their effectiveness in mitigating overconfidence—especially when such overconfidence results from uncertainty propagated and accumulated throughout the network during training. Previous studies \cite{b29, b30} utilize softmax-based residual scoring during training and apply GP models with combined input–output kernels to estimate variance and calibration error at inference time. Although these approaches demonstrate effective uncertainty quantification capabilities, they have not been systematically compared to established post-hoc calibration methods with respect to calibration performance in deep neural networks. Moreover, the single-layer residual framework is inherently limited in its ability to address persistent overconfidence arising from the training data, as it imposes a functional GP prior over the entire network \cite{b69, b71, b72}. This global treatment effectively abstracts the deep model as a black box and fails to account for the layer-wise contributions to prediction confidence. This oversimplification is particularly problematic for undertrained models, where important layer-specific behaviors are not adequately represented by a global GP structure. To overcome these challenges, modeling each layer with a local GP provides a more expressive framework that captures the correlations among intermediate representations and their impact on the final predictive uncertainty \cite{b60, b61, b62, b63}.

One possible approach is to employ a stacked GP framework, in which the output of each layer serves as the input to the next, forming a sequential structure that enables highly flexible, non-linear transformations. A prominent example of this architecture is the Deep Gaussian Process (DGP) \cite{b63}. However, unlike standard Hierarchical Gaussian Processes (HGPs), DGPs require sophisticated posterior inference techniques—such as variational inference or Markov Chain Monte Carlo (MCMC)—to approximate the intractable posterior distributions, significantly increasing computational complexity and training time. In DGPs, while the initial input and the final output are deterministic, the intermediate representations at each layer are modeled as latent random variables whose distributions are marginalized during inference. This marginalization, rather than conditioning on explicit feature values, causes uncertainty introduced at each layer to accumulate as it propagates through the network. Consequently, the predictive uncertainty at the output layer can be substantially amplified, potentially exceeding the true uncertainty associated with model calibration. As a result, DGPs may be less suitable for calibration frameworks that require tightly controlled and interpretable uncertainty quantification \cite{b62, b63, b64, b65}.

The challenges inherent to DGP frameworks motivate the concept of a layerwise HGP, wherein multiple interdependent GPs are organized across different levels, each explicitly conditioned on layerwise latent variables \cite{b60, b61, b62}. In the context of calibrating deep neural network classifiers with multiple layers, an HGP-based approach can incorporate both a global GP and layer-specific GPs, with each layer’s process conditionally dependent on others within the hierarchy \cite{b60, b61, b62, b63, b27}.

However, it remains an open question whether hierarchical constructions of local GPs yield meaningful advantages for next-generation GP-based calibration. Structuring each layer’s GP as a local process beneath a higher-level GP could potentially limit the expressiveness of distinctive layer-specific responses, as each GP may be constrained by the hierarchical dependency structure \cite{b60, b61}.

To address this, one can introduce multi-layer kernel constructions, such as those based on the Intrinsic Coregionalization Model (ICM) kernel \cite{b53, b54}. In this framework, in addition to the feature or latent representation kernel, a specific index kernel is defined for each layer to capture both inter- and intra-layer dependencies. This is achieved by modeling the cross-covariance between input features and layer indices, often through multiplicative kernel structures across layers \cite{b54, b55}. Nevertheless, the computational cost and scalability issues—especially with large training datasets or high-dimensional features—can hinder the practical application of full multiplicative kernel forms for learning cross-layer dependencies \cite{b55, b56}.

A tractable approximation is to use only diagonal or low-rank representations of the layer correlation structure, adopting an additive kernel form that is significantly more computationally efficient, albeit less expressive in modeling inter-layer dependencies \cite{b53, b56}. In this way, each layer’s semantic latent representation can be incorporated using layerwise kernels for neural network calibration, leading to the Semantic-Aware Layerwise Gaussian Process (SAL-GP) framework with a multi-layer kernel (ML) approach.

In this paper, we present three key contributions of the SAL-GP framework for calibrating deep neural networks, particularly under data-scarce and domain-shifted conditions:

First, we propose the SAL-GP framework, which addresses the limitations of standard GP calibration methods while reducing the computational overhead and prediction uncertainty typically associated with BNN augmentations. To the best of our knowledge, no previous work has investigated structured layerwise calibration by explicitly modeling internal neural network layers as tasks within a layerwise GP framework. Our approach advances calibration metrics and enables principled quantification of calibration uncertainty in accordance with Bayesian methodology. By extending GP-based uncertainty estimation to a layerwise structure, this method directly addresses persistent overconfidence arising from internal network representations—an issue that conventional post-hoc techniques often fail to resolve, especially under distribution shift. The layerwise design integrates local, layer-specific latent variables to effectively capture global calibration behavior and inter-layer dependencies throughout the network. In addition, we incorporate localized kernel design by assigning stage-specific kernels indexed to each layer, enabling the model to reflect the distinct statistical properties and uncertainty propagation characteristics at every stage. This architecture supports more expressive and interpretable modeling of uncertainty evolution across the network.

Second, we develop and analyze several architectural variants of layerwise GPs with intra-layer connectivity. Unlike standard DGPs \cite{b63}, our architecture is explicitly aligned with the internal structure of the target neural network. The GP hierarchy is decomposed into layerwise submodules, where local GPs are organized under a shared global prior, closely reflecting the hierarchical organization of modern neural classifiers.

Finally, we evaluate the proposed architectures across a diverse set of data types and network backbones, each representing different calibration challenges. Our results identify which SAL-GP configurations yield the most reliable calibration performance and provide practical recommendations for robust, real-world deployment.

\section{Related Works}

\subsubsection{Performance of Calibrations of Conventional Methods}

In this section, we compare the performance of established calibration methods for modern neural networks to identify the most appropriate baseline for evaluating the SAL-GP framework. As a representative dataset, we consider Synthetic Aperture Radar (SAR) imagery from the Moving and Stationary Target Acquisition and Recognition (MSTAR) database \cite{b45}, which comprises 3,671 SAR images. The training set includes images with a depression angle of 15°, while the test set contains images with depression angles of 17° or 30°. Domain shift between training and testing phases in SAR imagery presents a significant challenge for model generalization and calibration, especially when the training dataset is limited or drawn from a distribution different from that of the test set \cite{b47, b48}. While the intrinsic domain shift in the MSTAR dataset is relatively modest, more pronounced distributional discrepancies can be simulated by reducing the size of the training set through random subsampling. Additional factors such as azimuthal angle, background variation, and scene clutter further complicate calibration, often resulting in inaccurate confidence estimates and increased misclassification during deployment \cite{b47}.

\begin{figure*}[ht]
    \centering
    \begin{minipage}{0.02\textwidth}\end{minipage}
    \begin{minipage}{0.18\textwidth}\centering \textbf{Raw} \end{minipage}
    \begin{minipage}{0.18\textwidth}\centering \textbf{Isotonic}\\\textbf{Regression}\end{minipage}
    \begin{minipage}{0.18\textwidth}\centering \textbf{BBQ} \end{minipage}
    \begin{minipage}{0.18\textwidth}\centering \textbf{Temp-Scaled}\\\textbf{(Init=0.5)}\end{minipage}
    \begin{minipage}{0.18\textwidth}\centering \textbf{Temp-Scaled}\\\textbf{(Init=2)}\end{minipage}
    \\[0.2em]
    \begin{minipage}{0.02\textwidth}\centering\rotatebox{90}{\textbf{100\%}}\end{minipage}
    \begin{minipage}{0.18\textwidth}\centering\includegraphics[width=\linewidth]{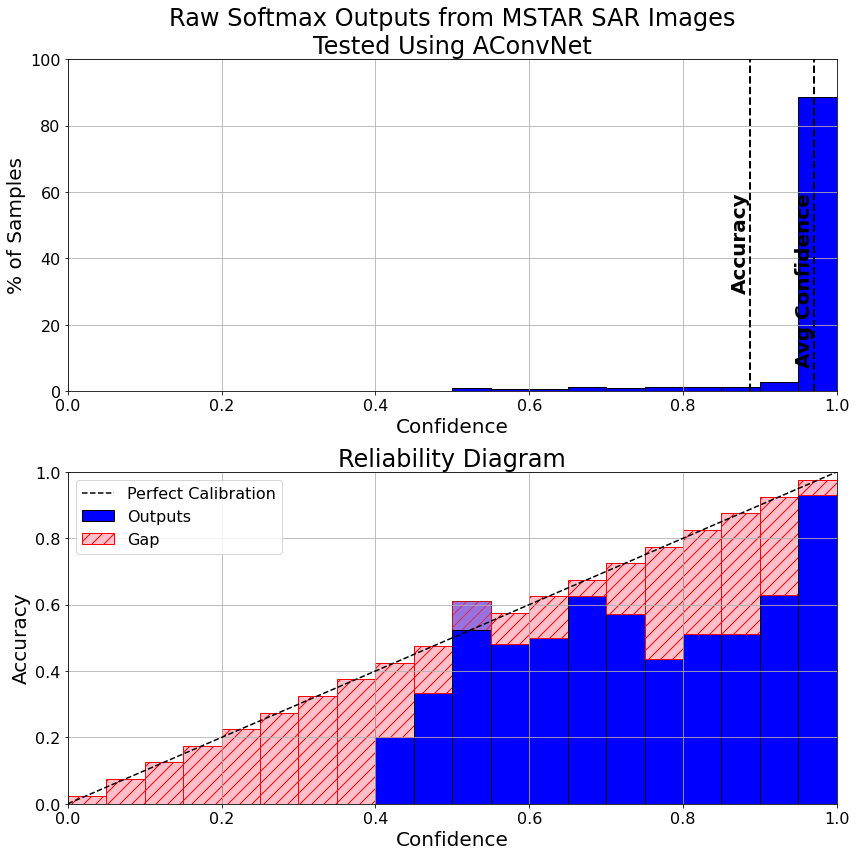}\end{minipage}
    \begin{minipage}{0.18\textwidth}\centering\includegraphics[width=\linewidth]{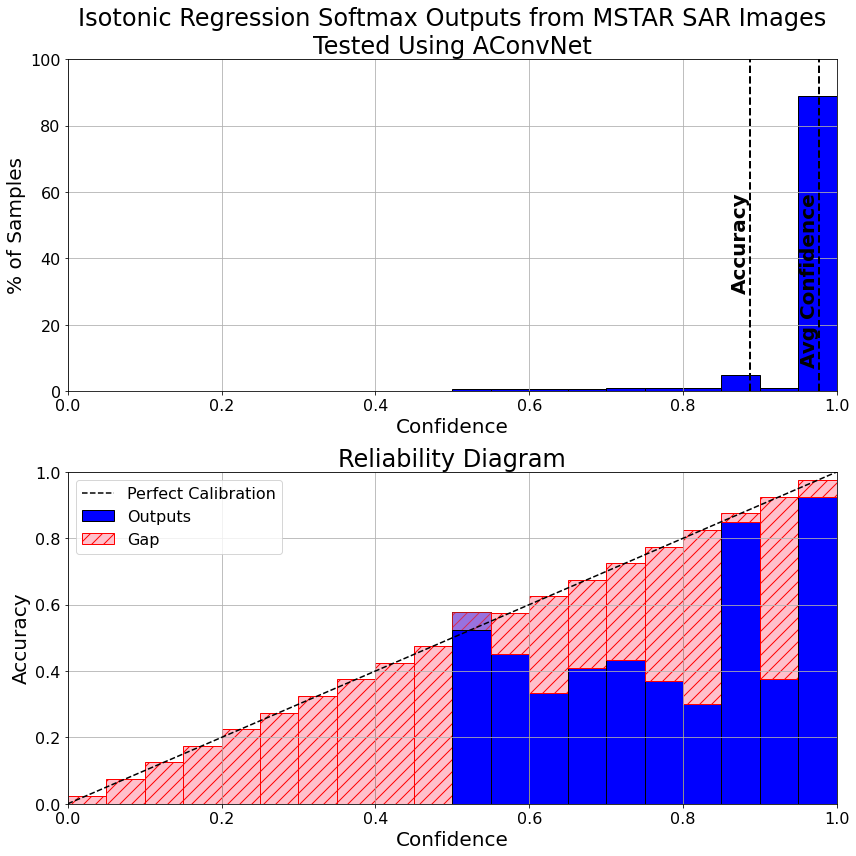}\end{minipage}
    \begin{minipage}{0.18\textwidth}\centering\includegraphics[width=\linewidth]{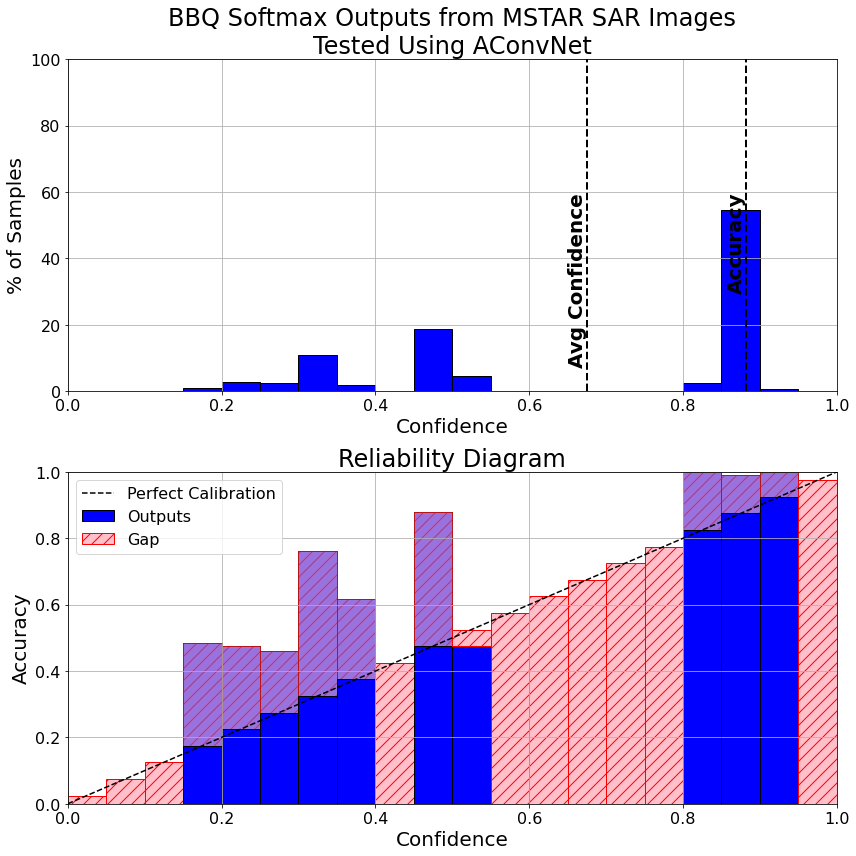}\end{minipage}
    \begin{minipage}{0.18\textwidth}\centering\includegraphics[width=\linewidth]{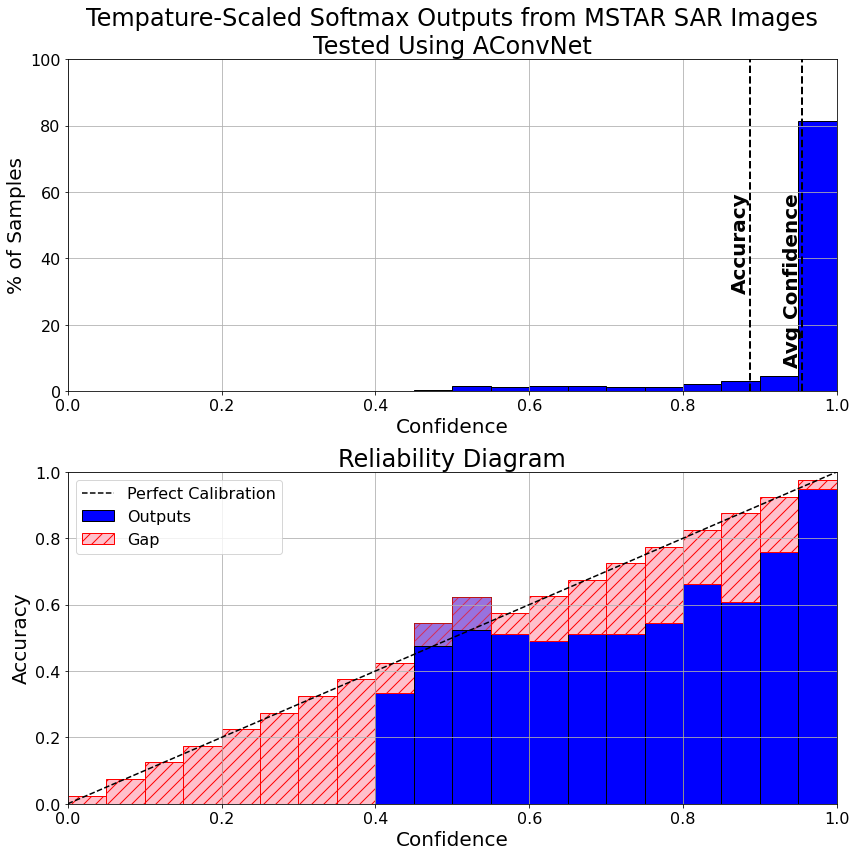}\end{minipage}
    \begin{minipage}{0.18\textwidth}\centering\includegraphics[width=\linewidth]{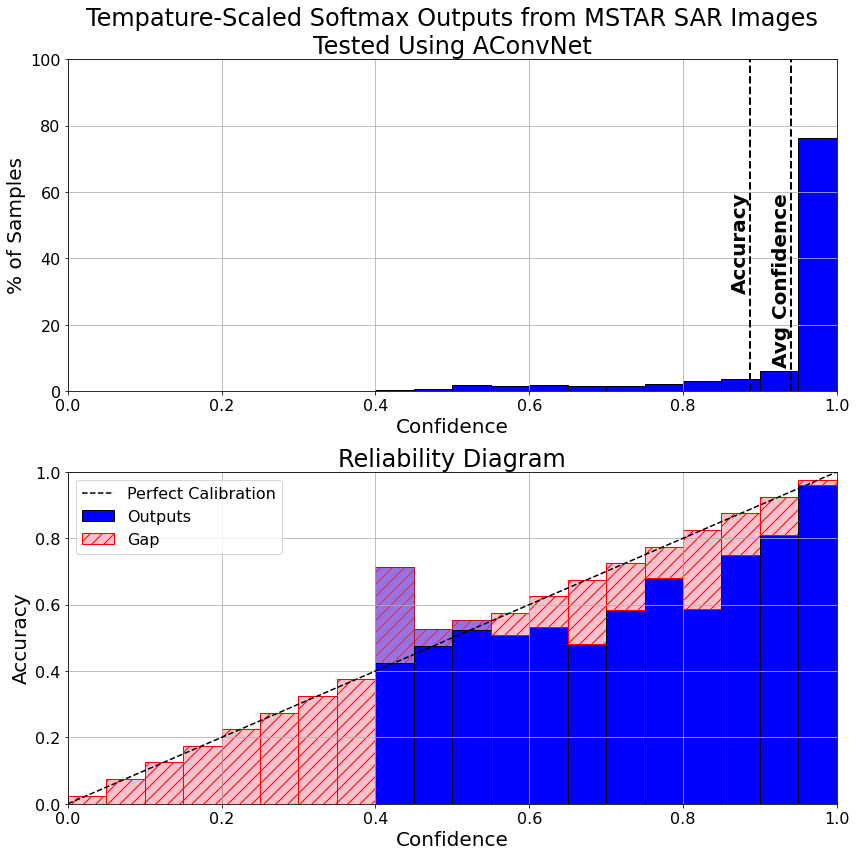}\end{minipage}
    \\[0.1em]
    \begin{minipage}{0.02\textwidth}\centering\rotatebox{90}{\textbf{50\%}}\end{minipage}
    \begin{minipage}{0.18\textwidth}\centering\includegraphics[width=\linewidth]{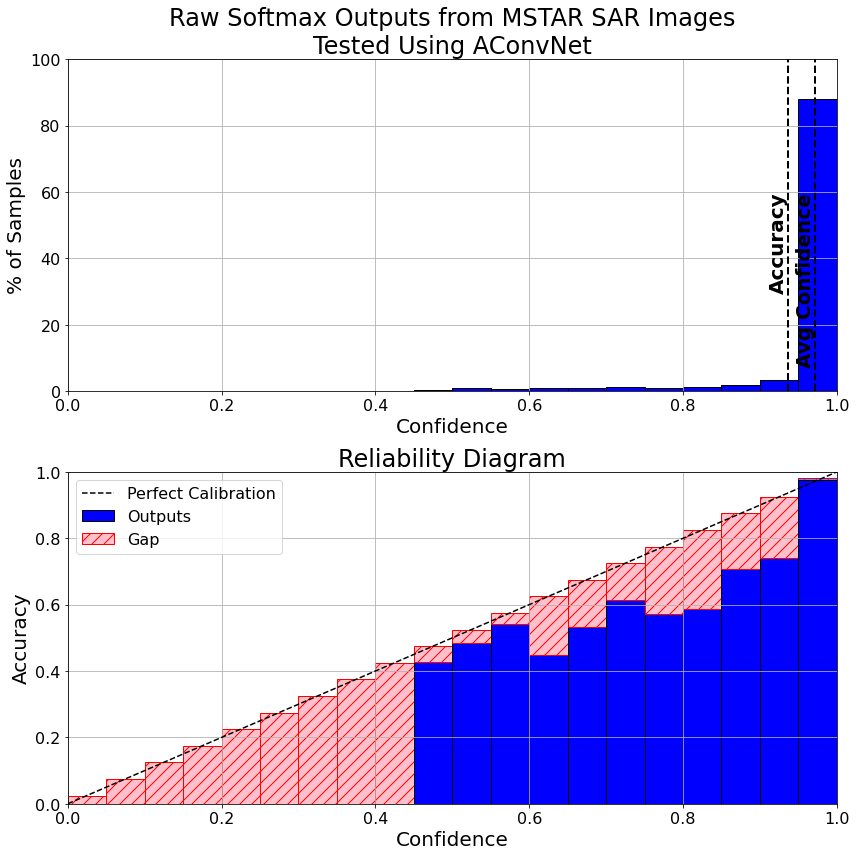}\end{minipage}
    \begin{minipage}{0.18\textwidth}\centering\includegraphics[width=\linewidth]{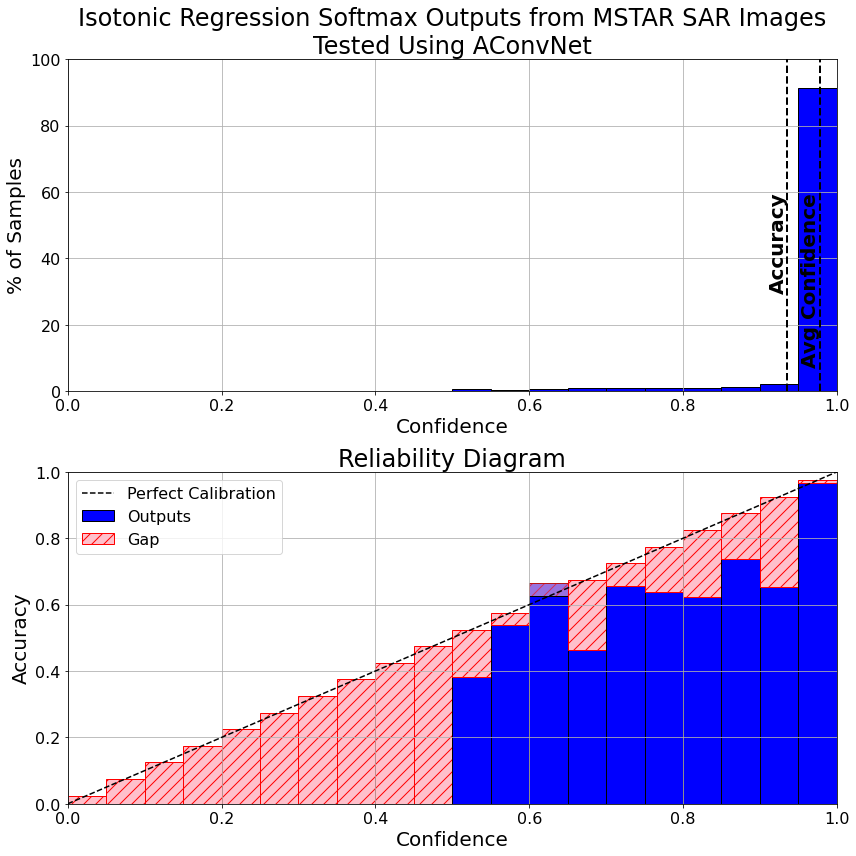}\end{minipage}
    \begin{minipage}{0.18\textwidth}\centering\includegraphics[width=\linewidth]{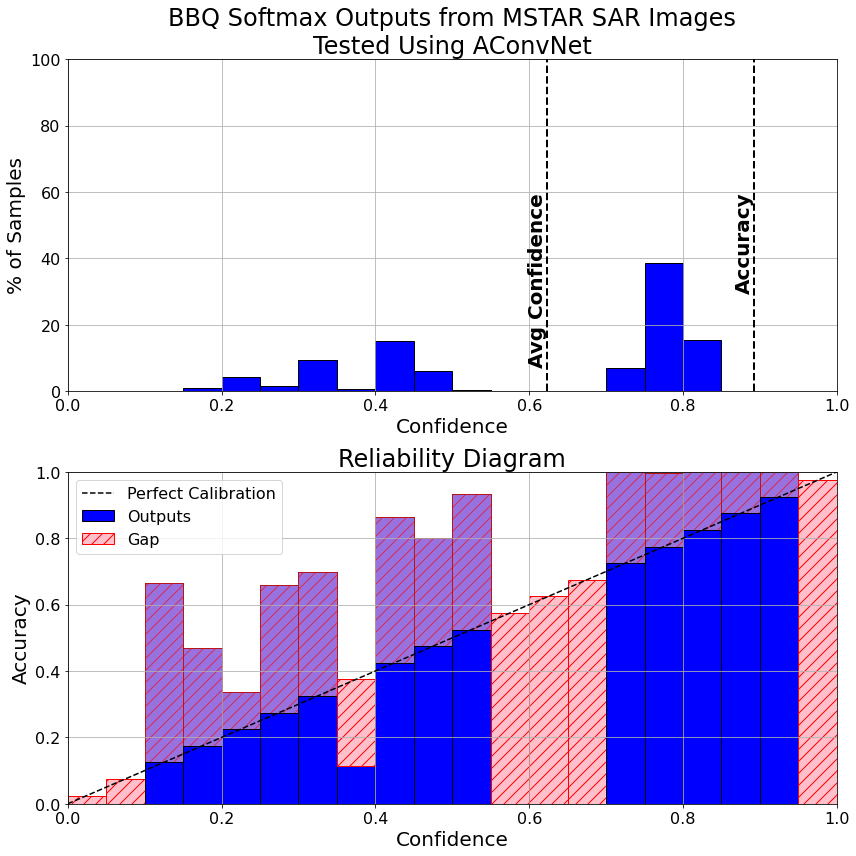}\end{minipage}
    \begin{minipage}{0.18\textwidth}\centering\includegraphics[width=\linewidth]{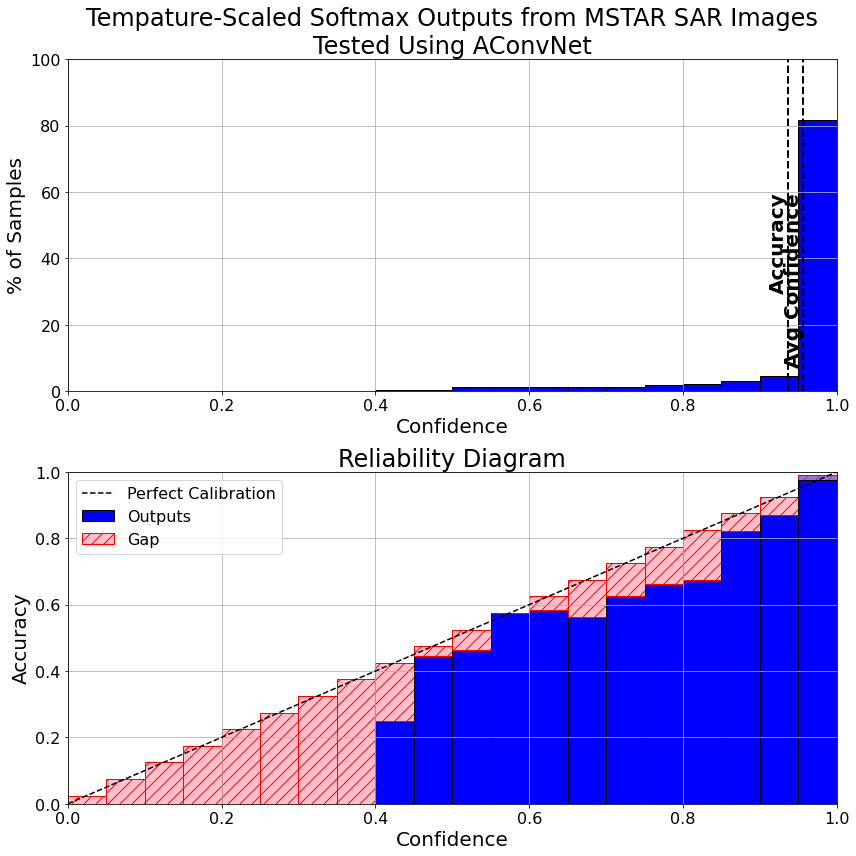}\end{minipage}
    \begin{minipage}{0.18\textwidth}\centering\includegraphics[width=\linewidth]{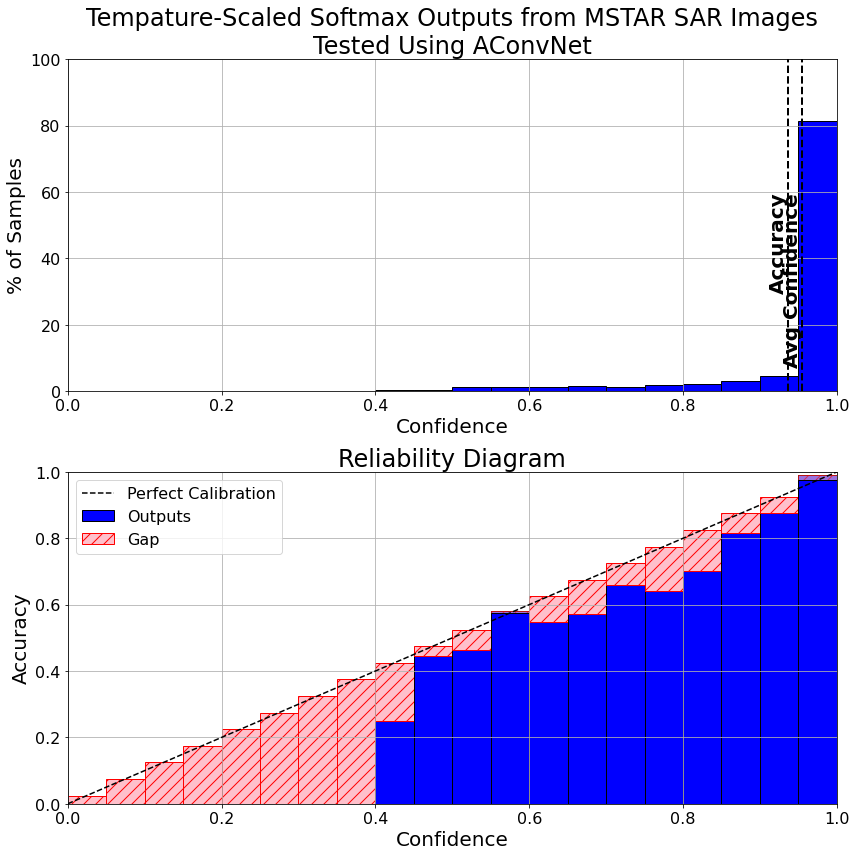}\end{minipage}
    \\[0.1em]
    \begin{minipage}{0.02\textwidth}\centering\rotatebox{90}{\textbf{25\%}}\end{minipage}
    \begin{minipage}{0.18\textwidth}\centering\includegraphics[width=\linewidth]{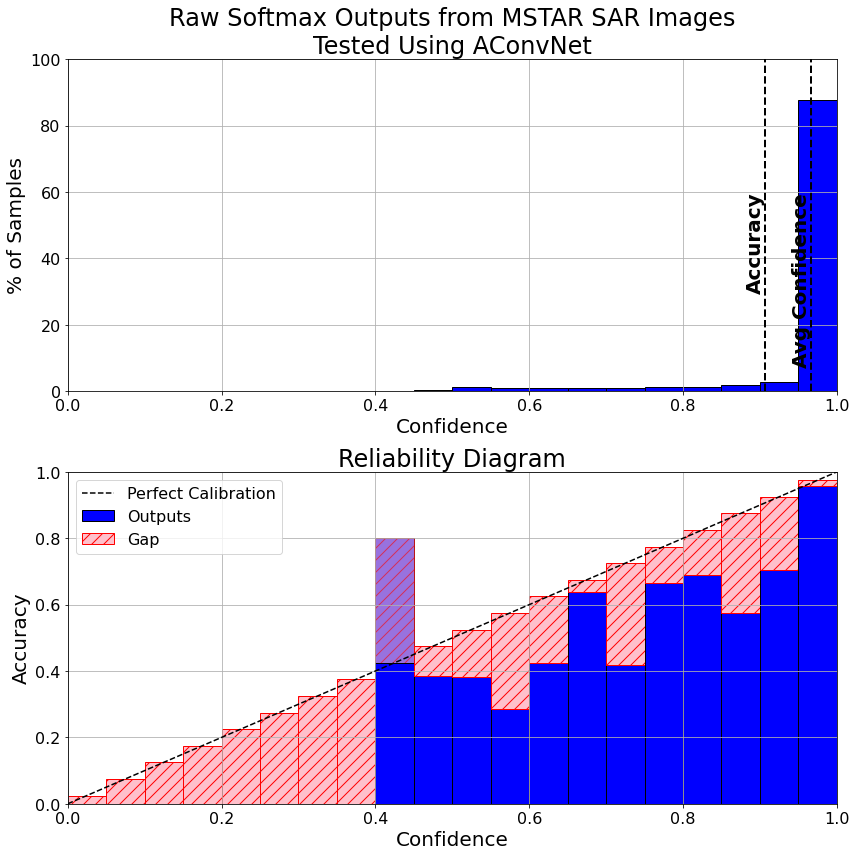}\end{minipage}
    \begin{minipage}{0.18\textwidth}\centering\includegraphics[width=\linewidth]{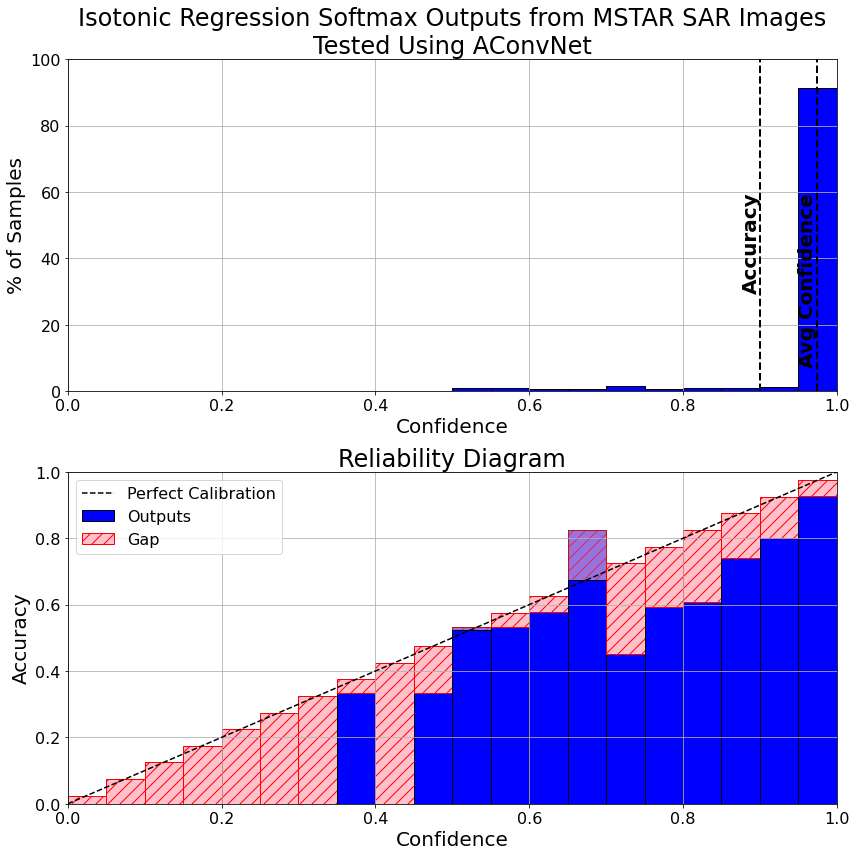}\end{minipage}
    \begin{minipage}{0.18\textwidth}\centering\includegraphics[width=\linewidth]{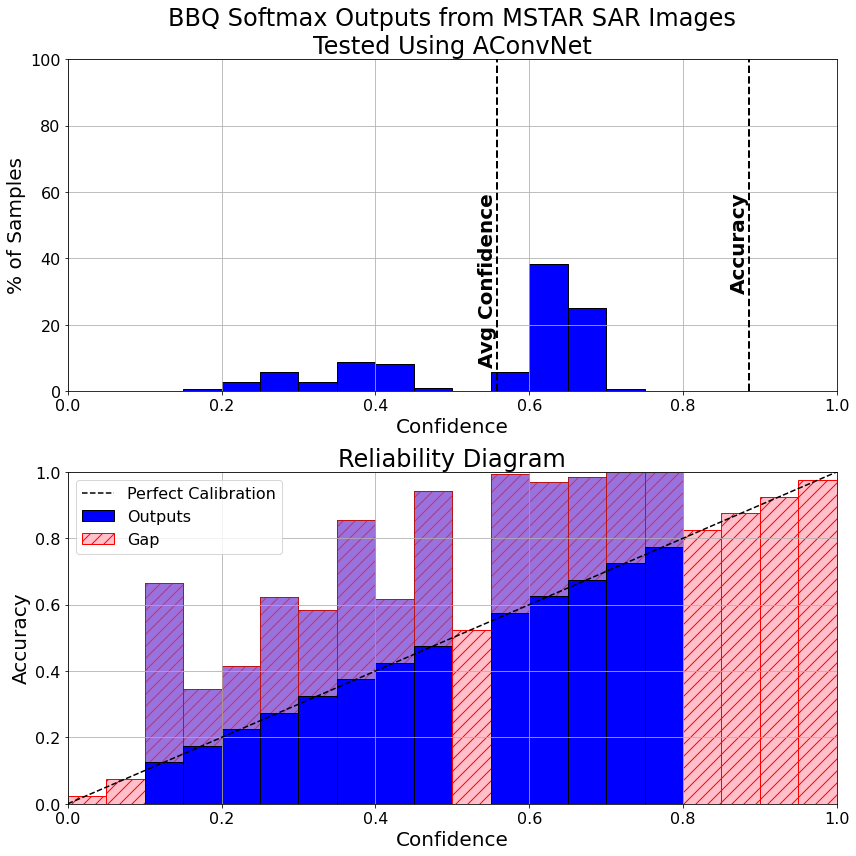}\end{minipage}
    \begin{minipage}{0.18\textwidth}\centering\includegraphics[width=\linewidth]{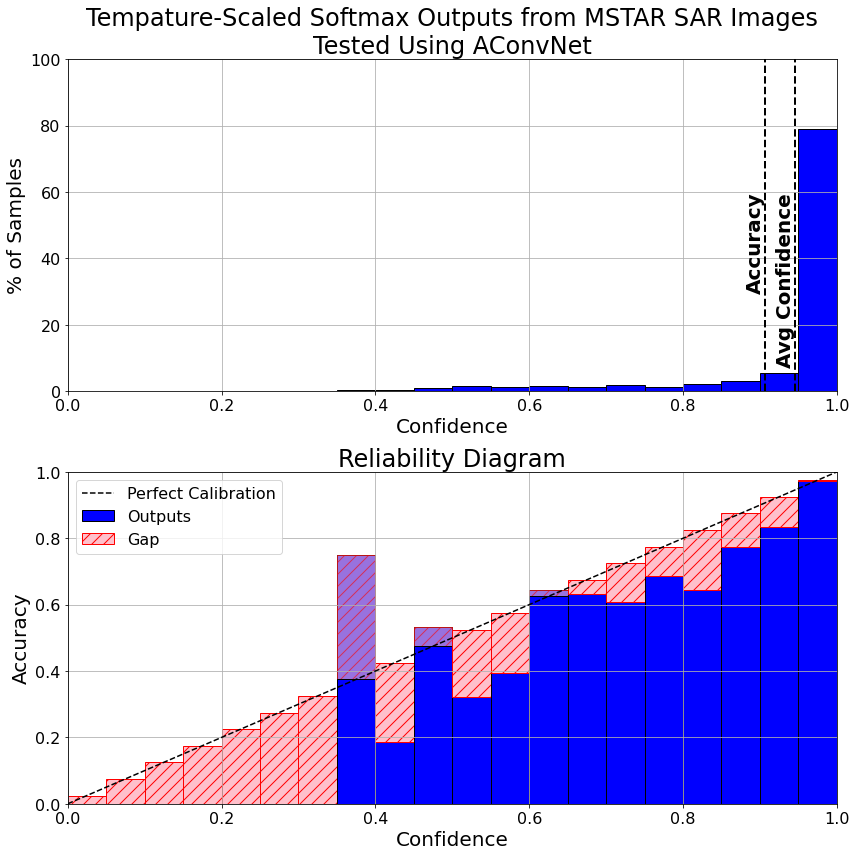}\end{minipage}
    \begin{minipage}{0.18\textwidth}\centering\includegraphics[width=\linewidth]{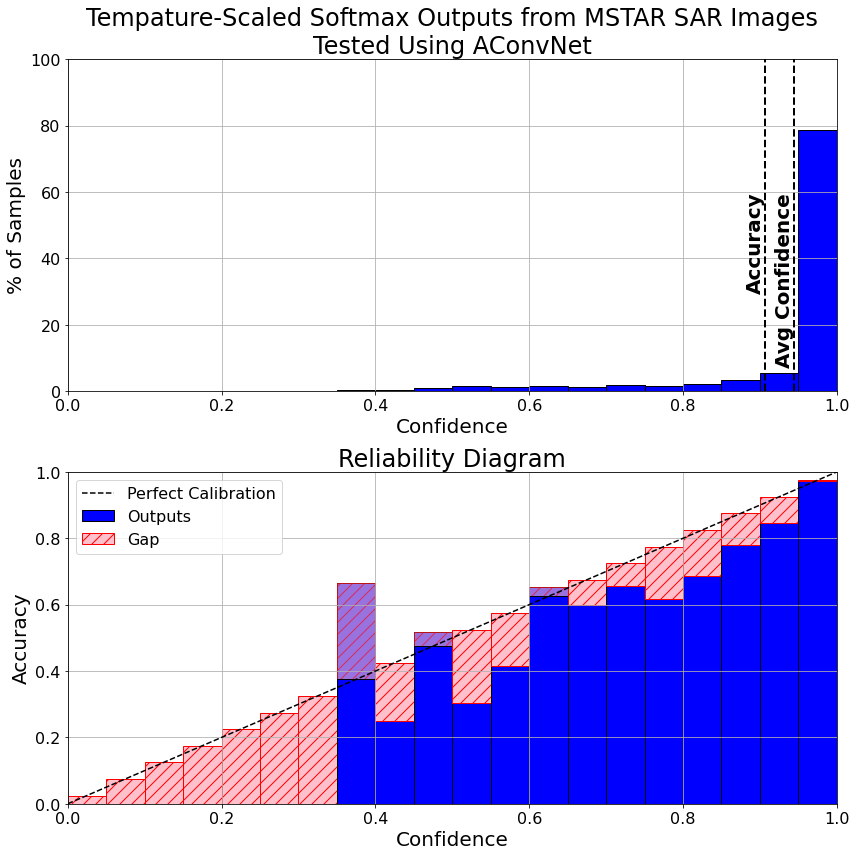}\end{minipage}
    \\[0.1em]
    \begin{minipage}{0.02\textwidth}\centering\rotatebox{90}{\textbf{10\%}}\end{minipage}
    \begin{minipage}{0.18\textwidth}\centering\includegraphics[width=\linewidth]{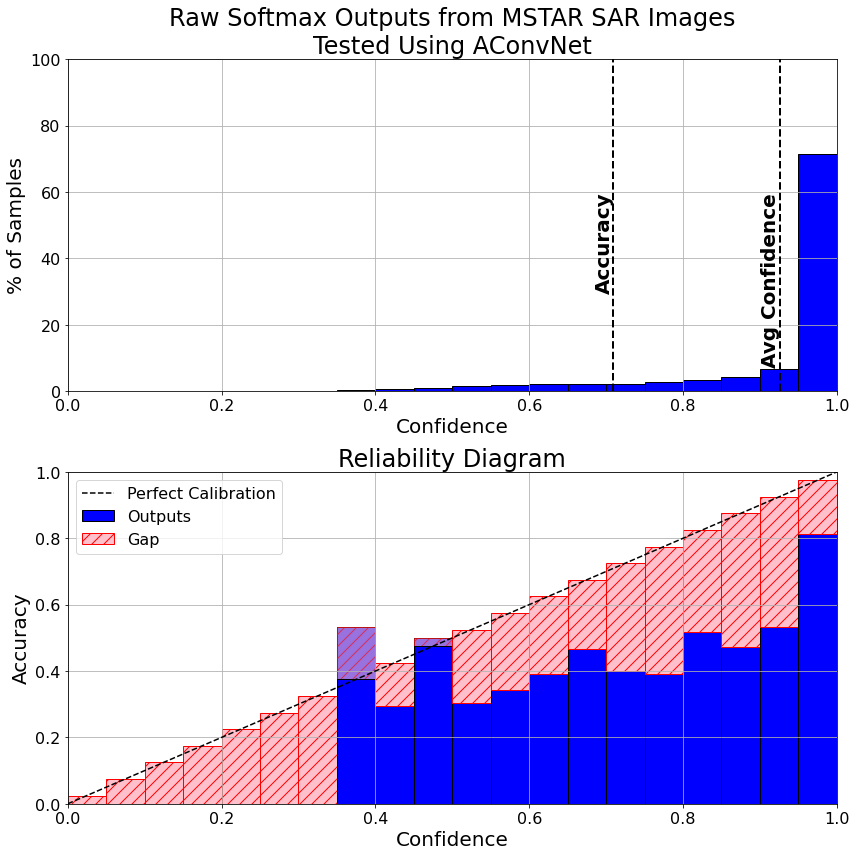}\end{minipage}
    \begin{minipage}{0.18\textwidth}\centering\includegraphics[width=\linewidth]{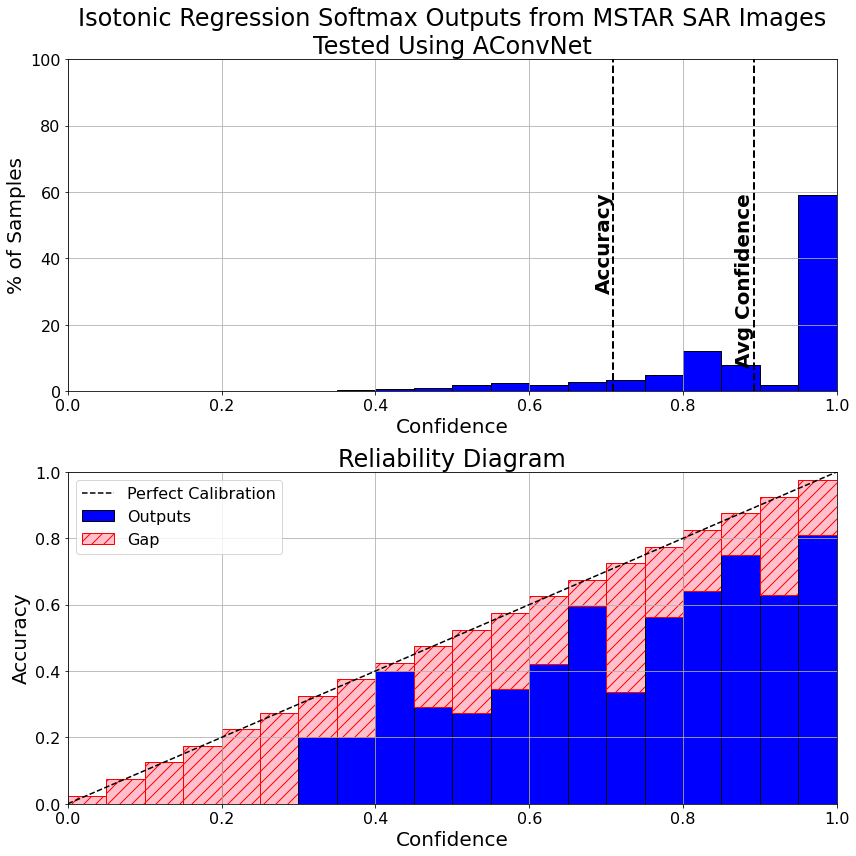}\end{minipage}
    \begin{minipage}{0.18\textwidth}\centering\includegraphics[width=\linewidth]{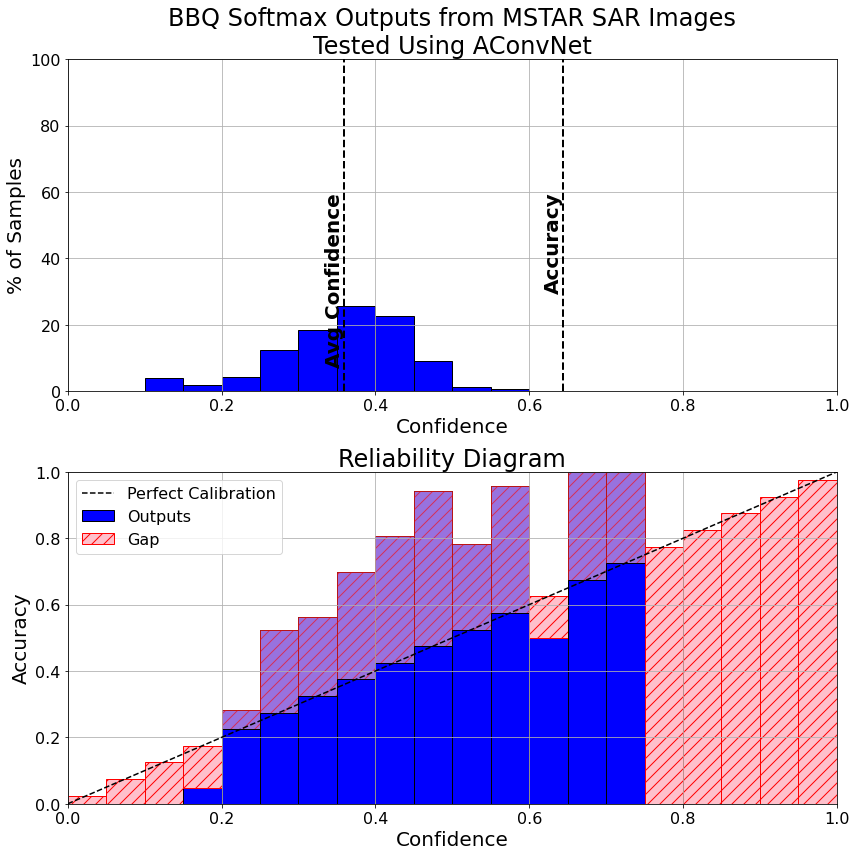}\end{minipage}
    \begin{minipage}{0.18\textwidth}\centering\includegraphics[width=\linewidth]{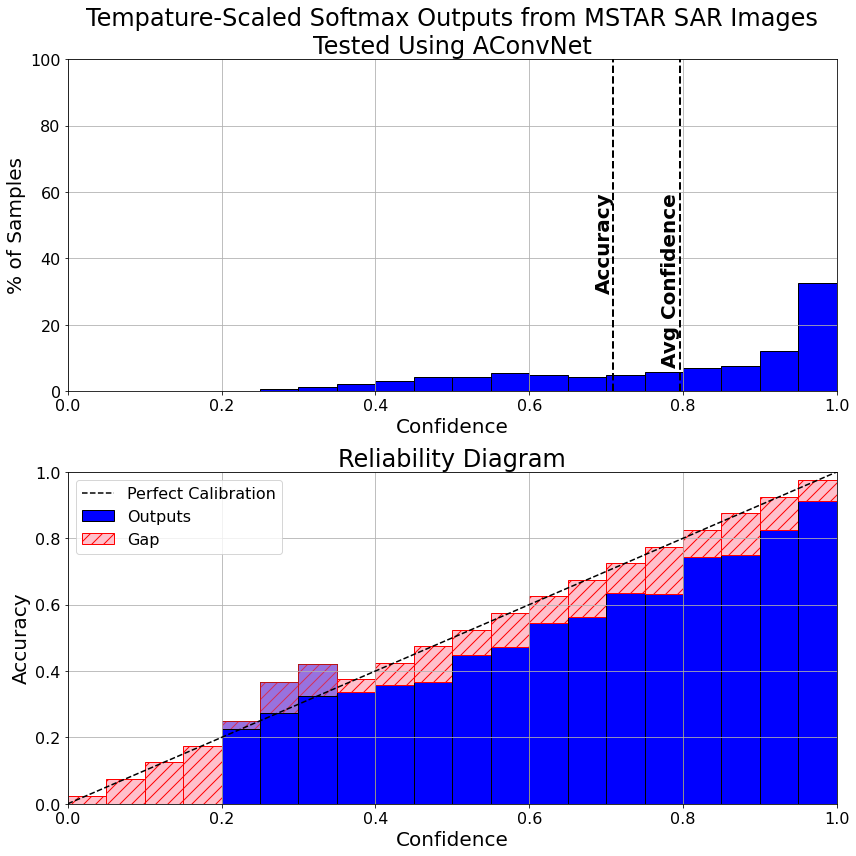}\end{minipage}
    \begin{minipage}{0.18\textwidth}\centering\includegraphics[width=\linewidth]{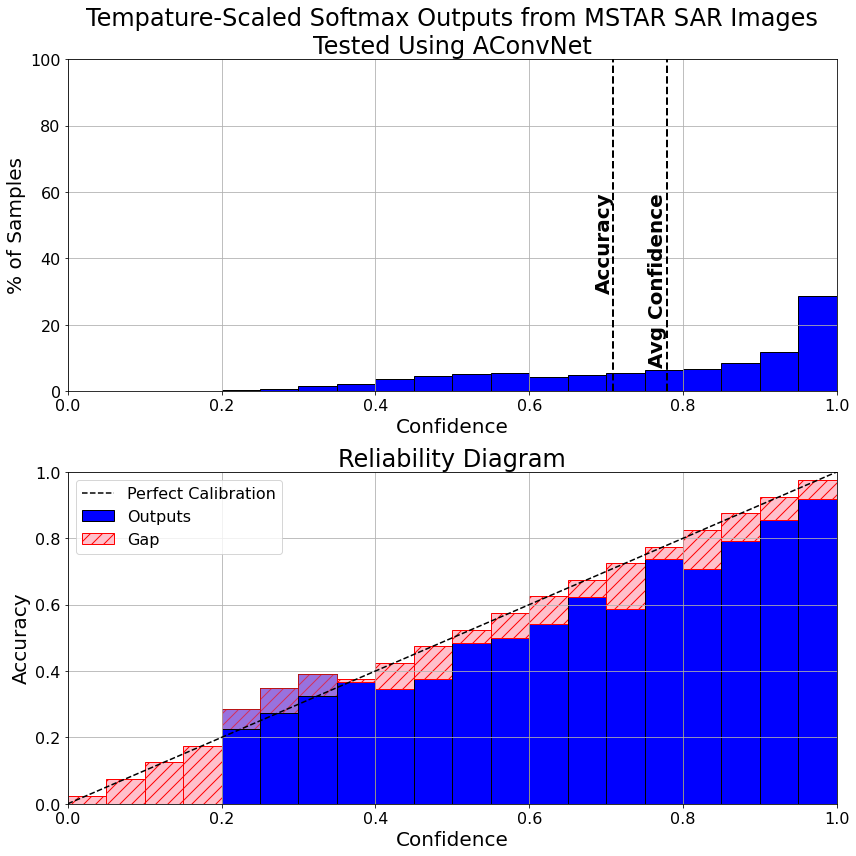}\end{minipage}

    \caption{Comparative performance of various calibration methods on MSTAR DB.
    \textbf{Column titles:} Raw, Isotonic Regression, BBQ, Temperature-Scaled (Init=0.5), Temperature-Scaled (Init=2).
    \textbf{Row titles:} 100\%, 50\%, 25\%, 10\% training data.}
    \label{fig:calibration_mstar_grid}
\end{figure*}

We compare the performance of several post-hoc calibration methods, including Isotonic Regression, BBQ\cite{b33}, and TS (with two initial temperature values)\cite{b14}, across the MSTAR dataset and different training sample sizes \cite{b45}. Fig.~\ref{fig:calibration_mstar_grid} presents reliability diagrams and calibration metrics for the test set, evaluated at training sample fractions from 10\% to 100\%.

Reliability diagrams visually assess model calibration by plotting expected accuracy against predicted confidence \cite{b12, b13, b14}. A perfectly calibrated model follows the diagonal, indicating that predicted confidence matches actual accuracy. Deviations from this line reveal miscalibration. Note that reliability diagrams also show the proportion of samples in each bin, indicating the distribution of calibration quality.

In each figure, the top panel includes dotted lines representing the network’s mean accuracy (ground truth) and mean predicted confidence. A visible gap between these lines signifies miscalibration. In the bottom panel, blue bars indicate observed accuracy within each confidence bin, while red-shaded areas denote the discrepancy between predicted confidence and actual accuracy, highlighting overconfidence.

Among the calibration methods evaluated, Isotonic Regression and BBQ provide limited or even negative improvement over uncalibrated neural network confidence outputs. TS consistently delivers the most effective calibration across different training set sizes and initial values, in line with prior findings~\cite{b14}. Notably, BBQ calibration \cite{b33} remains suboptimal for modern neural network classifiers, especially under limited data conditions. A frequently cited drawback of BBQ is its sensitivity to hyperparameters—specifically, the Beta prior parameters ($\alpha_{prior}$ and $\beta_{prior}$) that directly affect the degree of smoothing in sparsely populated bins. To address these issues, previous studies\cite{b13, b33} have proposed several strategies. One common approach is hyperparameter tuning via cross-validation, which involves performing grid searches over plausible $\alpha_{prior}$ and $\beta_{prior}$ values using a separate calibration set and selecting parameters that minimize calibration error metrics such as Expected Calibration Error (ECE), Maximum Calibration Error (MCE), or Negative Log-Likelihood (NLL). Another approach uses empirical Bayes estimation, in which prior parameters are estimated from validation data by maximizing the marginal likelihood, enabling data-driven adaptation of prior strength. Alternatively, employing weakly informative priors—such as uniform priors with $\alpha_{prior} = \beta_{prior} = 1$—can reduce sensitivity, as these exert minimal influence given sufficient data per bin. When bins contain very few samples, further adjustments, including increasing the number of bins or applying additional smoothing or regularization, may be necessary to avoid undue influence of the priors.

\begin{figure*}[ht]
    \centering
    \begin{subfigure}[t]{0.48\textwidth}
        \centering
        \includegraphics[width=\linewidth]{BBQ_10percent_data_AConvNet_Test.png}
        \caption{BBQ calibration results using $\alpha=1, \beta=1$: ECE=$0.29950$, MCE=$0.47306$, NLL=$2.97976$.}
        \label{fig:BBQ_Alpha1}
    \end{subfigure}
    \hfill
    \begin{subfigure}[t]{0.48\textwidth}
        \centering
        \includegraphics[width=\linewidth]{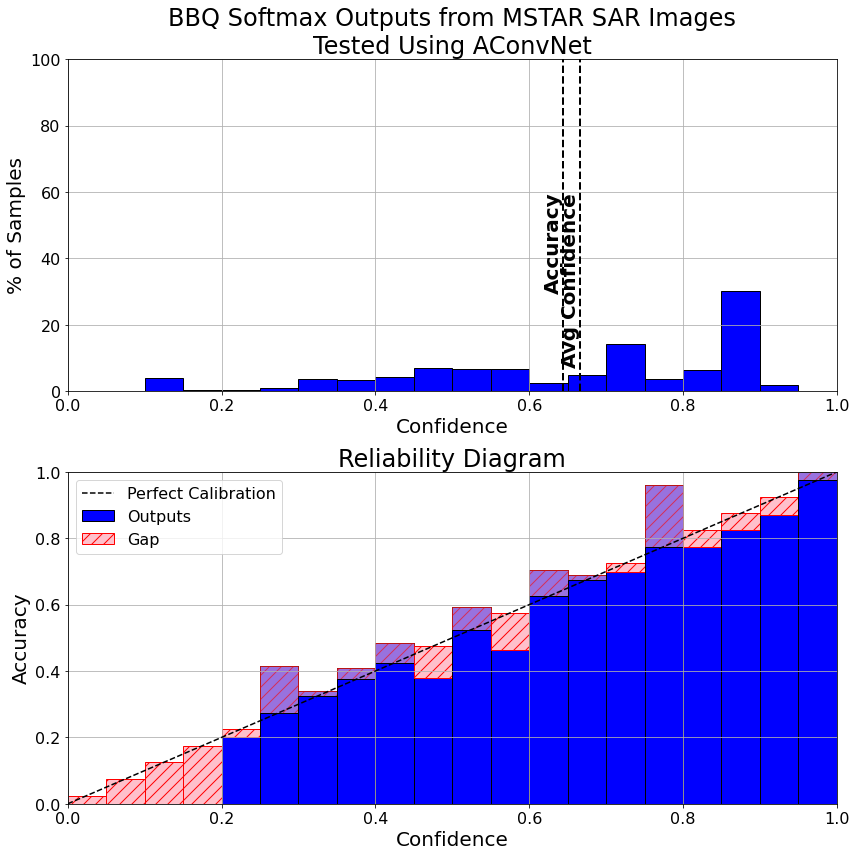}
        \caption{BBQ calibration results using $\alpha=0.1, \beta=1$: ECE=$0.06246$, MCE=$0.17871$, NLL=$2.78592$.}
        \label{fig:BBQ_Alpha0p1}
    \end{subfigure}
    \caption{Effect of varying BBQ prior hyperparameters on calibration performance for a $10\%$ subsample from the MSTAR dataset. Reducing $\alpha$ from $1$ to $0.1$ markedly improves ECE and MCE, as shown above.}
    \label{fig:BBQ_Calibration_Alternative}
\end{figure*}

For a subsample of $10\%$ from the MSTAR dataset, using $\alpha=1$ and $\beta=1$ yields an ECE of $0.29950$, MCE of $0.47306$, and NLL of $2.97976$. Reducing the prior to $\alpha=0.1$ significantly improves calibration, achieving an ECE of $0.06246$, MCE of $0.17871$, and NLL of $2.78592$, as shown in Fig.~\ref{fig:BBQ_Calibration_Alternative}. However, despite these modifications, BBQ's reliance on validation-driven updates and prior assumptions continues to produce empirical inconsistencies in calibration performance under varying experimental conditions. Therefore, for baseline comparisons in this study, we adopt TS as the most robust and reliable of the conventional calibration methods \cite{b14}.

\section{Method}

\subsection{Gaussian Process Regression}

We review a brief introduction to GP regression along with key notations; for a comprehensive discussion, refer to \cite{b58, b59}.

A GP is a joint multivariate Gaussian distribution over functions, expressed as 
$f(x) \sim \mathcal{GP}(\mu(x), k(x, x'))$, where $\mu(x)$ and $k(x, x')$ are the mean and covariance functions, respectively, which fully define the GP. The input space is the real-valued feature space $\Omega$, with $w \in \Omega$ and $x, x' \in \mathbb{X}$, where $x$ and $x'$ correspond to training and test inputs in a set of training inputs $X$ and a set of test inputs $X'$.

In our Bayesian framework for GP regression, we begin with a prior distribution over functions and update it in light of observed data to obtain a posterior distribution. Leveraging standard properties of Gaussian distributions, regression is reduced to straightforward linear algebra operations.

Throughout this work, we alternatively use the Radial Basis Function (RBF) and Matérn covariance functions, defined as follows:

For the RBF covariance function,
\begin{equation} k(x, x') = \alpha \exp{(-\gamma (x - x')^2)}, \end{equation}
where $\alpha$ scales the functional variance, and $\gamma$ controls the smoothness of the GP.

For the Matérn covariance function,
\begin{equation} k_{\nu}(r) = \sigma^2 \frac{2^{1-\nu}}{\Gamma(\nu)} \left( \frac{\sqrt{2\nu} r}{\ell} \right)^{\nu} K_{\nu} \left( \frac{\sqrt{2\nu} r}{\ell} \right), \end{equation}
where $r = | x - x' |$ is the Euclidean distance between input points $x$ and $x'$, $\sigma^2$ is the variance parameter, $\ell$ is the length-scale, and $\nu$ controls smoothness. Here, $K_{\nu}(\cdot)$ denotes the modified Bessel function of the second kind and $\Gamma(\nu)$ is the Gamma function.

The choice of covariance function reflects prior assumptions about the smoothness or potential discontinuities in the underlying function \cite{b59}. 

Regression is performed using the marginal and conditional properties of multivariate Gaussian distributions. Then, for training GP model, the marginal likelihood distribution over observed data y at input points X is given by 

\begin{equation}
p(y \mid X) = \mathcal{N}(\mu(X), K(X, X) + \sigma^2 I)
\end{equation}

Given function observations $f$ at input points $x$, we aim to predict function values at new input points $x'$, denoted $f'$. The joint probability distribution of $f$ and $f'$ is:

\begin{equation}
    p
    \begin{bmatrix} f \\ f' \end{bmatrix}
    =
    N
    \left(
    \begin{bmatrix} f \\ f' \end{bmatrix}
    \Bigg|
    0,
    \begin{bmatrix} K_{x,x} & K_{x,x'} \\ K_{x',x} & K_{x',x'} \end{bmatrix}
    \right)
\end{equation}
For simplicity, a zero-mean function is assumed without loss of generality.
Using the conditional properties of multivariate Gaussians, we derive the predictive distribution:

\begin{equation}
    p(f' | f) = N
    \left(
    f' | K_{x',x} K_{x,x}^{-1} f, K_{x',x'} - K_{x',x} K_{x,x}^{-1} K_{x,x'}
    \right)
\end{equation}

Observations include a noise-corrupted measurement vector $y$, modeled as:

\begin{equation}
    p(y | f) = N(y | f, \beta I)
\end{equation}

where $\beta$ is the noise variance and $I$ is the identity matrix. The final marginal likelihood for optimizing hyperparameters is:

\begin{align}
    \log p(y \mid \theta) =\ 
    &- \frac{D}{2} \log 2\pi \notag \\
    &- \frac{1}{2} \log \left| K_{x,x} + \beta I \right| \notag \\
    &- \frac{1}{2} y^\top \left(K_{x,x} + \beta I\right)^{-1} y
\end{align}

\subsection{Feature Map Structure and Processing}

Assume that for a given layer $l$ of a Convolutional Neural Network (CNN), the output feature map for a batch of images is represented as a four-dimensional tensor:
\begin{align}
    \mathbf{F}^{(l)} \in \mathbb{R}^{N \times C \times H \times W}
\end{align}
where $N$ denotes the number of samples or images in the batch, $C$ is the number of channels or feature maps produced by the convolutional layer, and $H$ and $W$ correspond to the height and width of each feature map, respectively.

For an individual sample $i$, the feature map is a three-dimensional array:
\begin{align}
    \mathbf{F}_i^{(l)} \in \mathbb{R}^{C \times H \times W}
\end{align}

Pooling operations are commonly applied to aggregate information across the channel dimension at each spatial location $(h, w)$.

For max pooling across the channel axis, the operation for each sample $i$ at spatial location $(h, w)$ is defined as
\begin{align}
    \mathbf{M}_i^{(l)}[h, w] = \max_{c} \mathbf{F}_i^{(l)}[c, h, w],
\end{align}
where the resulting pooled feature map has shape $\mathbf{M}_i^{(l)} \in \mathbb{R}^{H \times W}$.

Similarly, average pooling across the channel axis is defined for each sample $i$ and spatial location $(h, w)$ as
\begin{align}
    \mathbf{A}_i^{(l)}[h, w] = \frac{1}{C} \sum_{c=1}^C \mathbf{F}_i^{(l)}[c, h, w],
\end{align}
which computes the mean value across all channels at each spatial location. The resulting pooled feature map has shape $\mathbf{A}_i^{(l)} \in \mathbb{R}^{H \times W}$.

If the latent representation does not include a channel dimension—for example, when using feature representations obtained from a Recurrent Neural Network (RNN) \cite{b73, b74, b75}, the pooling step is omitted, and the layer's feature map is used directly.

\subsubsection{Gaussian Process Regression on Softmax Residuals}

Let the training dataset be
\begin{equation}
\mathcal{D} = \{(\mathbf{z}_i, s_i, r_i)\}_{i=1}^N
\end{equation}
Here, $\mathbf{z}_i \in \mathbb{R}^{d}$ denotes the flattened latent feature vector for the $i$-th sample, constructed by applying max or average pooling to the feature map at a fixed layer $l$—that is, $\mathbf{z}_i = \mathrm{vec}(A_i^{(l)})$ or $\mathrm{vec}(M_i^{(l)})$, where $A_i^{(l)}$ and $M_i^{(l)}$ denote the average- and max-pooled feature maps, respectively.

The scalar $s_i \in [0, 1]$ denotes the model confidence, defined as the maximum softmax probability for the $i$-th input. The target $r_i = c_i - s_i$ is the softmax residual, where $c_i$ is a correctness indicator equal to $1$ if the prediction is correct ($y_i = \hat{y}_i$) and $0$ otherwise \cite{b29, b30}.

Consider training a GP model to predict the softmax residual $r_i = c_i - s_i$ for each sample, where $\mathbf{z}_i = \mathrm{vec}(A_i^{(l)})$ or $\mathrm{vec}(M_i^{(l)})$ denotes the flattened latent feature vector from a fixed convolutional layer $l$, $s_i$ is the maximum softmax probability (network confidence), and $c_i$ is the correctness indicator.

A standard approach is to use the feature representation $\mathbf{z}_i \in \mathbb{R}^d$ as the input to the GP, which learns a mapping $f_{\mathrm{feat}}: \mathbf{z}_i \mapsto r_i$. The GP prior is specified as
\begin{equation}
    f_{\mathrm{feat}}(\mathbf{z}) \sim \mathcal{GP}(m(\mathbf{z}),\; k_{\mathrm{feat}}(\mathbf{z}, \mathbf{z}')),
\end{equation}
where $k_{\mathrm{feat}}(\mathbf{z}, \mathbf{z}')$ is a kernel defined on the feature space, for example,
\begin{equation}
    k_{\mathrm{feat}}(\mathbf{z}, \mathbf{z}') = \sigma_{\mathrm{feat}}^2 \exp\left(-\frac{\|\mathbf{z} - \mathbf{z}'\|^2}{2\ell_{\mathrm{feat}}^2}\right).
\end{equation}
In this setting, the GP is trained to minimize prediction error for the residuals $r_i$ using only the latent features.

To further improve calibration, recent work by \cite{b29, b30} proposes augmenting the GP input with the network confidence $s_i$, yielding a joint input vector
\begin{equation}
    \mathbf{x}_i =
    \begin{bmatrix}
        \mathbf{z}_i \\
        s_i
    \end{bmatrix}
    \in \mathbb{R}^{d+1}.
\end{equation}
The GP prior is then defined as
\begin{equation}
    f(\mathbf{x}) \sim \mathcal{GP}(m(\mathbf{x}),\; k(\mathbf{x}, \mathbf{x}')),
\end{equation}
where the kernel is constructed as the sum of independent feature and confidence kernels:
\begin{equation}
    k(\mathbf{x}, \mathbf{x}') = \sigma_{\mathrm{feat}}^2\, k_{\mathrm{feat}}(\mathbf{z}, \mathbf{z}') + \sigma_{\mathrm{conf}}^2\, k_{\mathrm{conf}}(s, s').
\end{equation}
Here, $k_{\mathrm{conf}}(s, s')$ adds another kenrel in relation to softmax confidence input space, for example,
\begin{equation}
    k_{\mathrm{conf}}(s, s') = \exp\left(-\frac{(s - s')^2}{2\ell_{\mathrm{conf}}^2}\right).
\end{equation}
This input+output kernel construction allows the GP to capture both feature similarity and confidence similarity for improved residual calibration.

Given a test sample with latent features $\mathbf{z}_*$ and softmax confidence $s_*$, the GP predicts the posterior distribution of the residual,
\begin{equation}
    r_* \sim \mathcal{N}(\bar{r}_*, \operatorname{Var}(r_*)),
\end{equation}
where $\bar{r}_*$ and $\operatorname{Var}(r_*)$ denote the GP posterior mean and variance, respectively. The calibrated confidence is then given by
\begin{equation}
    s_*' \sim \mathcal{N}(s_* + \bar{r}_*,\; \operatorname{Var}(r_*)).
\end{equation}

The feature-only kernel utilizes the latent representation from a fixed layer for calibration, potentially overlooking information contained in the model’s output confidence. In contrast, the input+output kernel, as proposed in \cite{b29, b30}, incorporates both the feature representation and network confidence, resulting in improved uncertainty calibration compared to the feature-only approach.

\subsection{Deep Gaussian Process (DGP)}
Standard GP-based calibration methods rely on a single pair of sequential input vectors from latent feature representations, which limits their ability to effectively model the decision-making process of the neural network classifier, though it keeps them close to the prior function. In contrast, DGPs propagate outputs as inputs to subsequent hidden layers in latent space because each layer $h^{(l)}$ in a DGP is treated as a latent variable \cite{b63, b64, b65} .
The input $x$ is passed through multiple GP layers, not as deterministic functions, but as probabilistic mappings:

\begin{align}
    h^{(1)} &\sim \mathcal{GP}_1(x), \notag\\
    h^{(2)} &\sim \mathcal{GP}_2(h^{(1)}), \notag\\
    &\;\;\vdots \notag\\
    y &\sim \mathcal{GP}_L(h^{(L-1)})
\end{align}

In order to estimate the marginal predictive distribution $p(y \mid x)$, integration latent variable over all layers is required.

\begin{equation}
p(y \mid x) = \int p(y \mid h^{(L-1)}) \cdots p(h^{(1)} \mid x) \, dh^{(1)} \cdots dh^{(L-1)}
\end{equation}

This integral is not tractable in closed form due to the hierarchical composition of non-linear, non-Gaussian functions at each layer. Specifically, the mappings between layers are non-linear and non-Gaussian after the first GP layer. In addition, each intermediate layer introduces a latent function, which must be integrated over, but the outputs after the first GP layer are no longer Gaussian-distributed. Furthermore, there is no closed-form expression for the marginalization of multiple nested GPs over intermediate layers.
As a result, approximate inference techniques such as variational inference (e.g., doubly stochastic variational inference) or Monte Carlo sampling are required to estimate the predictive distribution $\hat{p}(y \mid x)$ \cite{b65}.

As explained earlier, this intractability poses a significant limitation for confidence calibration as well as not being able to condition for intermediate layer's latent variables.

\subsection{Hierachical Gaussian Process}

Beyond the standard GP, we consider a HGP model with three levels of latent function composition to capture multiple levels of structure in the data \cite{b60, b61}. We can define the top-level function $f_p(x)$ as a global latent function that models the general response behavior of $p^{th}$ pattern cluster. The mid-level function $g_q(x)$ captures the cluster-specific deviations at the second hierarchical level, corresponding to the $q^{th}$ sub-cluster within the $p^{th}$ cluster. Finally, the bottom-level function $h_{qr}(x)$ represents the lowest-level variation within the constructed hierarchical structure, corresponding to the $r^{th}$ nested subcomponent of the $q^{th}$ mid-level cluster. 

Each level is modeled as a GP, where the output of a higher-level function conditioning on the lower-level function with drawn from its own corresponding kernel, which serves as the mean function for the GP at the next level:

\begin{equation}
f_p(x) \sim \mathcal{GP}\left( \mu(x), k_f(x, x') \right)
\end{equation}

\begin{equation}
g_q(x) \sim \mathcal{GP}\left( f_p(x), k_g(x, x') \right)
\end{equation}

\begin{equation}
h_{qr}(x) \sim \mathcal{GP}\left( g_q(x), k_h(x, x') \right)
\end{equation}

Then the joint distribution over all functions in three-layers hierarchy, for a fixed top-level cluster $p$, can be written compactly as with $\mu(x) = 0$ for simplicity:

\begin{equation}
\begin{bmatrix}
f_p(x) \\
g_1(x) \\
\vdots \\
g_{q}(x) \\
\vdots \\
g_{N_q}(x) \\
h_{11}(x) \\
\vdots \\
h_{qr}(x) \\
\vdots \\
h_{N_q N_r}(x)
\end{bmatrix}
\sim \mathcal{N}\left( \mathbf{0}, 
\begin{bmatrix}
K_f & K_{fg} & K_{fh} \\
K_{gf} & K_g & K_{gh} \\
K_{hf} & K_{hg} & K_h
\end{bmatrix}
\right)
\end{equation}

In this formulation, $K_f$ denotes the covariance matrix of the top-level global latent function $f_p(x)$. The block $K_g$ is a block-diagonal matrix representing the covariances of the mid-level functions $g_q(x)$, each of which is conditionally dependent on $f_p(x)$. Similarly, $K_h$ is a block-diagonal matrix corresponding to the covariances of the bottom-level functions $h_{qr}(x)$, which are conditionally dependent on their respective parent functions $g_q(x)$. The off-diagonal blocks $K_{fg}$, $K_{fh}$, and $K_{gh}$, along with their transposes, encode the cross-covariance terms between adjacent layers in the hierarchy, capturing the dependency structure across the three levels. $N_q$ and $N_r$ denote the number of mid-level and bottom-level functions per mid-level function, respectively. This implies that the total number of mid-level functions $g_q(x)$ is $N_q$, while the total number of bottom-level functions $h_{qr}(x)$ is $N_q \times N_r$.

This block structure of $\Sigma$—a full, symmetric, and positive semi-definite covariance matrix—encodes the hierarchical dependencies across the levels.

\begin{equation}
\Sigma =
\begin{bmatrix}
K_f & K_{fg} & K_{fh} \\
K_{gf} & K_g & K_{gh} \\
K_{hf} & K_{hg} & K_h
\end{bmatrix}
\end{equation}

Specifically, each block covariance matrix can be expanded:

\begin{equation}
K_f = k_f(X, X)
\end{equation}

\begin{align}
K_{fg} &=
\begin{bmatrix}
k_f(X, X) & \cdots & k_f(X, X)
\end{bmatrix}
\in \mathbb{R}^{n \times (N_q n)} \\
K_{gf} &= K_{fg}^\top
\end{align}

\begin{equation}
K_g = \mathrm{blockdiag}(K_{g_1}, K_{g_2}, \dots, K_{g_{N_q}})
\end{equation}

\begin{equation}
K_{g_q} = k_g(X, X)
\end{equation}

\begin{equation}
K_{fh} = 
\begin{bmatrix}
k_f(X, X) & \cdots & k_f(X, X)
\end{bmatrix}
\in \mathbb{R}^{n \times (N_q N_r n)}
\end{equation}

\begin{equation}
K_{hf} = K_{fh}^\top
\end{equation}

\begin{equation}
K_{gh} =
\begin{bmatrix}
k_g(X, X) & 0 & \cdots & 0 \\
0 & k_g(X, X) & \cdots & 0 \\
\vdots & \vdots & \ddots & \vdots \\
0 & 0 `& \cdots & k_g(X, X)
\end{bmatrix}
\end{equation}

\begin{equation}
K_{hg} = K_{gh}^\top
\end{equation}

\begin{equation}
K_h = \mathrm{blockdiag}(K_{h_{11}}, K_{h_{12}}, \dots, K_{h_{N_q N_r}}), 
\end{equation}

\begin{equation}
K_{h_{qr}} = k_h(X, X)
\end{equation}

Each block encodes the covariance between functions at different levels in the hierarchy, respecting conditional dependencies through shared kernel structure. Each mid-level function $g_q(x)$ depends on the top-level function $f_p(x)$, and each bottom-level function $h_{qr}(x)$ depends on its corresponding $g_q(x)$. Appendix 2 presents the relationships among covariance kernels in a HGP and demonstrates that two points from the bottom-level local GP are jointly Gaussian, with a covariance given by the sum of the covariance functions from the higher-level GP.

The observed data are structured hiearchically as well. The output at the top level, $\hat{y}_p$, is formed by concatenating outputs from the mid-level components $\hat{y}_q$, each of which in turn consists of outputs from the bottom level $\hat{y}_r$:

where

\begin{equation}
\hat{y}_p = [\hat{y}_1, \ldots, \hat{y}_q, \ldots, \hat{y}_{N_q}]
\end{equation}

\begin{equation}
\hat{y}_q = [\hat{y}_1, \ldots, \hat{y}_r, \ldots, \hat{y}_{N_r}]
\end{equation}

This hierarchical GP framework enables the model to capture structured variability across multiple levels, with the flexibility to reduce or extend the hierarchy depending on the presence of additional structure among subcomponents \cite{b60}.

\subsection{Architecture of Layerwise Gaussian Process}

\begin{figure*}[ht]
    \centering
    \begin{subfigure}{0.48\textwidth}  
        \centering
        \includegraphics[width=\linewidth]{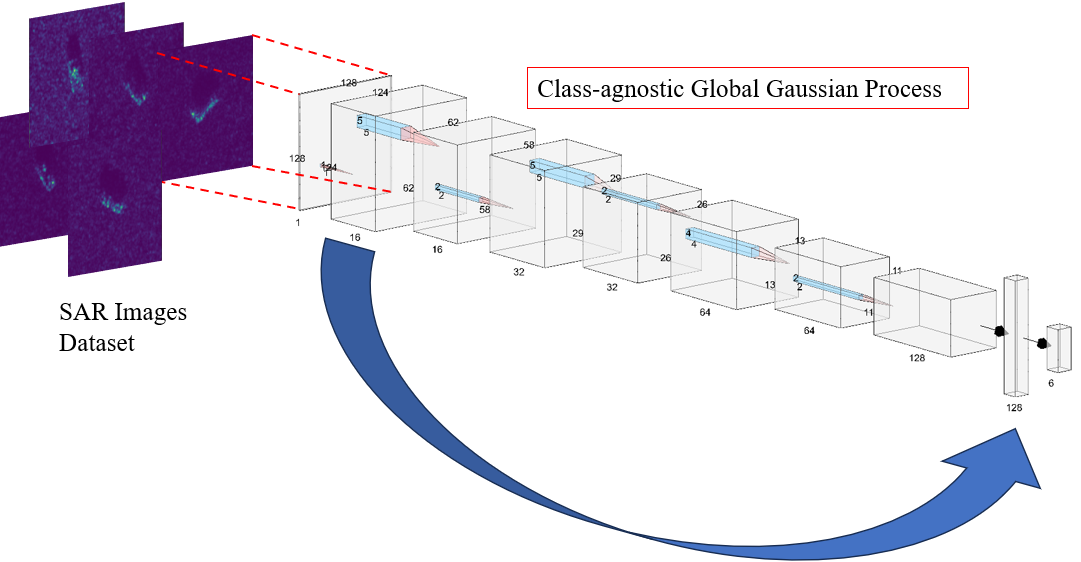}
        \caption{Class-agnostic Global GP}
        \label{fig:1}
    \end{subfigure}
    \begin{subfigure}{0.48\textwidth}  
        \centering
        \includegraphics[width=\linewidth]{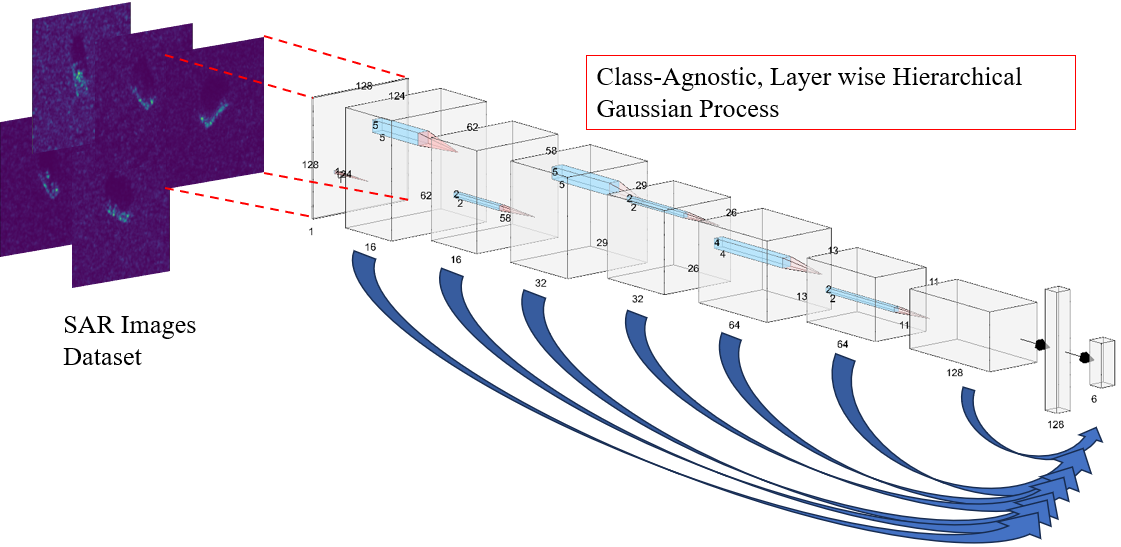}
        \caption{Class-agnostic Layer-wise HGP}
        \label{fig:2}
    \end{subfigure}
    
    \vspace{0.5cm}  
    
    \begin{subfigure}{0.48\textwidth}  
        \centering
        \includegraphics[width=\linewidth]{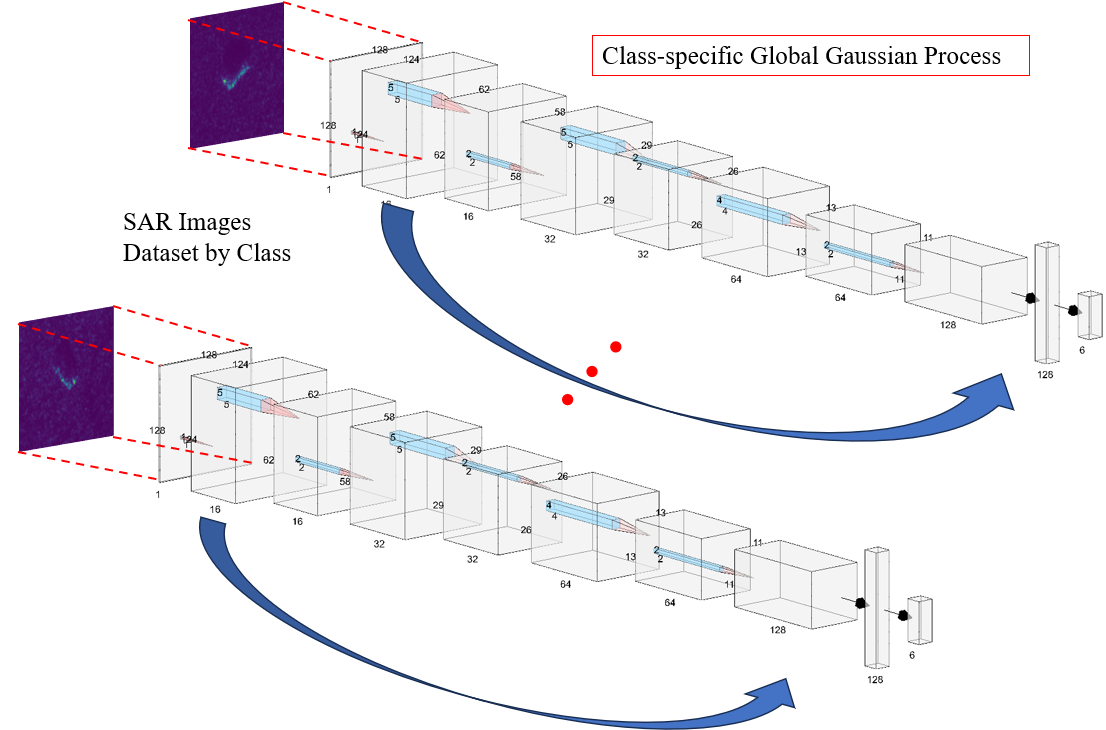}
        \caption{Class-specific Global HGP}
        \label{fig:3}
    \end{subfigure}
    \begin{subfigure}{0.48\textwidth}  
        \centering
        \includegraphics[width=\linewidth]{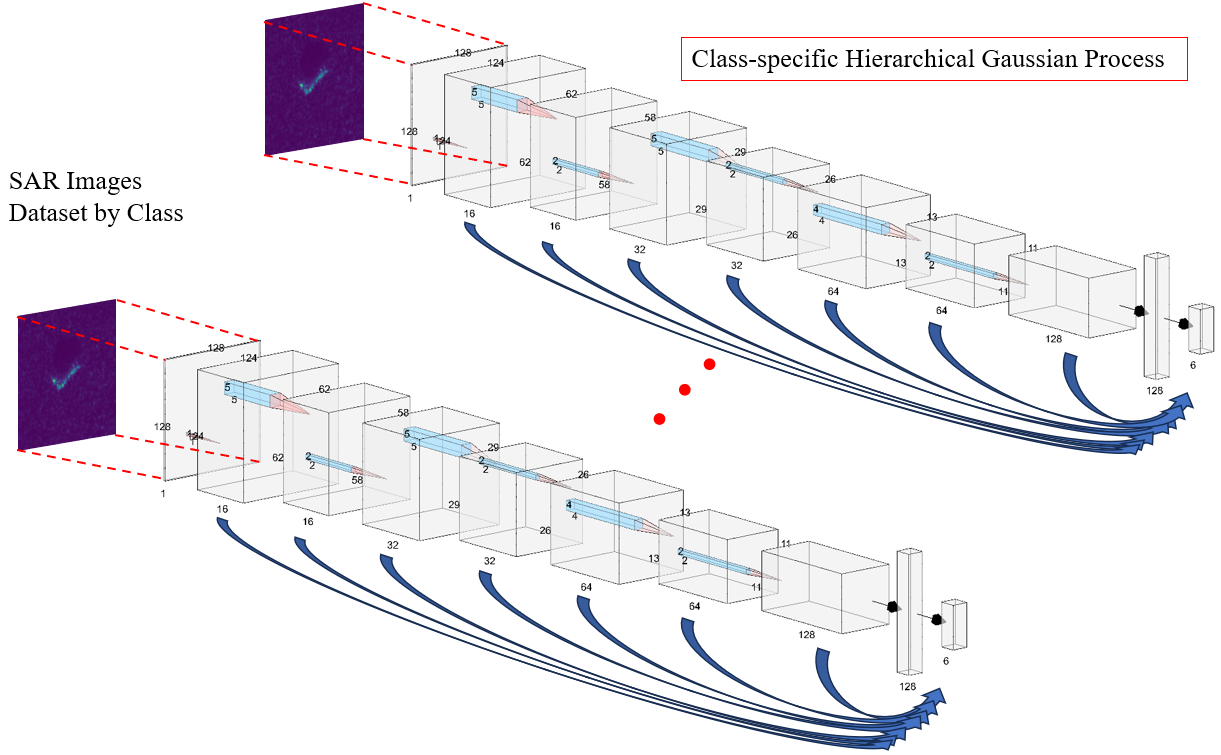}
        \caption{Class-specific Layer-wise HGP}
        \label{fig:4}
    \end{subfigure}

    \caption{Four possible interpretations for constructing a Hierarchical Gaussian Process based on class and neural network layer.}
    \label{fig:four_possible_interpretations_HGP}
\end{figure*}

Fig.~\ref{fig:four_possible_interpretations_HGP} illustrates possible GP structures, incorporating hierarchical or non-hierarchical configurations based on class and layer, to calibrate neural networks depending on the interpretation of input-output systems.
The figure categorizes these structures along two dimensions. The first and second columns distinguish between treating the entire network as a single, global GP versus constructing a hierarchical layer-wise GP. The first and second rows differentiate between class-agnostic and class-specific formulations, where the latter introduces an additional level of complexity by modeling patterns at the class level.

\subsubsection{Class-agnostic Global GP (top-left panel)}

A non-HGP in which the first neural network layer’s output for all training data—regardless of class labels—is used as input, with the residual correctness score and softmax output serving as the GP's output.
\begin{equation}
f(x) \sim \mathcal{GP}\left( \mu(x),\ k_f(x,\ x') \right)
\end{equation}

Here, \( f(x) \) is the global GP response function defined over all training data, regardless of class. The mean function \( \mu(x) \) represents the prior expectation of the function value at input \( x \), and the kernel \( k_f(x, x') \) defines the covariance structure across the input space, capturing global similarity between input points.

\subsubsection{Class-agnostic Layer-wise HGP (top-right panel)}

A hierarchical layer-wise GP, where each layer has an independent local GP. These layer-wise processes are structured with a global GP prior mean but remain independent in their responses.
\begin{equation}
f_\ell(x) \sim \mathcal{GP}\left( \mu(x),\ k_f(x,\ x') \right)
\end{equation}
\begin{equation}
g_{\ell r}(x) \sim \mathcal{GP}\left( f_\ell(x),\ k_g(x,\ x') \right)
\end{equation}

In this formulation, \( f_\ell(x) \) denotes the layer-specific GP at neural network layer \( \ell \), modeling shared structure across all classes. The function \( g_{\ell r}(x) \) captures residual variation indexed by \( r \) within layer \( \ell \), and is modeled as a GP whose mean is given by \( f_\ell(x) \). The kernels \( k_f(x, x') \) and \( k_g(x, x') \) define the intra-layer and residual-level covariance structures, respectively.

\subsubsection{Class-specfic Global HGP (bottom-left panel)}

A hierarchical GP where each class-label group forms a distinct local GP, establishing a hierarchy within the global structure.
\begin{equation}
f_c(x) \sim \mathcal{GP}\left( \mu(x),\ k_f(x,\ x') \right)
\end{equation}
\begin{equation}
g_{c r}(x) \sim \mathcal{GP}\left( f_c(x),\ k_g(x,\ x') \right)
\end{equation}

In the class-specific global model, \( f_c(x) \) represents the top-level GP for class \( c \), capturing the general behavior of samples within that class. The function \( g_{cr}(x) \) models fine-scale variations or subcomponents indexed by \( r \) within class \( c \), conditioned on \( f_c(x) \). The kernels \( k_f(x, x') \) and \( k_g(x, x') \) define the class-level and intra-class covariance structures, respectively.

\subsubsection{Class-specific Layer-wise HGP (bottom-right panel)}

A hierarchical GP model with both layer-wise and class-wise structures, where the global GP mean serves as a prior, and local GPs are defined at both levels.
\begin{equation}
f_\ell(x) \sim \mathcal{GP}\left( \mu(x),\ k_f(x,\ x') \right)
\end{equation}
\begin{equation}
g_c(x) \sim \mathcal{GP}\left( f_\ell(x),\ k_g(x,\ x') \right)
\end{equation}
\begin{equation}
h_{cr}(x) \sim \mathcal{GP}\left( g_c(x),\ k_h(x,\ x') \right)
\end{equation}

This hierarchical formulation combines both layer-wise and class-specific modeling. The function \( f_\ell(x) \) is the global GP defined at layer \( \ell \), shared across all classes. The function \( g_c(x) \) is the class-specific GP for class \( c \), conditioned on the global layer GP. Finally, \( h_{cr}(x) \) captures localized variations within class \( c \) for subcomponent \( r \), and is conditioned on \( g_c(x) \). The kernels \( k_f(x, x') \), \( k_g(x, x') \), and \( k_h(x, x') \) govern the dependencies across the global layer, class-specific, and residual levels, respectively.

\begin{figure*}[t]
  \centering
  \includegraphics[width=\textwidth]{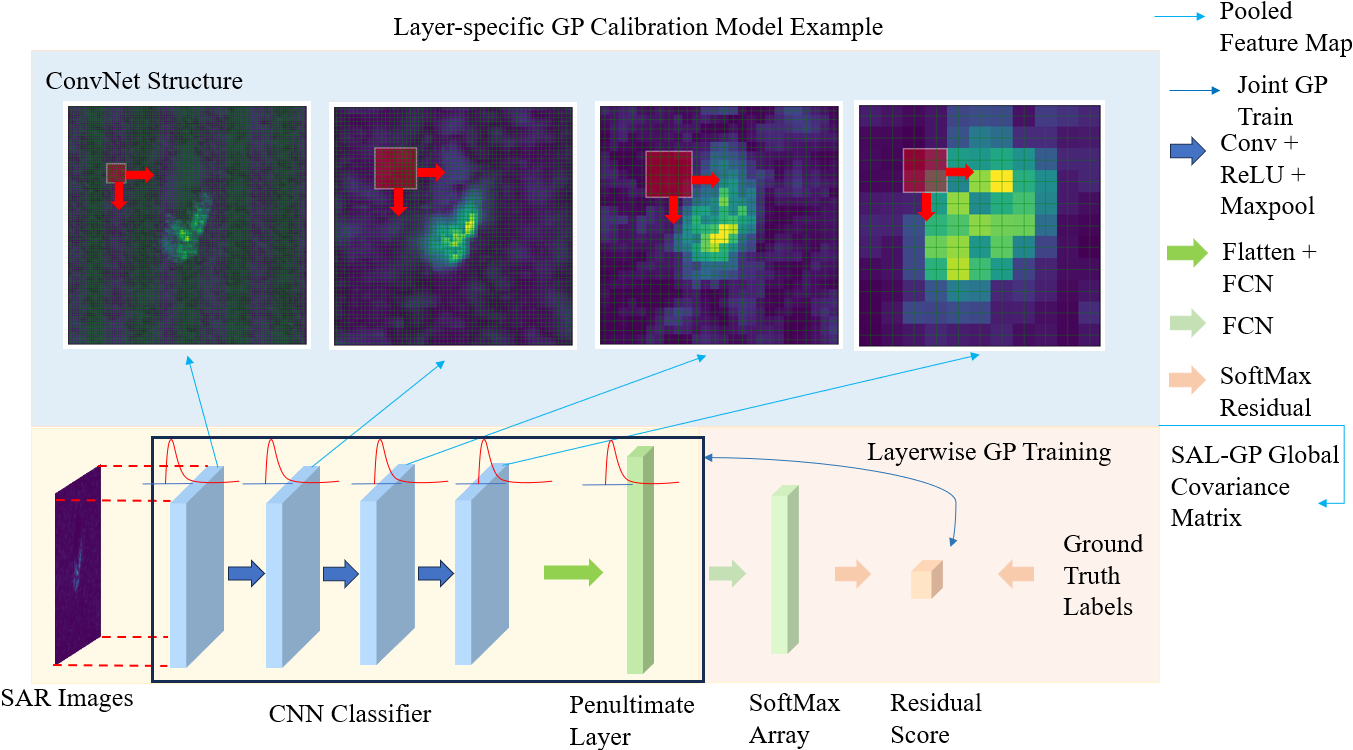}
  \caption{Layer-wise Semantic Representation Architecture of SAL-GP}
  \label{fig:Diagram of SAL-GP}
\end{figure*}

In our formalism, we adopt Fig.~\ref{fig:2} the class-agnostic layer-wise HGP architecture, focusing solely on the semantic structure of layerwise intra-connections. Although it is possible to construct class-wise local GPs within the hierarchy, this approach is less suitable than leveraging layerwise latent representations for calibration. We introduce two variants of the SAL-GP framework, both motivated by the design in Fig.~\ref{fig:2}; a detailed schematic of the proposed architecture is presented in Fig.~\ref{fig:Diagram of SAL-GP}.

\subsection{SAL-GP variant: Hierarchical Layer Kernel}
\label{sec:salgp_hierarchical_layer} 

Each data sample is represented by a set of feature vectors from each neural network layer, the maximum softmax value, and a layer index. The composite input vector is
\begin{equation}
    \mathbf{x}_i^{(\ell)} = \left[\, \mathbf{f}_i^{(\ell)}\,;\; s_i\,;\; \ell\, \right]
\end{equation}
where $\mathbf{f}_i^{(\ell)} \in \mathbb{R}^D$ denotes the feature vector for sample $i$ at layer $\ell$, zero-padded to a common width $D$; $s_i \in \mathbb{R}$ is the maximum softmax score for sample $i$; and $\ell$ represents the layer index, which is treated as a categorical or layer-specific variable. This joint input space is used for both training and prediction.


In Hierarchical structured GP modeling, it is often necessary to capture both global relationships across all outputs as well as local dependencies specific to certain groups or hierarchical structures \cite{b60, b61}. The general form of a HGP kernel for an input tuple $\big(\mathbf{x}, \boldsymbol{\ell}\big)$, where $\boldsymbol{\ell} = (\ell^{(1)}, \ldots, \ell^{(J)})$ denotes a vector of hierarchical group indices across $J$ levels, is given by
\begin{equation}
    k\big((\mathbf{x}, \boldsymbol{\ell}), (\mathbf{x}', \boldsymbol{\ell}')\big) = k_0(\mathbf{x}, \mathbf{x}') + \sum_{j=1}^J \delta_{\ell^{(j)} \ell'^{(j)}}\, k_j(\mathbf{x}, \mathbf{x}')
\end{equation}
where $k_0(\mathbf{x}, \mathbf{x}')$ represents the global (shared) kernel across all groups, $k_j(\mathbf{x}, \mathbf{x}')$ is a kernel specific to the $j$-th hierarchical group, and $\delta_{\ell^{(j)} \ell'^{(j)}}$ is the Kronecker delta that is 1 when the $j$-th group indices are equal, and 0 otherwise. This additive structure allows the GP to flexibly model correlations arising from different levels of hierarchy \cite{b60}.

In this work, we consider the case where there is a single local group structure corresponding to neural network layers (i.e., $J = 1$), which leads to a two-level hierarchical kernel comprising a global component and a layerwise local component. This specialized kernel is detailed below

A hierarchical kernel is employed to couple global (layer-agnostic) and local (layer-specific) similarity:
\begin{equation}
    k\big((\mathbf{x}, \ell), (\mathbf{x}', \ell')\big) = k_{\mathrm{global}}(\mathbf{x}, \mathbf{x}') + \delta_{\ell\ell'}\, k_{\mathrm{layer}}(\mathbf{x}, \mathbf{x}')
\end{equation}
Here, $k_{\mathrm{global}}$ denotes a stationary kernel function (such as the Matern or RBF kernel) that models similarity across all layers, while $k_{\mathrm{layer}}$ is a stationary kernel that captures similarity within individual layers. The term $\delta_{\ell\ell'}$ represents the Kronecker delta, which equals 1 when $\ell = \ell'$, and 0 otherwise.

For two input vectors $\mathbf{x}_1 = (\mathbf{f}_1, s_1, \ell_1)$ and $\mathbf{x}_2 = (\mathbf{f}_2, s_2, \ell_2)$, the kernel becomes
\begin{align}
    k(\mathbf{x}_1, \mathbf{x}_2) =\; & 
    k_{\mathrm{global}}\big([\mathbf{f}_1, s_1], [\mathbf{f}_2, s_2]\big) \nonumber \\
    & +\, \delta_{\ell_1, \ell_2}\; k_{\mathrm{layer}}\big([\mathbf{f}_1, s_1], [\mathbf{f}_2, s_2]\big)
\end{align}
where the layer index is excluded from the base kernels.

We implement GP regression using the hierarchical layerwise kernel and the joint input space $[\mathbf{f}, s, \ell]$, with the calibration residual as the regression target. The posterior mean and variance for a test input $(\mathbf{z}_*, \ell_*)$ are computed using the GP regression formulas with the hierarchical kernel and structured inputs.

Local (layerwise) predictions are computed at each $(\mathbf{z}, \ell)$. For global (layer-agnostic) predictions, we use a test layer index $\ell_* = -1$ (absent from training), so $\delta_{\ell_*, \ell_i} = 0$ for all training points. In this case,
\begin{align}
    k\big((\mathbf{z}_*, \ell_*), (\mathbf{z}_i, \ell_i)\big) &= k_{\mathrm{global}}(\mathbf{z}_*, \mathbf{z}_i), \\
    k\big((\mathbf{z}_*, \ell_*), (\mathbf{z}_*, \ell_*)\big) &= k_{\mathrm{global}}(\mathbf{z}_*, \mathbf{z}_*) + k_{\mathrm{layer}}(\mathbf{z}_*, \mathbf{z}_*)
\end{align}
The corresponding global mean and variance are
\begin{align}
    \mu_{\mathrm{global}}(\mathbf{z}_*) &= \mathbf{k}_{\mathrm{global}}^\top ( K + \sigma^2 I )^{-1} \mathbf{y}, \\
    \sigma^2_{\mathrm{global}}(\mathbf{z}_*) &= k_{\mathrm{global}}(\mathbf{z}_*, \mathbf{z}_*) - \mathbf{k}_{\mathrm{global}}^\top ( K + \sigma^2 I )^{-1} \mathbf{k}_{\mathrm{global}}
\end{align}
where $\mathbf{k}_{\mathrm{global}} = [k_{\mathrm{global}}(\mathbf{z}_*, \mathbf{z}_i)]_{i=1}^n$.

Because $k_{\mathrm{layer}}(\mathbf{z}_*, \mathbf{z}_*)$ does not contribute to global uncertainty, the pure global predictive variance is
\begin{align}
    \mathrm{Var}\big[ f_{\mathrm{global}}(\mathbf{z}_*) \big]
    &= 
    \left\{
        k_{\mathrm{global}}(\mathbf{z}_*, \mathbf{z}_*) 
        + k_{\mathrm{layer}}(\mathbf{z}_*, \mathbf{z}_*) \right. \notag \\
    &\left. 
        \quad -\, \mathbf{k}_{\mathrm{global}}^\top 
        ( K + \sigma^2 I )^{-1} \mathbf{k}_{\mathrm{global}}
    \right\} \notag \\
    &\quad
    -\, k_{\mathrm{layer}}(\mathbf{z}_*, \mathbf{z}_*)
\end{align}
This quantity reflects uncertainty attributable solely to the global GP structure.

\subsection{SAL-GP variant: Reduced Multi-layer Kernel from ICM Kernel}
\label{sec:reduced-multilayer-kernel}

While the hierarchical layerwise GP models each layer's feature representations as local GPs, organizing these as subordinate to a higher-level GP can restrict the ability to represent unique, layer-specific responses, since each local GP may be constrained by the overarching hierarchical dependencies. In this regard, a more flexible way to incorporate multi-layer kernels is through the ICM kernel, as in the design analogous to Fig.~\ref{fig:2}.

To jointly model correlations across neural network layers, we employ the ICM kernel \cite{b53, b55} for layerwise kernel. For input tuples $(\mathbf{x}, \ell)$ and $(\mathbf{x}', \ell')$, where $\mathbf{x}, \mathbf{x}'$ are input features (possibly including softmax/confidence values), and $\ell, \ell'$ are discrete layer indices, the ICM kernel is defined as
\begin{equation}
    k_{\mathrm{ICM}}\big((\mathbf{x}, \ell), (\mathbf{x}', \ell')\big)
    = k_{\mathrm{feat}}(\mathbf{x}, \mathbf{x}') \; B_{\ell \ell'},
\end{equation}
where \(k_{\mathrm{feat}}(\mathbf{x}, \mathbf{x}')\) denotes a standard positive-definite kernel function on the input features (such as the RBF or Matern kernel), and \(B \in \mathbb{R}^{L \times L}\) is a learnable, positive semi-definite coregionalization matrix that encodes the inter-layer correlations, with \(L\) denoting the total number of layers.

The total covariance matrix for $N$ samples and $L$ layers is then of size $(NL) \times (NL)$, and can be expressed as the Kronecker product
\begin{equation}
    K_{\mathrm{ICM}} = K_{\mathrm{feat}} \otimes B,
\end{equation}
where $K_{\mathrm{feat}}$ is the $N \times N$ kernel matrix over features, and $B$ encodes all pairwise correlations among layers.

To further increase expressivity, the Linear Model of Coregionalization (LCM) with the multi-layer covariance as a sum of $Q$ separable (ICM) kernels \cite{b53, b55}:
\begin{equation}
    k_{\mathrm{LCM}}\big((\mathbf{x}, \ell), (\mathbf{x}', \ell')\big)
    = \sum_{q=1}^Q k_{\mathrm{feat}}^{(q)}(\mathbf{x}, \mathbf{x}') \; B^{(q)}_{\ell \ell'},
\end{equation}
where each $k_{\mathrm{feat}}^{(q)}$ is a distinct base kernel, and $B^{(q)}$ is a corresponding coregionalization matrix for latent process $q$. The resulting covariance matrix is
\begin{equation}
    K_{\mathrm{LCM}} = \sum_{q=1}^Q K_{\mathrm{feat}}^{(q)} \otimes B^{(q)}.
\end{equation}
As $Q$ increases, the LCM is capable of modeling highly complex and non-separable inter-layer dependencies. However, both computational cost and memory usage grow rapidly with $Q$, $L$, and $N$, rendering LCM with large $Q$ impractical for deep networks with many layers, especially when combined with high-dimensional input feature kernels.

To alleviate the computational burden of full ICM/LCM kernels, one can consider an additive kernel structure as computational approximation:
\begin{equation}
    k_{\mathrm{add}}\big((\mathbf{x}, \ell), (\mathbf{x}', \ell')\big)
    = k_{\mathrm{global}}(\mathbf{x}, \mathbf{x}') 
    + \delta_{\ell \ell'}\, k_{\mathrm{layer}}(\mathbf{x}, \mathbf{x}')
\end{equation}
where $k_{\mathrm{global}}$ is a kernel shared across all layers, and $k_{\mathrm{layer}}$ is a kernel restricted to within-layer pairs (activated only when $\ell = \ell'$ via the Kronecker delta). This construction yields a block-diagonal or block-sparse covariance matrix, significantly reducing both memory and computational cost:
\begin{equation}
    K_{\mathrm{add}} = K_{\mathrm{global}} \otimes I_L + \bigoplus_{\ell=1}^{L} K_{\mathrm{layer}}^{(\ell)}
\end{equation}
where $I_L$ is the $L \times L$ identity, and $\bigoplus$ denotes the block-diagonal sum across layers.

The ICM kernel reduces to the additive layerwise kernel under specific constraints. If the coregionalization matrix $B$ is diagonal, i.e., $B = \operatorname{diag}(\beta_1, \ldots, \beta_L)$, then
\begin{align}
    k_{\mathrm{ICM}}\big((\mathbf{x}, \ell), (\mathbf{x}', \ell')\big)
    &= k_{\mathrm{feat}}(\mathbf{x}, \mathbf{x}') \cdot \delta_{\ell \ell'}\, \beta_\ell \\
    &= \delta_{\ell \ell'}\, \left[\beta_\ell k_{\mathrm{feat}}(\mathbf{x}, \mathbf{x}') \right]
\end{align}
This form corresponds to a sum of independent, layerwise GPs—i.e., the additive kernel with no global/shared component. If one further adds a global kernel across all layers (i.e., $B = \alpha \mathbf{1} \mathbf{1}^\top + \operatorname{diag}(\beta_1, \ldots, \beta_L)$), then the kernel becomes
\begin{equation}
    k((\mathbf{x}, \ell), (\mathbf{x}', \ell')) = \alpha\, k_{\mathrm{feat}}(\mathbf{x}, \mathbf{x}') + \delta_{\ell \ell'}\, \beta_\ell k_{\mathrm{feat}}(\mathbf{x}, \mathbf{x}')
\end{equation}
which is the sum of a global (shared) GP and layer-specific (local) GPs.

While the ICM (rank-1 LCM) flexibly models both shared and inter-layer correlations through a coregionalization matrix, it incurs substantial computational cost as the number of layers $L$ increases. The LCM further enhances expressivity by incorporating multiple latent processes, but its computational demands grow even more rapidly, often becoming prohibitive for practical applications. In contrast, the additive kernel provides a scalable and interpretable approximation of ICM by discarding off-diagonal (cross-layer) terms in $B$, thus enabling efficient layerwise GP inference. However, this simplification comes at the expense of ignoring direct statistical coupling between layers, which may be important for further reducing calibration error in certain scenarios.

\section{Experimental Setup}

\subsection{Network backbones}
We evaluate the proposed SAL-GP calibration framework across a range of neural network architectures, including a standard sequential CNN \cite{b76}, a fully convolutional network without dense layers \cite{b49}, a deep residual network \cite{b3} for image-based tasks, and a modified RNN \cite{b73, b74, b75} for time-series astronomical data \cite{b77, b78}. This diversity enables a comprehensive assessment of calibration performance and uncertainty estimation across models of varying complexity.

\begin{enumerate}
\item \textbf{Standard Convolutional Neural Network (CNN) \cite{b76}:} Standard CNN is a classic four-layer convolutional neural network tailored for moderate-sized input images (128×128). The architecture consists of a sequence of convolutional and max-pooling layers, progressively increasing channel depth from 16 to 128 while reducing spatial resolution. After the final convolutional block, feature maps are flattened and passed through a fully connected layer for further feature abstraction before classification. Dropout regularization is applied to mitigate overfitting. The network outputs both class logits and softmax probabilities, as well as feature maps from each intermediate layer to support layerwise calibration.

\begin{align*}
&\rightarrow \mathrm{Conv}(16) \rightarrow \mathrm{MaxPool} 
\rightarrow \mathrm{Conv}(32) \rightarrow \mathrm{MaxPool} \\
&\rightarrow \mathrm{Conv}(64) \rightarrow \mathrm{MaxPool} 
\rightarrow \mathrm{Conv}(128) \rightarrow \mathrm{Dropout} \\
&\rightarrow \mathrm{Flatten} \rightarrow \mathrm{Linear}(128)
\rightarrow \mathrm{Classifier} \rightarrow \mathrm{Softmax}
\end{align*}

(Intermediate features are extracted after each convolutional layer.)

\item \textbf{All-Convolutional Network (Modified AConvNet) \cite{b49}:} Modified AConvNet implements a purely convolutional architecture that eschews fully connected layers in favor of deep hierarchical feature extraction. The model comprises five successive convolutional blocks, interleaved with pooling and dropout, with channel dimensions increasing up to 128 and then reduced before the final classification stage. The final classification is achieved using a 1×1 convolution over the feature map, followed by global flattening. This design enables end-to-end feature learning and provides rich spatial representations at multiple layers. All intermediate feature maps are retained for downstream calibration analysis.

\begin{align*}
&\rightarrow \mathrm{Conv}(16) \rightarrow \mathrm{MaxPool}
\rightarrow \mathrm{Conv}(32) \rightarrow \mathrm{MaxPool} \\
&\rightarrow \mathrm{Conv}(64) \rightarrow \mathrm{MaxPool}
\rightarrow \mathrm{Conv}(128) \rightarrow \mathrm{Conv}(32) \\
&\rightarrow \mathrm{Dropout} \rightarrow \mathrm{Conv}(n_\mathrm{classes})
\rightarrow \mathrm{Flatten} \rightarrow \mathrm{Softmax}
\end{align*}

(Intermediate features are extracted after each convolutional layer.)

\item \textbf{Residual Network (ResNet-18) \cite{b3}} ResNet-18 follows the widely adopted residual learning paradigm, featuring skip connections to alleviate vanishing gradient problems in deep networks. The architecture begins with an initial convolutional stem, followed by four residual stages with increasing channel widths (64, 128, 256, 512), each comprising multiple Basic Block units with identity shortcuts. Global average pooling condenses the feature map prior to a final fully connected classifier. The design allows extraction of pre-activation feature maps at each major stage, aligning naturally with the hierarchical calibration scheme.

\begin{align*}
&\rightarrow~\mathrm{Conv}(64) \rightarrow~\mathrm{BatchNorm} \rightarrow~\mathrm{ReLU} \rightarrow~\mathrm{MaxPool} \\
&\rightarrow~[\mathrm{ResBlock} \times 2,~64] \rightarrow~[\mathrm{ResBlock} \times 2,~128] \\
&\rightarrow~[\mathrm{ResBlock} \times 2,~256] \rightarrow~[\mathrm{ResBlock} \times 2,~512] \\
&\rightarrow~\mathrm{GlobalAvgPool} \rightarrow~\mathrm{Linear} \rightarrow~\mathrm{Softmax}
\end{align*}
(Intermediate feature maps are extracted after each major residual layer.)

\item \textbf{Recurrent Neural Network (RNN) \cite{b73, b74, b75, b80}} 

The time-series RNN model is a multi-passband, multi-layer classifier consisting of six independent 5-layer bidirectional Gated Recurrent Units (GRUs), one for each of the ugrizy passbands. Each passband is processed separately through its own stack of five bidirectional GRU layers, enabling the model to capture hierarchical temporal patterns in both forward and backward directions. Following sequence encoding, temporal max-pooling is applied to the outputs of the final GRU layer for each passband, reducing variable-length sequences to fixed-size summary vectors. Simultaneously, the hidden states from all GRU layers are retained to preserve intermediate temporal representations for downstream calibration and uncertainty analysis. The max-pooled passband outputs are then concatenated and fused with auxiliary metadata, such as positional and photometric information, to enable context-aware classification. This combined feature vector is passed through two fully connected layers with batch normalization and ReLU activation to learn higher-level representations, followed by a final linear layer that produces class logits. These logits are transformed via softmax to generate probabilistic predictions over the target classes \cite{b80}.

\begin{align*}
&\text{(a) For each passband (6 total):} \\
&\rightarrow \mathrm{5\text{-}layer\ BiGRU} \\
&\rightarrow \mathrm{MaxPool\ over\ time\ steps} \\
&\quad \text{(extract last-layer pooled output)} \\
\\
&\text{(b) After all passbands:} \\
&\rightarrow \mathrm{Concat\ pooled\ outputs\ from\ 6\ passbands} \\
&\quad + \mathrm{Meta\text{-}features} \\
&\qquad \left(\text{equatorial coordinates,\ host galaxy, redshift,\ etc.}\right) \\
&\rightarrow \mathrm{Linear}(64) \rightarrow \mathrm{BatchNorm} \rightarrow \mathrm{ReLU} \\
&\rightarrow \mathrm{Linear}(64) \rightarrow \mathrm{BatchNorm} \rightarrow \mathrm{ReLU} \\
&\rightarrow \mathrm{Linear}(n_{\mathrm{classes}}) \\
&\rightarrow \mathrm{Softmax}
\end{align*}

(Intermediate hidden states are extracted from each BiGRU layer for calibration analysis.)

\end{enumerate}

\subsection{Dataset}
To evaluate the effectiveness and generalizability of the proposed calibration framework, we conduct experiments on two benchmark datasets spanning distinct domains: SAR image classification and time-domain astronomical event classification (PLAsTiCC dataset).

\begin{enumerate}
\item \textbf{The Moving and Stationary Target Acquisition and Recognition (MSTAR) \cite{b45}:} The MSTAR dataset is a widely used benchmark for SAR image classification. It consists of 3,671 SAR images for training, acquired at a depression angle of 15°, and 3,502 SAR images for testing, acquired at depression angles of 17° and 30° \cite{b45}. Each image depicts military ground vehicles under varying configurations, enabling the study of domain adaptation, limited training, and robust target recognition in SAR applications. The detailed comparison SAR images from MSTAR and optical images of military targets are illustrated in Fig.~\ref{fig:MSTAR_optical_images}.

\begin{figure*}[htb!]
    \centering
    \begin{subfigure}[b]{0.1\textwidth}
        \centering
        \includegraphics[width=\textwidth]{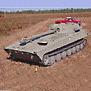}
    \end{subfigure}
    \begin{subfigure}[b]{0.1\textwidth}
        \centering
        \includegraphics[width=\textwidth]{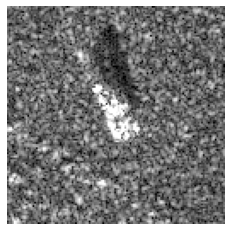}
    \end{subfigure}
    \begin{subfigure}[b]{0.1\textwidth}
        \centering
        \includegraphics[width=\textwidth]{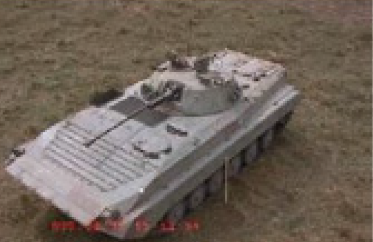}
    \end{subfigure}
    \begin{subfigure}[b]{0.1\textwidth}
        \centering
        \includegraphics[width=\textwidth]{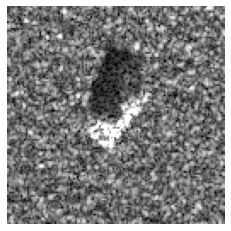}
    \end{subfigure}
    \begin{subfigure}[b]{0.1\textwidth}
        \centering
        \includegraphics[width=\textwidth]{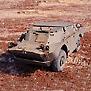}
    \end{subfigure}
    \begin{subfigure}[b]{0.1\textwidth}
        \centering
        \includegraphics[width=\textwidth]{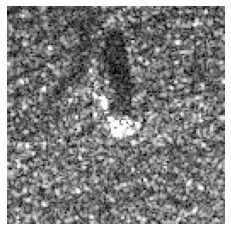}
    \end{subfigure}
    \begin{subfigure}[b]{0.1\textwidth}
        \centering
        \includegraphics[width=\textwidth]{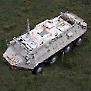}
    \end{subfigure}
    \begin{subfigure}[b]{0.1\textwidth}
        \centering
        \includegraphics[width=\textwidth]{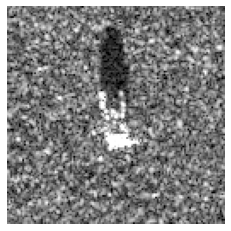}
    \end{subfigure}
    
    \vskip\baselineskip 
    
    \begin{subfigure}[b]{0.1\textwidth}
        \centering
        \includegraphics[width=\textwidth]{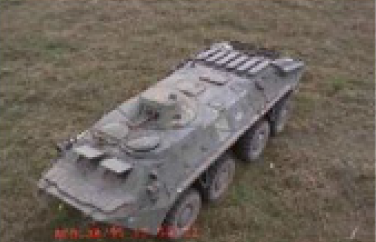}
    \end{subfigure}
    \begin{subfigure}[b]{0.1\textwidth}
        \centering
        \includegraphics[width=\textwidth]{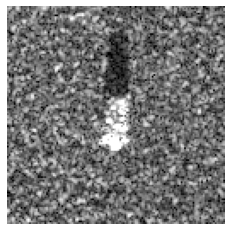}
    \end{subfigure}
    \begin{subfigure}[b]{0.1\textwidth}
        \centering
        \includegraphics[width=\textwidth]{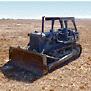}
    \end{subfigure}
    \begin{subfigure}[b]{0.1\textwidth}
        \centering
        \includegraphics[width=\textwidth]{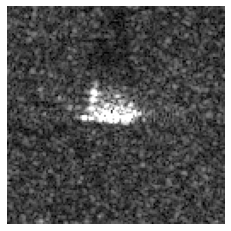}
    \end{subfigure}
    \begin{subfigure}[b]{0.1\textwidth}
        \centering
        \includegraphics[width=\textwidth]{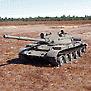}
    \end{subfigure}
    \begin{subfigure}[b]{0.1\textwidth}
        \centering
        \includegraphics[width=\textwidth]{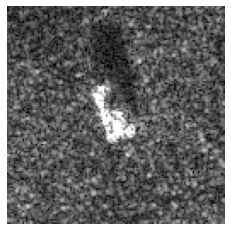}
    \end{subfigure}
    \begin{subfigure}[b]{0.1\textwidth}
        \centering
        \includegraphics[width=\textwidth]{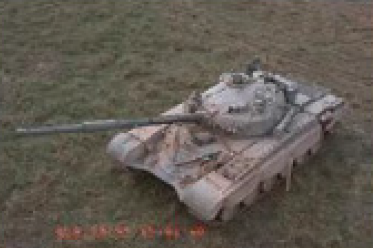}
    \end{subfigure}
    \begin{subfigure}[b]{0.1\textwidth}
        \centering
        \includegraphics[width=\textwidth]{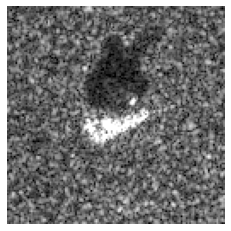}
    \end{subfigure}
    
    \vskip\baselineskip 
    
    \begin{subfigure}[b]{0.1\textwidth}
        \centering
        \includegraphics[width=\textwidth]{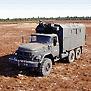}
    \end{subfigure}
    \begin{subfigure}[b]{0.1\textwidth}
        \centering
        \includegraphics[width=\textwidth]{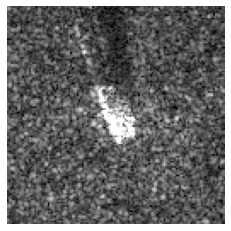}
    \end{subfigure}
    \begin{subfigure}[b]{0.1\textwidth}
        \centering
        \includegraphics[width=\textwidth]{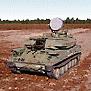}
    \end{subfigure}
    \begin{subfigure}[b]{0.1\textwidth}
        \centering
        \includegraphics[width=\textwidth]{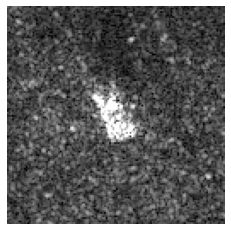}
    \end{subfigure}

    \caption{MSTAR Database: Optical images of military targets versus SAR images \cite{b1}. \\ Optical images of MSTAR database targets and their corresponding SAR images. \textit{Top to bottom rows, From left to right}: the targets are 2S1, BMP2, BRDM2, BTR60, BTR70, D7, T62, T72, ZIL, and ZSU.}
    \label{fig:MSTAR_optical_images}
\end{figure*}

\item \textbf{The Photometric LSST Astronomical Time-series Classification Challenge (PLAsTiCC) \cite{b77, b78}:} The PLAsTiCC dataset is a large-scale \textit{simulated} collection designed to benchmark machine learning approaches to time-domain astronomical classification. The dataset is divided into a training set of 8,000 labeled sources and a test set containing approximately $3.5$x$10^{7}$ unlabeled sources, reflecting the realistic class imbalance and domain shift anticipated in LSST observations. For all experiments with this dataset, we randomly selected 25,000 test samples from the approximately 3.5 million sources in the PLAsTiCC test set.

PLAsTiCC provides multi-band light curves for each object, recorded in six LSST passbands (u, g, r, i, z, y), with associated flux measurements and uncertainties. The accompanying metadata file includes object identifiers, sky coordinates (RA, Dec, Galactic longitude and latitude), Milky Way extinction, photometric and spectroscopic redshifts (for a subset), and host galaxy properties. Each light curve is stored as a time series of observations indexed by Modified Julian Date (MJD), filter, and detection status. Notably, the training set is dominated by nearby, spectroscopically reliable sources, while the test set comprises a larger and more diverse population, including fainter and more distant objects. This separation poses a challenging domain shift for classification and calibration algorithms. Examples of light curve shapes for different target classes across each photometric passband are shown in Fig.~\ref{fig:plasticc_lc_grid}, while the distributions of simulated photometric flux and flux error measurements for each of the six passbands in the training dataset are shown in Fig.~\ref{fig:plasticc_flux_dist} and Fig.~\ref{fig:plasticc_fluxerr_dist}, respectively.

\begin{figure*}[!ht]
    \centering
    \includegraphics[width=0.98\textwidth]{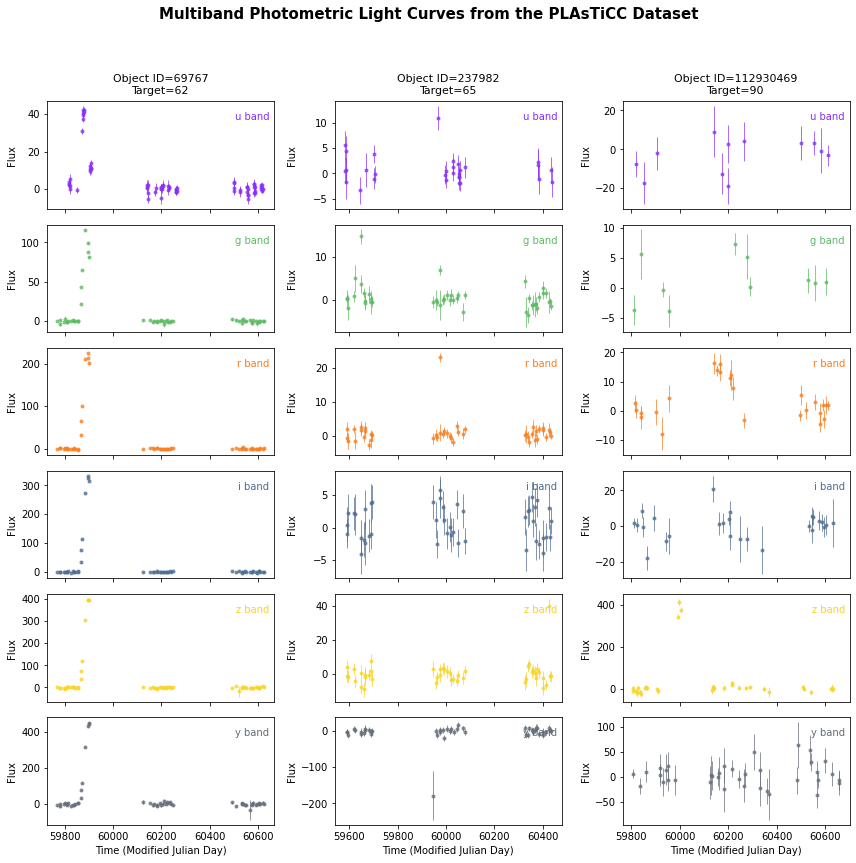}
    \caption{%
        Multiband photometric light curves for three representative PLAsTiCC objects \cite{b77, b78} (shown in columns). 
        Each row corresponds to a photometric band ($u$, $g$, $r$, $i$, $z$, $y$). 
        Error bars indicate photometric uncertainties. The object ID and target class are shown at the top of each column.%
    }
    \label{fig:plasticc_lc_grid}
\end{figure*}

\begin{figure*}[htbp]
    \centering
    \begin{minipage}{0.48\textwidth}
        \centering
        \includegraphics[width=\linewidth]{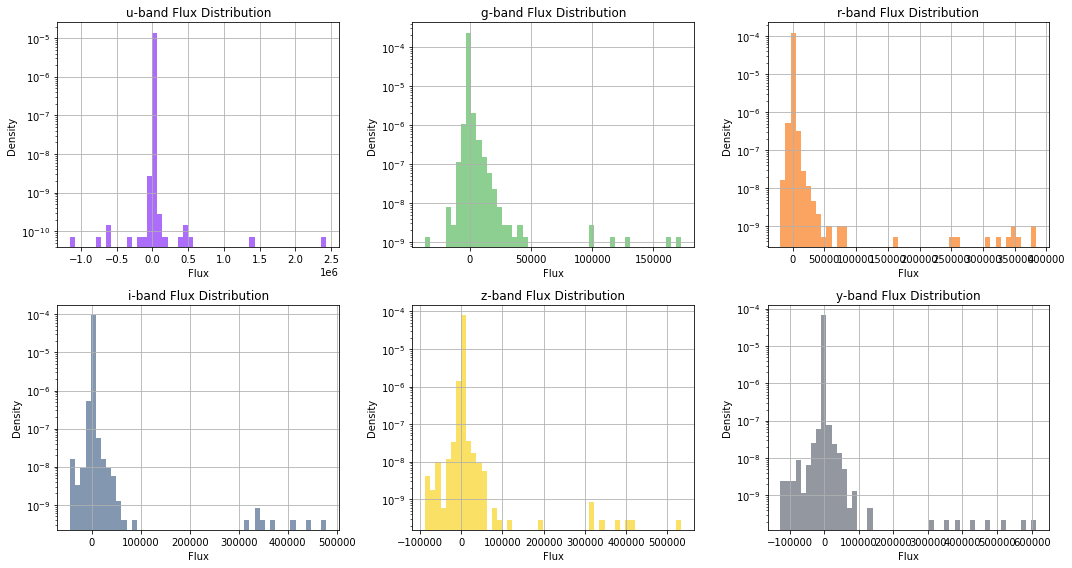}
        \caption{
            Distributions of photometric flux measurements for each of the six PLAsTiCC bands ($u$, $g$, $r$, $i$, $z$, $y$) in the training dataset. Each panel shows the kernel density estimate of flux values, with a logarithmic $y$-axis to highlight the long-tailed, non-Gaussian nature of astronomical photometric measurements.
        }
        \label{fig:plasticc_flux_dist}
    \end{minipage}\hfill
    \begin{minipage}{0.48\textwidth}
        \centering
        \includegraphics[width=\linewidth]{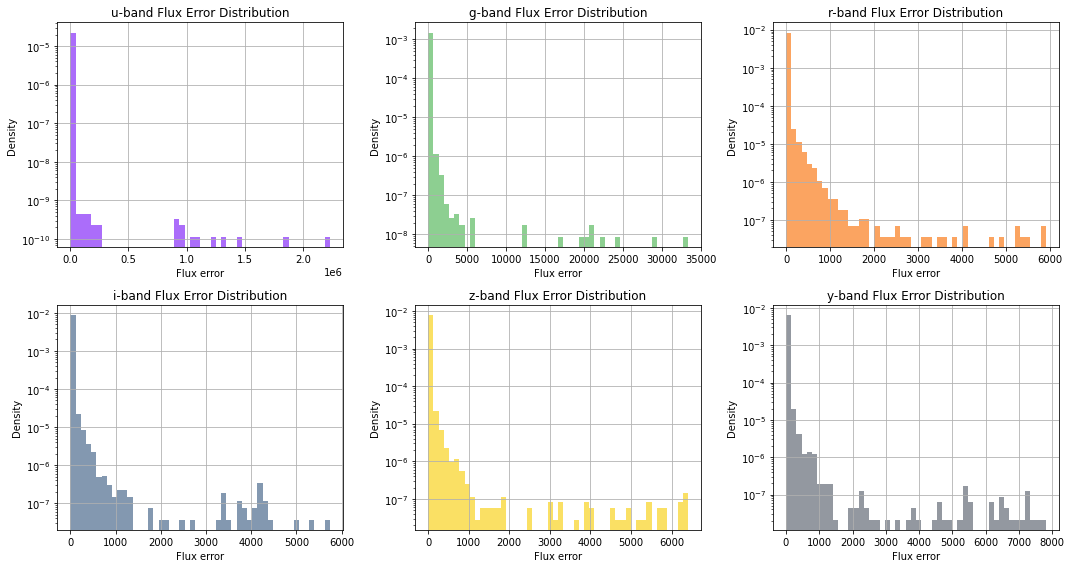}
        \caption{
            Distributions of photometric flux uncertainties for each PLAsTiCC band ($u$, $g$, $r$, $i$, $z$, $y$) in the training set. The panels show kernel density estimates of reported measurement errors, with a logarithmic scale.
        }
        \label{fig:plasticc_fluxerr_dist}
    \end{minipage}
\end{figure*}

\end{enumerate}

\subsection{Calibration Protocol}
Calibration pipelines and their performance are evaluated using a three-stage protocol:

\begin{enumerate}
\item \textbf{Uncalibrated Baseline:} The predictive confidence of each neural network is first evaluated directly using the maximum softmax probability, serving as an uncalibrated reference.

\item \textbf{Single-Layer GP Calibration:} A single-layer Gaussian Process is trained to map individual layer feature representations and their associated softmax scores to calibration residuals, serving as a baseline for comparison with hierarchical methods.

\item \textbf{SAL-GP Calibration:} The proposed SAL-GP calibration framework leverages structured layerwise feature maps from intermediate neural network layers as inputs to a global GP, employing either a hierarchical layer kernel or a reduced multi-layer kernel, as detailed in Sections~\ref{sec:salgp_hierarchical_layer} and~\ref{sec:reduced-multilayer-kernel}.

\end{enumerate}

\subsection{Calibration Metrics}

\subsubsection{Expected Calibration Error (ECE)}
While reliability diagrams provide useful visual insights into model calibration, a scalar summary statistic is often more convenient. Since statistical comparisons between two distributions cannot be fully comprehensive, various calibration metrics have been proposed, each emphasizing different aspects of miscalibration.

One common measure is the ECE \cite{b14, b33}, which quantifies the difference in expectation between confidence and accuracy:

\begin{equation}
    \mathbb{E}_p \left[ |\mathbb{P}(Y = \hat{Y} \mid \mathbb{P} = p) - p| \right].
\end{equation}

ECE approximates this expectation by partitioning predictions into $M$ equally spaced bins—similar to reliability diagrams—and computing a weighted average of the accuracy-confidence difference across bins:

\begin{equation}
    \text{ECE} = \sum_{m=1}^{M} \frac{|B_m|}{n} \left| \text{acc}(B_m) - \text{conf}(B_m) \right|,
\end{equation}

where $n$ is the total number of samples, and $B_m$ denotes the set of predictions falling into bin $m$. The calibration gap for a given bin (illustrated as red bars in reliability diagrams, e.g., Figure~1) is represented by the difference between accuracy and confidence. We use ECE as the primary empirical metric to assess calibration. See Section~S1 for further analysis.

\subsubsection{Maximum Calibration Error (MCE)}
In high-risk applications where precise confidence estimates are crucial, minimizing the worst-case deviation between confidence and accuracy is essential. The MCE \cite{b14, b33} captures this worst-case miscalibration:

\begin{equation}
    \max_{p \in [0,1]} \left| \mathbb{P}(Y = \hat{Y} \mid \mathbb{P} = p) - p \right|.
\end{equation}

Similar to ECE, MCE is estimated via binning:

\begin{equation}
    \text{MCE} = \max_{m \in \{1, \dots, M\}} \left| \text{acc}(B_m) - \text{conf}(B_m) \right|.
\end{equation}

On reliability diagrams (for example, see Fig.~\ref{fig:calibration_mstar_grid}), MCE corresponds to the largest calibration gap (red bars) across all bins, while ECE represents the weighted average of all gaps. A perfectly calibrated classifier has both $\text{MCE} = 0$ and $\text{ECE} = 0$. However, it is important to note that MCE may not be a reliable calibration metric in the presence of untrained or out-of-distribution (OOD) samples, as a single extreme gap can dominate the metric and mask the overall calibration quality of the confidence distribution

\subsubsection{Negative Log-Likelihood (NLL)}
NLL is a standard measure of the quality of probabilistic models \cite{b37}. In deep learning, it is commonly referred to as the cross-entropy loss \cite{b66}. Given a probabilistic model $P(Y \mid X)$ and a dataset of $n$ samples, NLL is defined as:

\begin{equation}
    L = -\sum_{i=1}^{n} \log P(y_i \mid x_i).
\end{equation}

It is well known \cite{b37} that, in expectation, NLL is minimized if and only if $P(Y \mid X)$ perfectly recovers the ground truth conditional distribution.

\subsection{Ablation Studies}
To rigorously isolate the contributions of each architectural and modeling choice in the proposed SAL-GP calibration framework, we conduct the following ablation experiments using the protocols described above. All ablations are evaluated using standard calibration and uncertainty metrics: ECE, MCE, NLL, and potentially including reliability diagrams, as well as classification accuracy. Model hyperparameters (kernel type, learning rate, iteration count) are held fixed across ablation settings.

\begin{enumerate}
\item \textbf{Uncalibrated Baseline:} We first evaluate the uncalibrated confidence estimates produced by the neural network's native softmax outputs. This baseline serves to establish the intrinsic calibration and predictive performance of the backbone models without any post-hoc adjustment, thereby quantifying the extent of miscalibration present in standard deep neural networks.

\item \textbf{Single-Layer GP Calibration:} To benchmark traditional GP calibration, we train a standard GP using only the certain layer's feature representation, after appropriate pooling or flattening, concatenated with the maximum softmax confidence as input. The GP is tasked with regressing the calibration residual, defined as the difference between the network's maximum softmax score and the binary correctness indicator. Formally, the input to the GP is the concatenation of the pooled $l^{th}$ layer's feature vector and the maximum softmax score, i.e., $[\mathbf{f}_{\mathrm{l}},\, p_{\mathrm{max}}]$. This experiment quantifies the calibration improvement attainable through conventional GP-based regression on the most informative deep features, without any notion of hierarchy or task structure. 

\comment{
\item \textbf{Layerwise GP without Task/Layer Kernel:}
Next, we assess the impact of multi-layer input aggregation absent explicit task structure. Here, the GP is trained using concatenated features from all CNN layers, along with the maximum softmax score, but omits the explicit encoding of the layer index in the kernel function. That is, all data points are modeled as originating from a single latent process with a kernel defined on the high-dimensional concatenated input, without a task or layer kernel. This ablation isolates the effect of hierarchical modeling, contrasting it with a naïve, high-dimensional input approach that ignores the semantic distinction between features at different layers.}

\item \textbf{SAL-GP Variant with Hierarchical Layer Kernel:}
This variant models the calibration residual using a GP with a hierarchical layerwise kernel, expressed as $k_{\mathrm{global}} + \delta_{ll'}\, k_{\mathrm{layer}}$. Each input consists of pooled feature representation, maximum softmax confidence, and layer index, i.e., $[\mathbf{f}_l,\, p_{\mathrm{max}},\, l]$. The global component $k_{\mathrm{global}}$ captures layer-agnostic similarity, while the local component $k_{\mathrm{layer}}$ (weighted by the Kronecker delta $\delta_{ll'}$) captures within-layer structure.

\item \textbf{SAL-GP with Structured Multi-Layer Kernel (Reduced ICM Approximation, Proposed):}
Our primary experiment implements a structured multi-layer GP calibration model, using a reduced form of the ICM as an additive kernel approximation. In this framework, each neural network layer is treated as a distinct layer (task) in a multi-index GP, with the input at each layer $l$ consisting of the pooled feature representation, the maximum softmax confidence, and the corresponding layer index, i.e., $[\mathbf{f}l, p{\mathrm{max}},, l]$. The multi-layer kernel is constructed as an additive combination of a kernel over the feature vector, a kernel over the confidence score, and a discrete index kernel over the layer identity. This reduced ICM approximation discards direct cross-layer (off-diagonal) covariance terms, yielding a computationally efficient and interpretable multi-layer GP model. The additive structure enables the model to capture within-layer (local) calibration behavior and aggregate global effects across all layers, while allowing coherent uncertainty quantification through joint marginalization.

\item \textbf{Backbone Network Ablation:}
To verify architectural generality, we replicate all calibration experiments across several widely used convolutional neural network backbones: a standard sequential CNN (ConvNet) \cite{b76}, a fully convolutional network without dense layers (AConvNet) \cite{b49}, and a deep residual network (ResNet-18) \cite{b73, b74, b75}. This suite of architectures allows us to assess the robustness of SAL-GP calibration under varying depth, connectivity, and representational complexity.

\item \textbf{Dataset Generalization:}
Experiments are conducted on both the MSTAR SAR image dataset \cite{b45} and the PLAsTiCC astronomical time-series dataset \cite{b77, b78}. This dual-domain evaluation ensures that our findings on calibration quality and uncertainty estimation are robust to significant shifts in data modality, class distribution, and domain-specific challenges.

\item \textbf{Pooling and Feature Aggregation:}
Where applicable, we further analyze the sensitivity of calibration results to the method of feature aggregation. We compare the effect of using max pooling, average pooling when extracting representations from intermediate feature maps prior to GP calibration. This ablation confirms that observed improvements are not artifacts of specific pooling choices.

\item \textbf{Uncertainty Propagation Comparison:}
Finally, we compare the predictive variance estimates produced by the various calibration methods, including the single-layer GP, the hierarchical GP, and multi-layer GP. By analyzing uncertainty propagation through the network hierarchy, we demonstrate that SAL-GP calibration yields more reliable and interpretable uncertainty quantification, which is especially critical for risk-sensitive and out-of-distribution detection scenarios.

\end{enumerate}

\subsection{Deep Neural Network Classifiers Training Setup}
For ConvNet and AConvNet, all models were trained for 100 epochs with a batch size of 32, an initial learning rate of $5 \times 10^{-4}$, and 10 output classes, using the Adam optimizer and a step learning rate scheduler (step size 100, gamma 0.1). For ResNet-18, the training configuration was identical except that models were trained for 50 epochs. All classifiers were trained with the cross-entropy loss function. Unless otherwise specified, these settings were consistently applied across all experiments to ensure fair comparison.

\subsection{Gaussian Process Training Setup}
All experiments were conducted on a workstation equipped with an Intel Core i7-14700KF CPU (20 cores, 28 threads, 3.40~GHz base frequency), 64~GB DDR5 RAM (5600~MT/s), and an NVIDIA GeForce RTX 3060 GPU with 12~GB VRAM. The system ran Microsoft Windows 11 Education. Model training and evaluation were performed using PyTorch 2.4.1 with CUDA 12.1 support. Unless otherwise specified, all GP models were trained using a Matérn kernel with $\nu = 2.5$. For the MSTAR dataset \cite{b45}, models were trained for 2000 iterations with a learning rate of $5 \times 10^{-3}$. For the PLAsTiCC dataset \cite{b77, b78}, models were trained for 2500 iterations with a learning rate of $2 \times 10^{-3}$.

\section{Result}

Across datasets and network architectures, we present calibration and accuracy results with an ablation study. All GP frameworks provide predictive uncertainty (variance) for calibration, as shown in the tables, whereas uncalibrated and temperature scaling methods do not quantify uncertainty since they are not Bayesian approaches.

Our analysis focuses on the stability, robustness, and improvement of calibration performance relative to the uncalibrated and temperature-scaled baselines. For single-layer GP calibration, large variations in performance across layers are regarded as indicative of instability and unpredictability in the absence of prior knowledge regarding optimal layer selection.

For clarity, we denote the SAL-GP variant with a multi-layer structured kernel as SAL-GP(ML), and the variant with a hierarchical layerwise kernel as SAL-GP(HL). For both the multi-task and hierarchical SAL-GP frameworks, we report calibration performance for joint-likelihood-optimized local layers as a reference, although the global prediction serves as the primary result for all local GPs.

ECE is selected as the primary metric for comparing the best-performing single-layer GP with the proposed SAL-GP framework, as it directly quantifies the average discrepancy between predicted probabilities and observed frequencies \cite{b14, b33}. ECE is widely used in the literature and provides an intuitive assessment of model miscalibration. In contrast, the MCE is less stable and more sensitive to binning and outliers, limiting its reliability. NLL, while a proper scoring rule important for overall probabilistic assessment, conflates calibration with predictive sharpness and accuracy, and is therefore not a pure calibration metric. Similarly, the Brier score combines calibration with aspects of refinement and discrimination. ECE is the most widely reported calibration metric in the literature \cite{b14, b81}, and specifically targets the reliability of probabilistic predictions, which is the central concern in model calibration.

\subsection{Calibration and Accuracy with Ablation Study: MSTAR Databse}

\subsubsection{Standard Convolutional Network}

\begin{table*}[htbp]
\centering
\caption{\textsc{Main Calibration and Accuracy Metrics: Standard Convolutional Network (ConvNet)}}
\label{tab:calibration_metrics_convnet}  
\begin{threeparttable}
\begin{tabular}{ll ccccc}
\hline\hline
\multicolumn{7}{c}{\textbf{Calibration Performance Metrics: Standard ConvNet}} \\ \hline
\textbf{Pooling} & \textbf{Method} & \textbf{ECE$\downarrow$} & \textbf{MCE$\downarrow$} & \textbf{NLL$\downarrow$} & \textbf{Brier$\downarrow$} & \textbf{Var.} \\ \hline
- & Uncalibrated (Baseline)      & 0.01054 & 0.66202 & 0.07991 & 0.03289 & - \\
- & Temperature Scaled (Baseline) & 0.02088 & 0.14369 & 0.51153 & 0.04170 & - \\
Max & Single GP (Layer 1)          & 0.00913 & 0.65807 & 0.08016 & 0.03284 & 0.00016 \\
Avg & Single GP (Layer 1)          & 0.01044 & 0.66626 & 0.08006 & 0.03285 & 0.00010 \\
Max & Single GP (Layer 2)          & 0.00980 & 0.65047 & 0.08009 & 0.03286 & 0.00019 \\
Avg & Single GP (Layer 2)          & 0.00877 & 0.60842 & 0.08032 & 0.03285 & 0.00011 \\
Max & Single GP (Layer 3)          & 0.00894 & 0.65612 & 0.08028 & 0.03287 & 0.00020 \\
Avg & Single GP (Layer 3)          & 0.00900 & 0.66271 & 0.08041 & 0.03292 & 0.00009 \\
Max & Single GP (Layer 4)          & 0.00915 & 0.66582 & 0.08049 & 0.03283 & 0.00019 \\
Avg & Single GP (Layer 4)          & 0.00832 & 0.65491 & 0.08047 & 0.03287 & 0.00010 \\
- & Single GP (Layer 5)          & 0.01005 & 0.65681 & 0.07958 & 0.03287 & 0.00023 \\
Max & SAL-GP (ML) G        & 0.01359 & 0.05553 & 0.07599 & 0.03371 & 0.00009 \\
Max & SAL-GP (ML) L1        & 0.01359 & 0.05550 & 0.07599 & 0.03371 & 0.00008 \\
Max & SAL-GP (ML) L2        & 0.01358 & 0.05550 & 0.07600 & 0.03371 & 0.00011 \\
Max & SAL-GP (ML) L3        & 0.01359 & 0.05551 & 0.07600 & 0.03371 & 0.00009 \\
Max & SAL-GP (ML) L4        & 0.01360 & 0.05554 & 0.07598 & 0.03371 & 0.00005 \\
Max & SAL-GP (ML) L5        & 0.01359 & 0.05549 & 0.07599 & 0.03371 & 0.00024 \\
Avg & SAL-GP (ML) G         & 0.00797 & 0.20182 & 0.07453 & 0.03372 & 0.00015 \\
Avg & SAL-GP (ML) L1        & 0.00797 & 0.19893 & 0.07453 & 0.03372 & 0.00002 \\
Avg & SAL-GP (ML) L2        & 0.00797 & 0.19893 & 0.07453 & 0.03372 & 0.00001 \\
Avg & SAL-GP (ML) L3        & 0.00797 & 0.20036 & 0.07453 & 0.03372 & 0.00002 \\
Avg & SAL-GP (ML) L4        & 0.00797 & 0.20182 & 0.07453 & 0.03372 & 0.00000480 \\
Avg & SAL-GP (ML) L5        & 0.00797 & 0.20182 & 0.07453 & 0.03372 & 0.00025 \\
Max & SAL-GP (HL) G  & 0.01651 & 0.71947 & 0.07596 & 0.03346 & 0.00025 \\
Max & SAL-GP (HL) L1  & 0.01728 & 0.76691 & 0.07553 & 0.03365 & 0.0001 \\
Max & SAL-GP (HL) L2  & 0.01669 & 0.72093 & 0.07576 & 0.03361 & 0.00018 \\
Max & SAL-GP (HL) L3  & 0.01727 & 0.73021 & 0.07552 & 0.03363 & 0.00014 \\
Max & SAL-GP (HL) L4  & 0.01738 & 0.73243 & 0.07541 & 0.03370 & 0.00007 \\
Max & SAL-GP (HL) L5  & 0.00955 & 0.63518 & 0.07883 & 0.03291 & 0.00051 \\
Avg & SAL-GP (HL) G  & 0.01168 & 0.60836 & 0.07820 & 0.03303 & 0.00044 \\
Avg & SAL-GP (HL) L1  & 0.01457 & 0.61912 & 0.07529 & 0.03376 & 0.00005 \\
Avg & SAL-GP (HL) L2  & 0.01413 & 0.62375 & 0.07517 & 0.03376 & 0.00004 \\
Avg & SAL-GP (HL) L3  & 0.01425 & 0.59463 & 0.07511 & 0.03377 & 0.00002 \\
Avg & SAL-GP (HL) L4  & 0.01364 & 0.56681 & 0.07510 & 0.03377 & 0.00001 \\
Avg & SAL-GP (HL) L5  & 0.00941 & 0.65855 & 0.07940 & 0.03287 & 0.00068 \\ \hline
\textbf{} & \textbf{Accuracy(Train, Validation, Test)}      & \multicolumn{5}{c}{100\%, 100\%, 97.92\%} \\ \hline
\end{tabular}
\begin{tablenotes}
\footnotesize
\item Calibration metrics: ECE (Expected Calibration Error), MCE (Maximum Calibration Error), NLL (Negative Log-Likelihood), Brier score, and mean predictive variance (Var.).
\item Accuracy is reported for the original neural network and is unchanged across calibration methods.
\item For "Single GP," each row corresponds to a GP trained using only the $l$-th layer's feature map plus softmax as input.
\item "SAL-GP (ML)" uses all layers as input with the structured multi-layer index kernel.
\item "SAL-GP (HL)" uses all layers as input with the hierarchical layer kernel.
\item Optimized T = 0.0308
\end{tablenotes}
\end{threeparttable}
\end{table*}
Table~\ref{tab:calibration_metrics_convnet} presents the main calibration and accuracy metrics for the standard convolutional network, including predictive variance and the results of ablation studies. The standard convolutional network (ConvNet) baseline is already well-calibrated, yielding a low ECE of 0.01054 and a Brier score of 0.03289. This indicates that, for this dataset and architecture, the uncalibrated network produces confidence scores that closely match empirical correctness, suggesting limited room for further improvement through post-hoc calibration.

Applying TS actually degrades calibration in this setting, as indicated by the increased ECE (0.02088) and substantially higher NLL. This failure is due to the validation set achieving perfect accuracy, which causes the NLL loss,
\begin{equation}
L(T) = -\frac{z_{y}}{T} + \log \left( \sum_{j=1}^{K} \exp(z_j / T) \right),
\end{equation}
(where $z_y$ is the logit for the true class $y$, $z_j$ is the logit for class $j$, and $T$ is the temperature parameter) to become nearly flat with respect to $T$. 

The gradient of the loss with respect to temperature is
\begin{equation}
\frac{dL}{dT} = \frac{1}{T^2} \left[z_{y} - \mathbb{E}_{p(T)}[z] \right],
\end{equation}
where $\mathbb{E}_{p(T)}[z]$ denotes the expected logit under the temperature-scaled softmax, i.e., $\mathbb{E}_{p(T)}[z] = \sum_{j=1}^K p_j(T) z_j$, with $p_j(T)$ the softmax probability for class $j$ at temperature $T$.

When all predictions are correct and highly confident ($z_y = 0$, $z_{j \neq y} \ll 0$), this gradient remains small and positive, so the optimizer keeps reducing $T$, making the softmax outputs even more overconfident. As a result, calibration actually worsens—a well-known pathology of TS when validation accuracy is perfect \cite{b14, b69, b82}.

Applying single-layer GP calibration yields a clear, nontrivial improvement over the uncalibrated baseline. Although the absolute reduction in ECE, from 0.01054 (uncalibrated) to 0.00832 (Layer 4, average pooling) and 0.00894 (Layer 3, max pooling), may appear numerically modest, it is significant given the already well-calibrated state of the network. Achieving further ECE reduction at such low baseline error rates is mathematically challenging, so the relative improvement should not be understated.

Notably, however, the calibration quality from single-layer GPs varies noticeably across both layer choice and pooling method: for average pooling, ECEs across layers range from 0.00832 to 0.01044, and for max pooling from 0.00894 to 0.01005. This inter-layer variability, though quantitatively small, introduces practical uncertainty because the optimal layer is only observable post hoc with access to test labels, which are unavailable in deployment. Consequently, relying on a single, fixed layer chosen from one dataset domain under specific experimental conditions introduces calibration risk, as selecting a suboptimal layer may result in unpredictable or even worse performance when applied to different environments.

The proposed SAL-GP(ML) model achieves the best overall calibration under average pooling. Its ECE reaches 0.00797 (global prediction, average pooling), while also significantly addressing the intrinsic issue of high MCE in GP-based calibration that single-layer GPs failed to resolve (reducing MCE from 0.66202 to 0.20182). This model outperforms all single-layer GP and baseline configurations. Importantly, this improvement is not a result of simply averaging or taking the median of local layerwise GP outputs. Instead, the SAL-GP framework fuses information from all internal layers by constructing a joint likelihood that fully couples their predictions. In this Bayesian formulation, the global mean and variance are determined by maximizing the posterior over all local GPs, leading to a calibrated prediction that genuinely reflects uncertainty propagation across the entire network rather than just aggregating independent sources.

Mathematically, the global mean prediction is
\begin{equation}
\begin{split}
p(f_{\text{global}} \mid \mathbf{X}, \mathbf{y}) = 
    \int \left[ \prod_{\ell=1}^L p(y_\ell \mid f_\ell)\, p(f_\ell \mid f_{\text{global}}) \right] \\
    \times\, p(f_{\text{global}})\, d f_{1:L}
\end{split}
\end{equation}
where $f_\ell$ denotes the local GP at layer $\ell$, $f_{\text{global}}$ is the shared latent function, and $L$ is the number of layers. This structure enables the optimal pooling of predictive uncertainty, resulting in a single Bayesian posterior that leverages all available information.

Crucially, the SAL-GP(ML) demonstrates remarkable stability: global ECE values remain tightly grouped (0.00797–0.0136 across pooling and layers) with consistently low predictive variance. This robustness directly mitigates the layer-selection sensitivity inherent in single-layer GP calibration. Even within each pooling strategy, the global SAL-GP prediction offers modestly superior and more consistent calibration than local (layerwise) outputs, in line with the theoretical benefit of layerwise Bayesian modeling.

Another variant of SAL-GP(HL) model does not outperform single-layer GP calibration or the uncalibrated baseline in this setting. Specifically, global predictions from SAL-GP(HL) yield higher ECE values (0.01651 with max pooling and 0.01168 with average pooling) compared to both the best single-layer GP and the uncalibrated network. Additionally, the calibration metrics for local GPs within this specific variant of SAL-GP framework exhibit greater variability across layers, and only the Layer 5 local prediction (ECE of 0.00941 with average pooling) approaches the performance of single-layer GP calibration. This indicates that, unlike the SAL-GP(ML), hierarchical layer kernel does not provide consistently robust or superior calibration across layers and pooling strategies for this ConvNet architecture. In this context, the benefit of this variant is limited, and its performance is unstable with respect to layer selection.

When comparing pooling strategies, average pooling consistently yields moderately lower ECEs than max pooling across all models, suggesting that spatial averaging before calibration can further regularize confidence scores and improve reliability. However, this effect is relatively minor in single GP models within the present architecture due to the already well-calibrated confidence.

\begin{figure*}[t]
    \centering
    \hspace*{0.19\textwidth}
    \hspace*{0.19\textwidth}
    \includegraphics[width=0.19\textwidth]{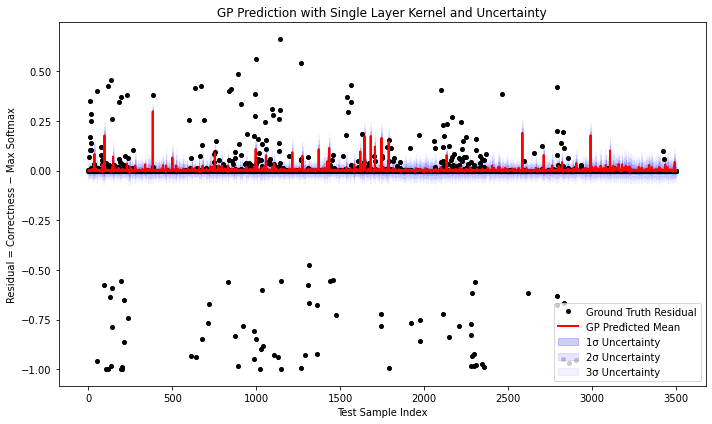}
    \includegraphics[width=0.19\textwidth]{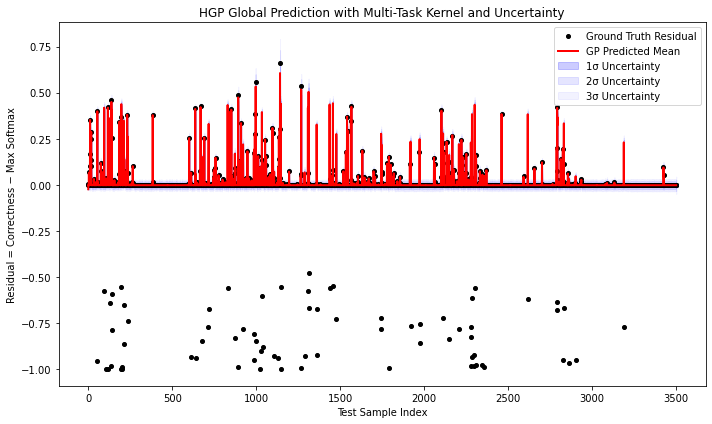}
    \includegraphics[width=0.19\textwidth]{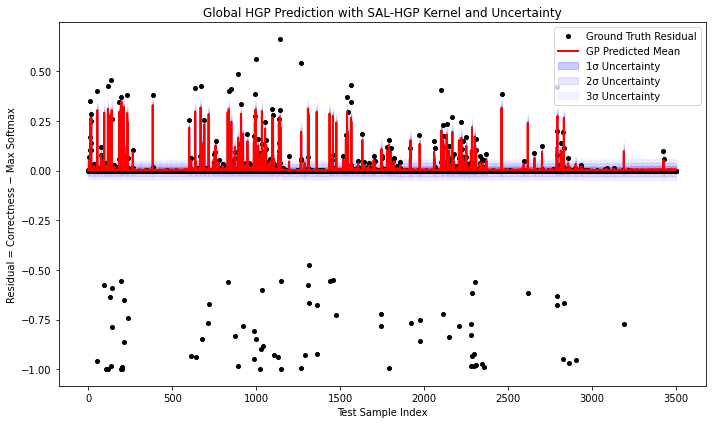} \\[2pt]
    \includegraphics[width=0.19\textwidth]{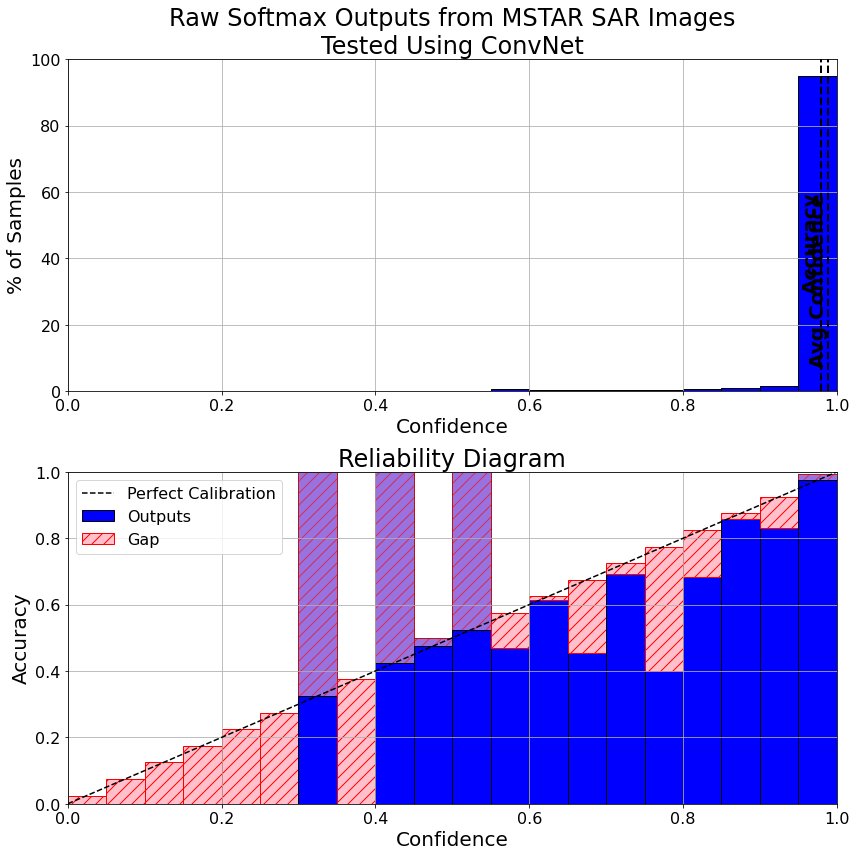}
    \includegraphics[width=0.19\textwidth]{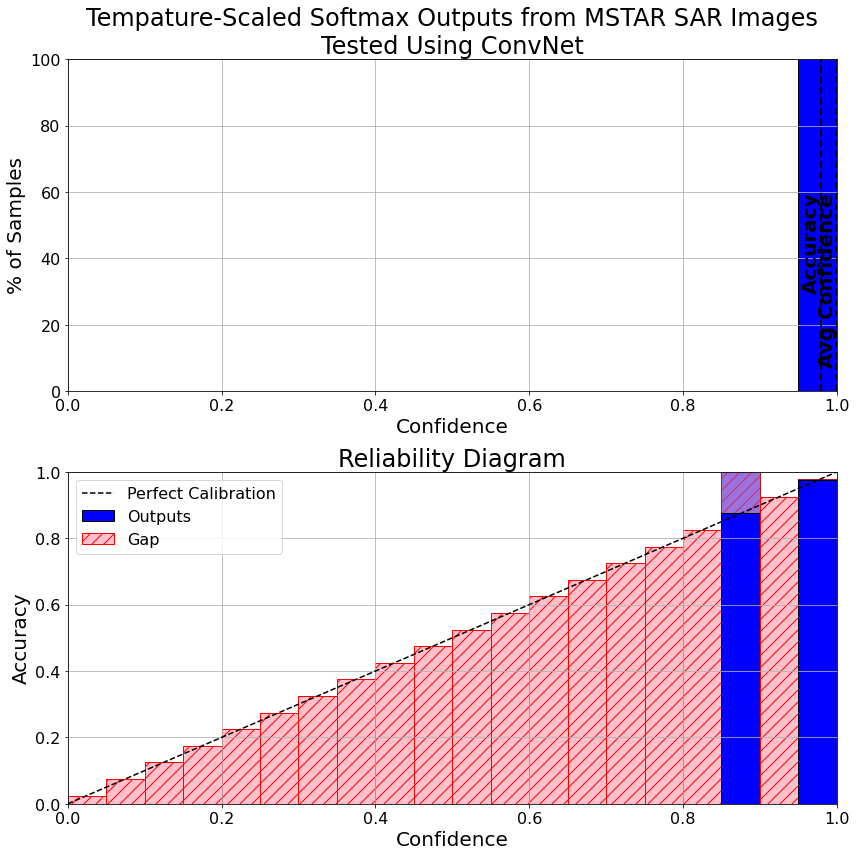}
    \includegraphics[width=0.19\textwidth]{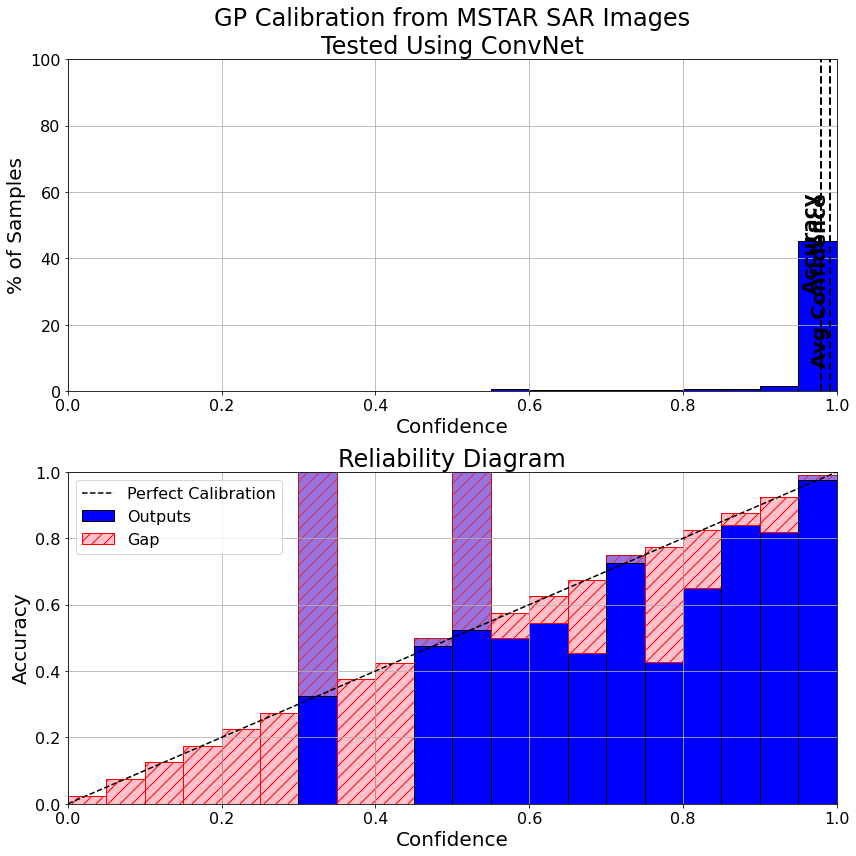}
    \includegraphics[width=0.19\textwidth]{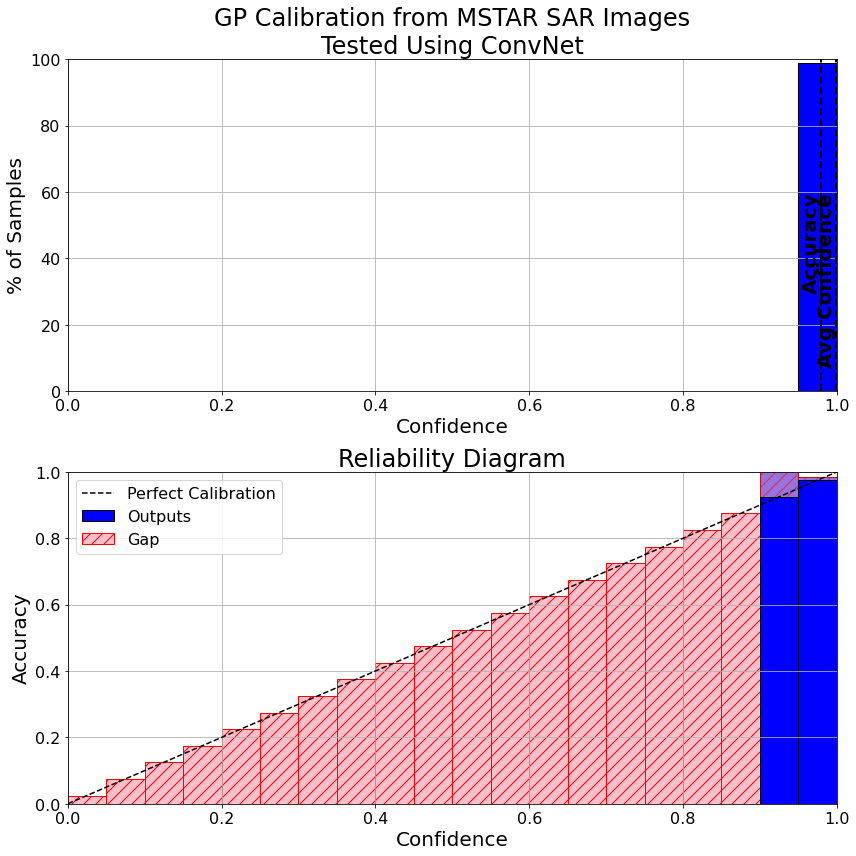}
    \includegraphics[width=0.19\textwidth]{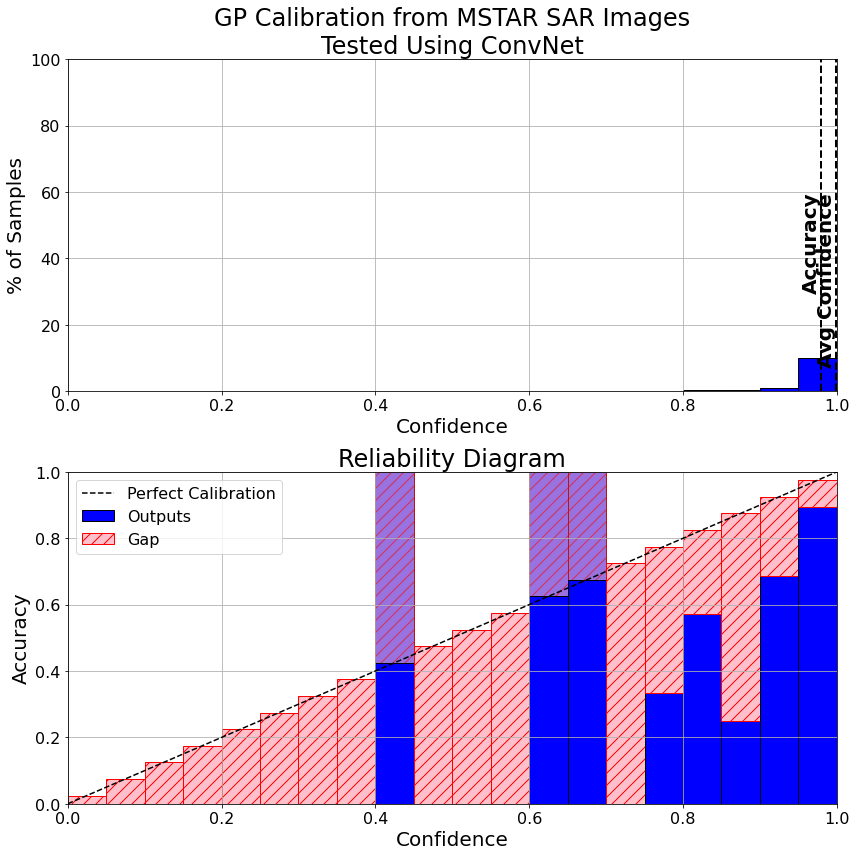} \\

    \begin{minipage}{0.19\textwidth}\centering Uncalibrated\end{minipage}
    \begin{minipage}{0.19\textwidth}\centering Temp. Scaled\end{minipage}
    \begin{minipage}{0.19\textwidth}\centering Single GP (Best Layer)\end{minipage}
    \begin{minipage}{0.19\textwidth}\centering SAL-GP (ML)\end{minipage}
    \begin{minipage}{0.19\textwidth}\centering SAL-GP (HL)\end{minipage}

    \caption{Calibration residual fit plots (top row) and reliability diagrams (bottom row) for each calibration approach on the standard convolutional neural network, evaluated on the MSTAR dataset with max pooling. The top row displays only the residual fit plots for GP-based methods. For uncalibrated and temperature scaled columns, no residual fit is shown, since these methods do not provide predictive uncertainty estimates. Each column corresponds to a different calibration method: uncalibrated, temperature scaled, single-layer GP (best-performing layer), SAL-GP (multi-layer), and proposed SAL-GP (hierarchical layer kernel).}
    \label{fig:calibration_comparison_convnet_mstarWithMaxPooling}
\end{figure*}

\begin{figure*}[t]
    \centering
    \hspace*{0.19\textwidth}
    \hspace*{0.19\textwidth}
    \includegraphics[width=0.19\textwidth]{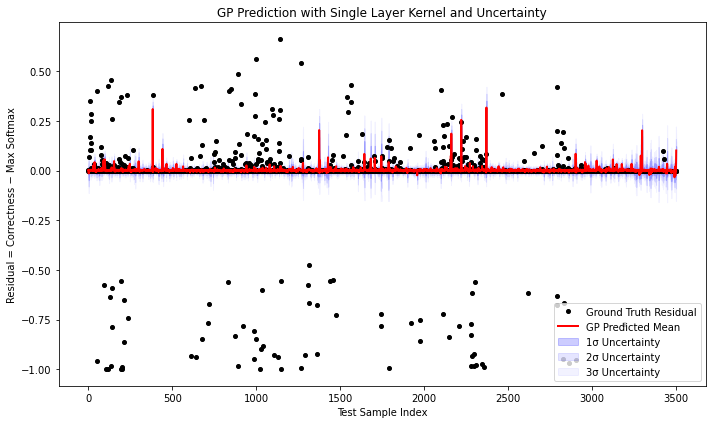}
    \includegraphics[width=0.19\textwidth]{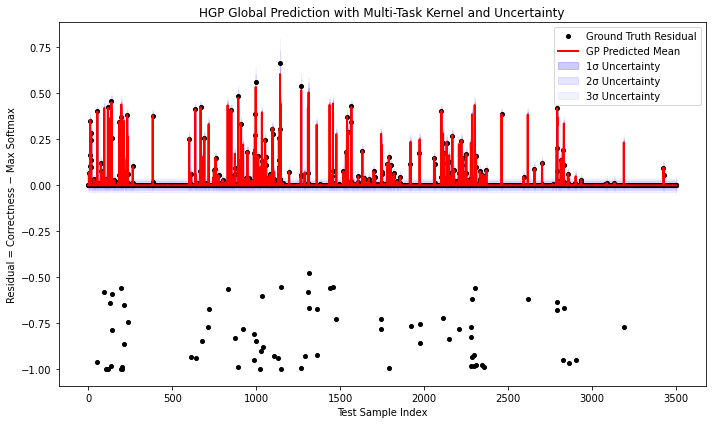}
    \includegraphics[width=0.19\textwidth]{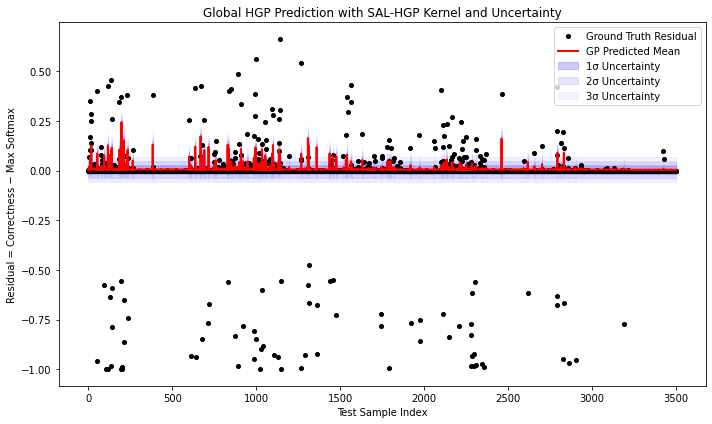} \\[2pt]
    \includegraphics[width=0.19\textwidth]{ConvNet_Uncalibrated_RD_MSTAR.png}
    \includegraphics[width=0.19\textwidth]{ConvNet_Temperature_Scaled_RD_MSTAR.png}
    \includegraphics[width=0.19\textwidth]{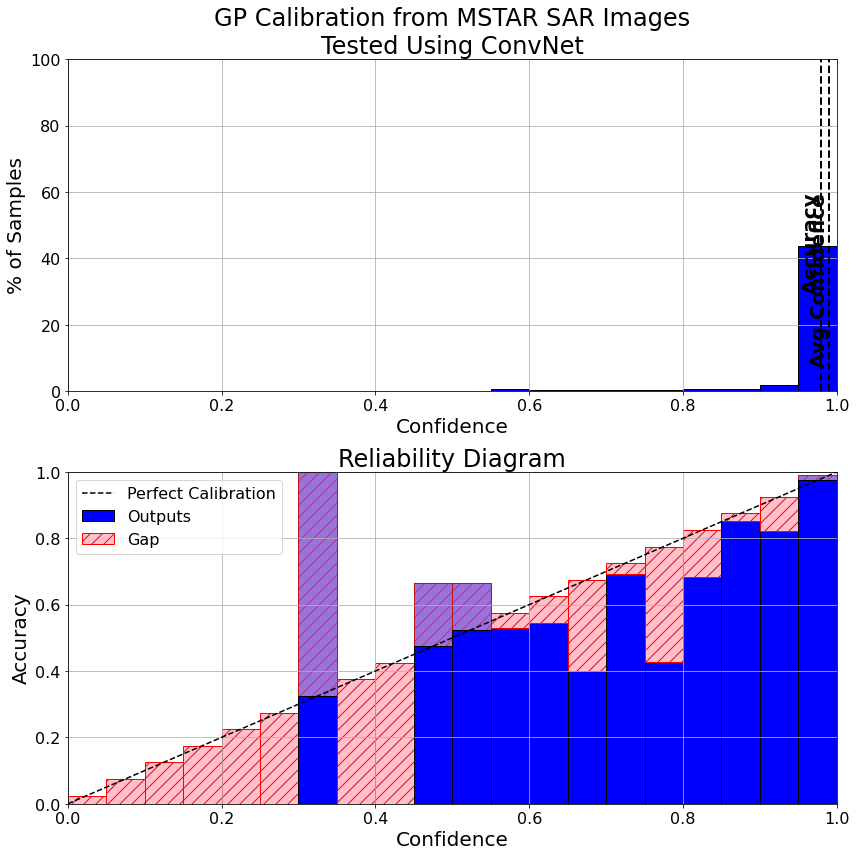}
    \includegraphics[width=0.19\textwidth]{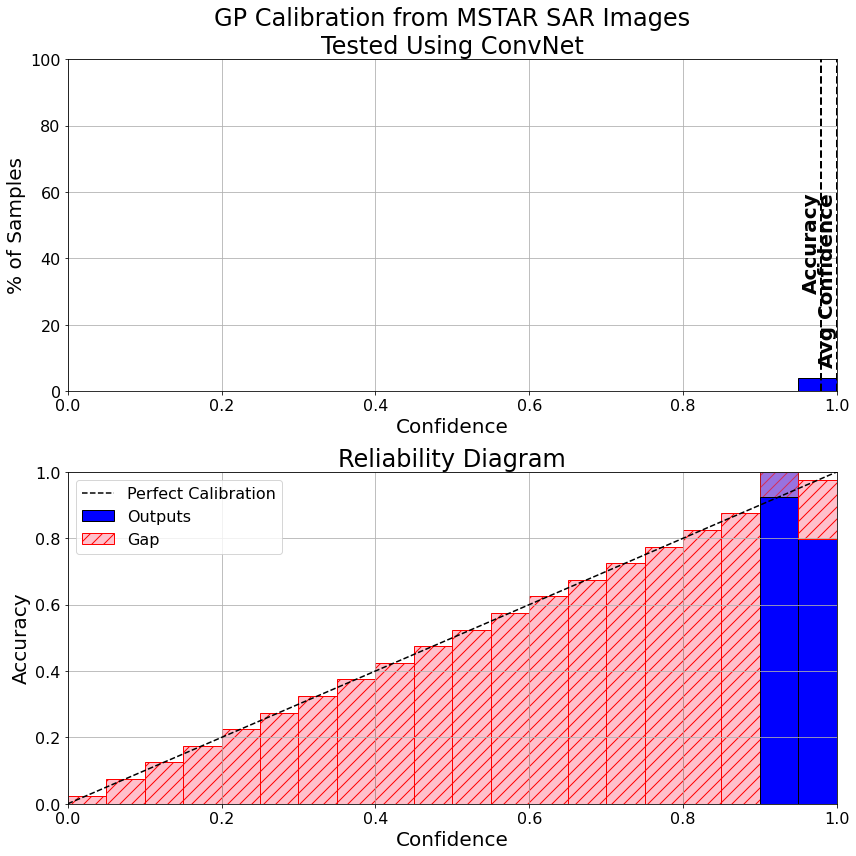}
    \includegraphics[width=0.19\textwidth]{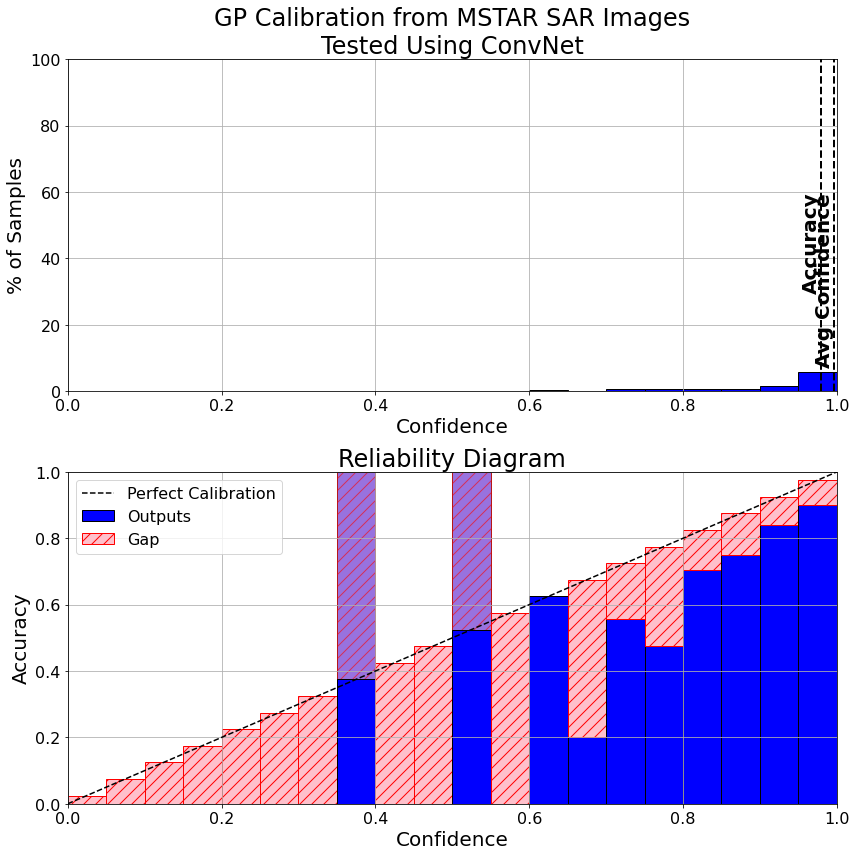} \\

    \begin{minipage}{0.19\textwidth}\centering Uncalibrated\end{minipage}
    \begin{minipage}{0.19\textwidth}\centering Temp. Scaled\end{minipage}
    \begin{minipage}{0.19\textwidth}\centering Single GP (Best Layer)\end{minipage}
    \begin{minipage}{0.19\textwidth}\centering SAL-GP (ML)\end{minipage}
    \begin{minipage}{0.19\textwidth}\centering SAL-GP (HL)\end{minipage}

    \caption{Calibration residual fit plots (top row) and reliability diagrams (bottom row) for each calibration approach on the standard convolutional neural network, evaluated on the MSTAR dataset with average pooling. The top row displays only the residual fit plots for GP-based methods. For uncalibrated and temperature scaled columns, no residual fit is shown, since these methods do not provide predictive uncertainty estimates. Each column corresponds to a different calibration method: uncalibrated, temperature scaled, single-layer GP (best-performing layer), SAL-GP (multi-layer), and proposed SAL-GP (hierarchical layer kernel).}
    \label{fig:calibration_comparison_convnet_mstarWithAvgPooling}
\end{figure*}

Inspection of the residual fit plots provides additional evidence supporting the superiority of the layerwise kernel approaches. In particular, the SAL-GP(ML) and SAL-GP(HL) models accurately capture positive residuals, that is, samples where the true label is correct but the network is underconfident, which none of the single-layer GP models are able to fit effectively. This improved modeling of the positive residual regime is crucial for correcting systematic underconfidence under very high accuracy and already well-calibrated neural network outputs, and for achieving sharp, reliable calibration, as is directly observable in Fig.~\ref{fig:calibration_comparison_convnet_mstarWithMaxPooling} and Fig.~\ref{fig:calibration_comparison_convnet_mstarWithAvgPooling} Both hierarchical models, however, still show some limitations in capturing negative residuals, although this effect is less consequential for real-world risk calibration in models with very high test accuracy.

In summary, although the ConvNet baseline is already well-calibrated, the proposed SAL-GP(ML) framework delivers meaningful and, more importantly, robust improvements in calibration. SAL-GP(ML) achieves stable performance across layers and pooling methods, minimizing the risk associated with layer selection—a key practical advantage over single-layer GP approaches. The global prediction from SAL-GP(ML) is consistently as good as or better than local predictions, and offers reliable uncertainty estimates that are particularly valuable in safety-critical and risk-aware applications. In contrast, the SAL-GP(HL) model does not consistently improve upon the baseline or single-layer GP calibration, with global and most local predictions underperforming except for Layer 5 under average pooling.

\subsubsection{AConvNet Network}
Table~\ref{tab:calibration_metrics_aconvnet} presents the main calibration and accuracy metrics for the standard all-convolutional network, including predictive variance and the results of ablation studies. The uncalibrated AConvNet baseline exhibits substantial miscalibration, with a notably high ECE of 0.18257 and a Brier score of 0.40602. During training, the test accuracy one epoch before the final iteration was 93.29\%, but it suddenly dropped to 76.44\% at the end. Nevertheless, actual test accuracy remains unknown until the test labels are revealed, while only near-perfect training and validation accuracy are available. This situation reflects a real-world deployment scenario. These results indicate that, unlike the standard ConvNet, the AConvNet architecture is significantly overconfident and unreliable out of the box, leaving considerable room for improvement through calibration methods.

Temperature scaling provides only a minor improvement, reducing the ECE to 0.16855. However, the reductions in NLL and Brier scores are noteworthy compared to other models. However, the absolute calibration remains poor. This limited effect is consistent with the theoretical expectation for temperature scaling in high-entropy, miscalibrated settings, and is visualized by the reliability diagrams in Figs. 6 and 7, which show persistent calibration gaps across confidence bins.

Single-layer GP calibration results in a substantial improvement over both the uncalibrated and temperature-scaled models. The best ECEs are achieved at Layer 2 with average pooling (0.08800) and at Layer 1 with max pooling (0.09335), while the worst ECEs are observed at Layer 5 with average pooling (0.12613) and max pooling (0.14431). This represents a nearly two times reduction in calibration error compared to the baseline. The improvement is clearly observable in both the reliability diagrams and residual fit plots, where the GP model more accurately aligns predicted confidence with empirical accuracy.

\begin{figure*}[t]
    \centering
    \hspace*{0.19\textwidth}
    \hspace*{0.19\textwidth}
    \includegraphics[width=0.19\textwidth]{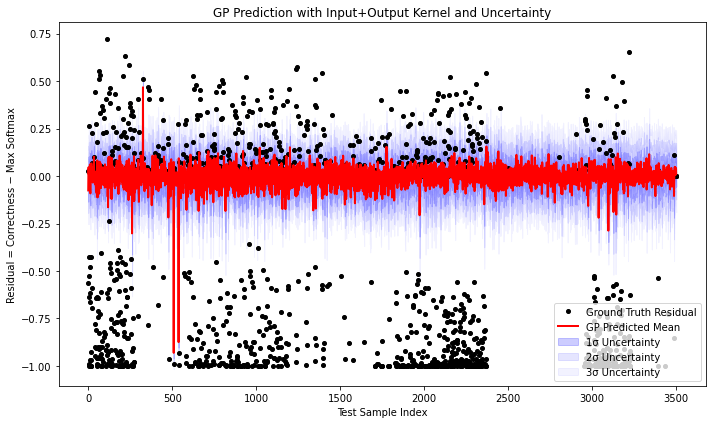}
    \includegraphics[width=0.19\textwidth]{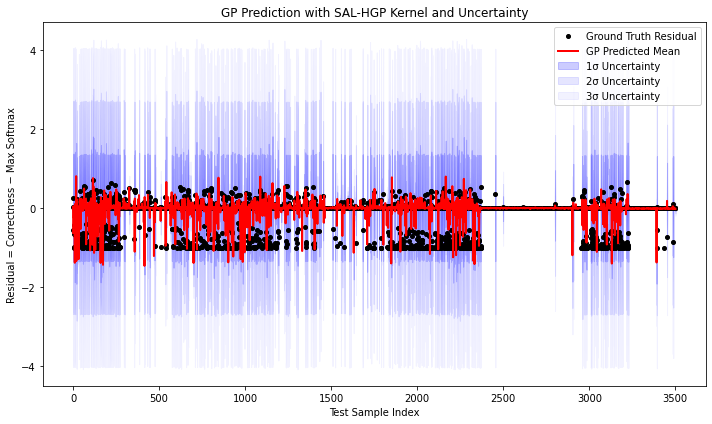}
    \includegraphics[width=0.19\textwidth]{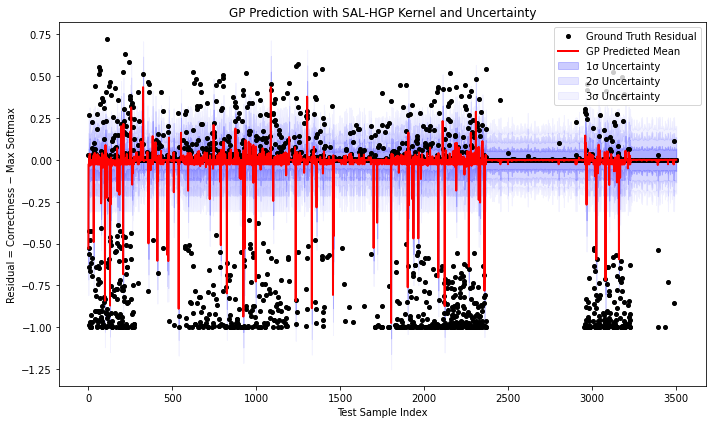} \\[2pt]
    \includegraphics[width=0.19\textwidth]{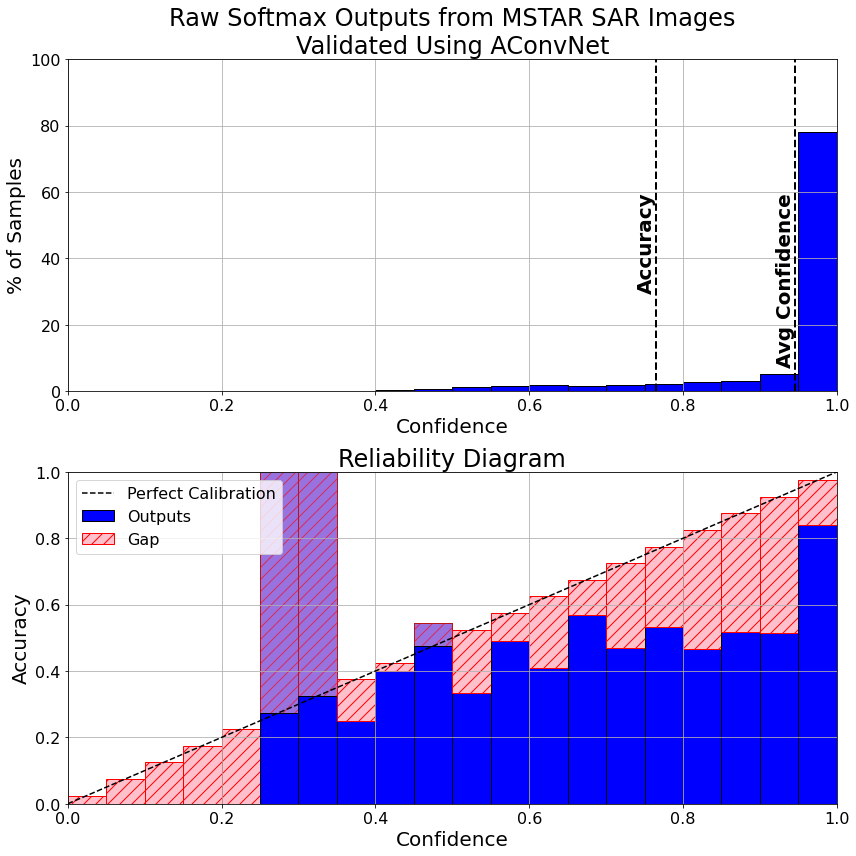}
    \includegraphics[width=0.19\textwidth]{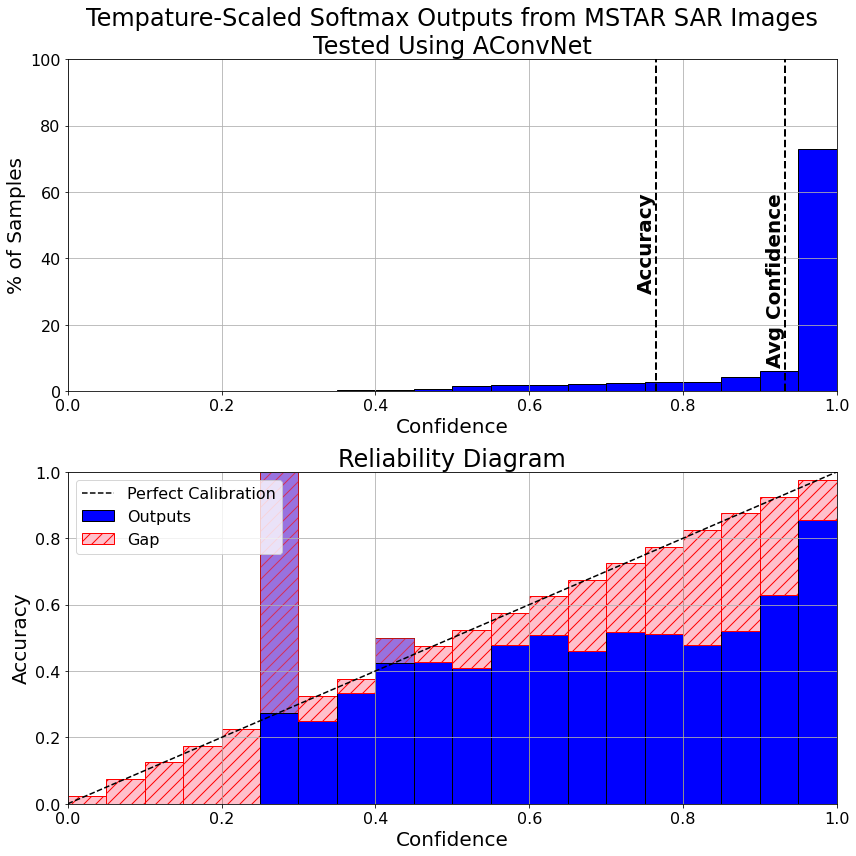}
    \includegraphics[width=0.19\textwidth]{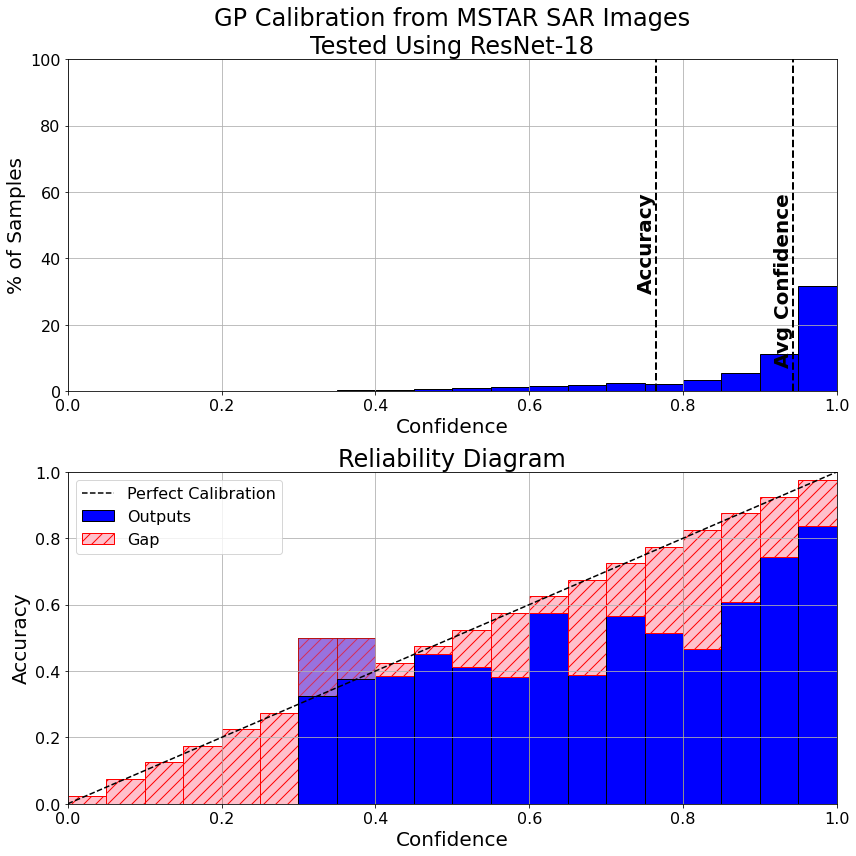}
    \includegraphics[width=0.19\textwidth]{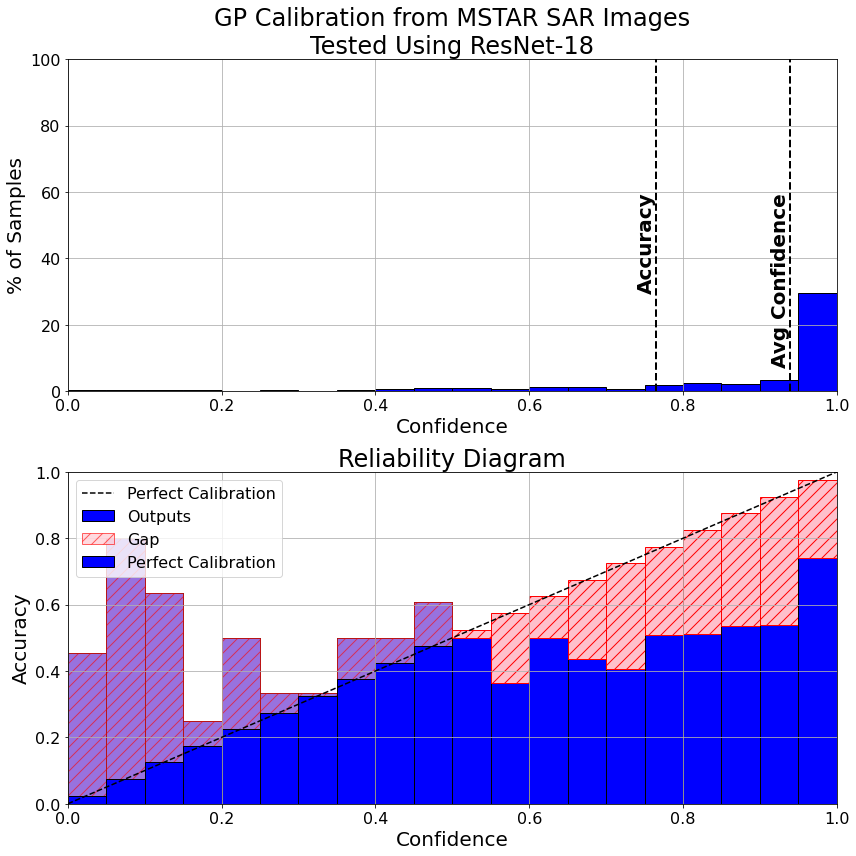}
    \includegraphics[width=0.19\textwidth]{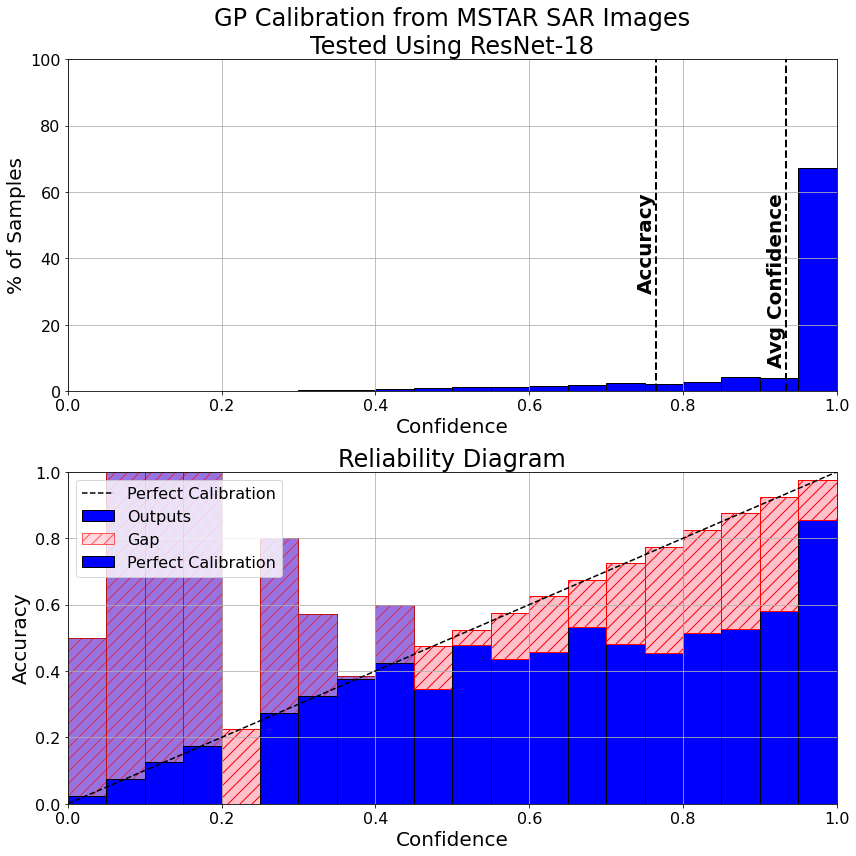} \\

    \begin{minipage}{0.19\textwidth}\centering Uncalibrated\end{minipage}
    \begin{minipage}{0.19\textwidth}\centering Temp. Scaled\end{minipage}
    \begin{minipage}{0.19\textwidth}\centering Single GP (Best Layer)\end{minipage}
    \begin{minipage}{0.19\textwidth}\centering SAL-GP (ML)\end{minipage}
    \begin{minipage}{0.19\textwidth}\centering SAL-GP (HL)\end{minipage}

    \caption{Calibration residual fit plots (top row) and reliability diagrams (bottom row) for each calibration approach on the all convolutional neural network(AConvNet), evaluated on the MSTAR dataset with max pooling. The top row displays only the residual fit plots for GP-based methods. For uncalibrated and temperature scaled columns, no residual fit is shown, since these methods do not provide predictive uncertainty estimates. Each column corresponds to a different calibration method: uncalibrated, temperature scaled, single-layer GP (best-performing layer), SAL-GP (multi-layer), and proposed SAL-GP (hierarchical layer kernel).}
    \label{fig:calibration_comparison_AConvNet_mstarWithMaxPooling}
\end{figure*}

\begin{figure*}[t]
    \centering
    \hspace*{0.19\textwidth}
    \hspace*{0.19\textwidth}
    \includegraphics[width=0.19\textwidth]{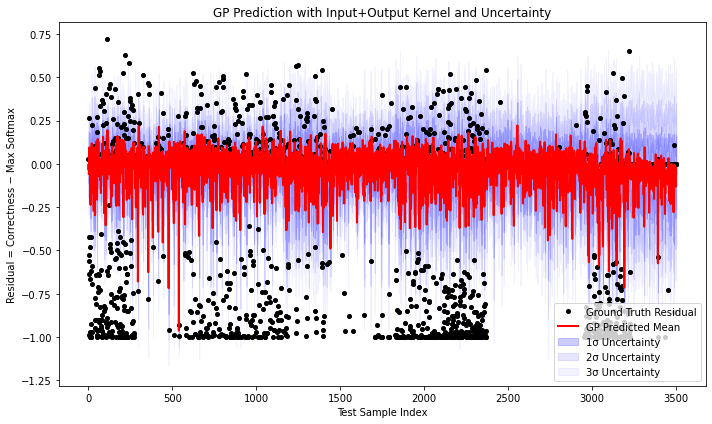}
    \includegraphics[width=0.19\textwidth]{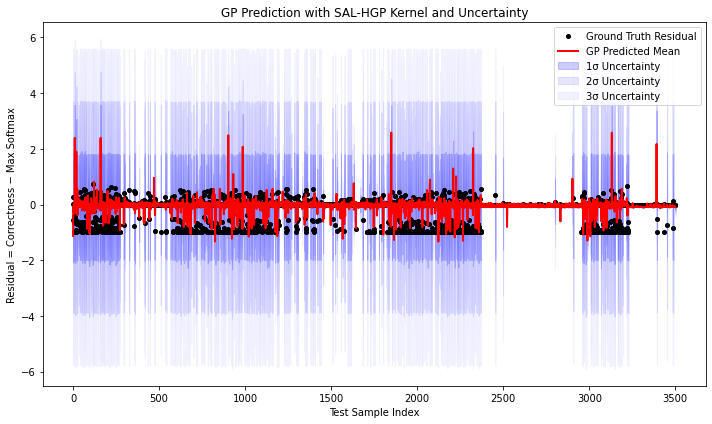}
    \includegraphics[width=0.19\textwidth]{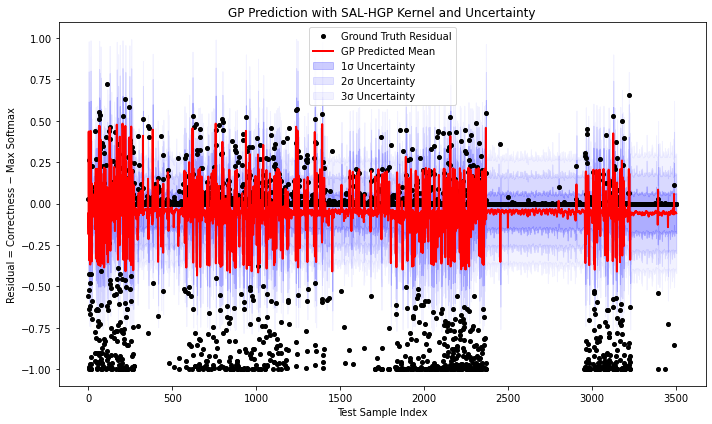} \\[2pt]
    \includegraphics[width=0.19\textwidth]{AConvNet_Uncalibrated_RD_MSTAR.png}
    \includegraphics[width=0.19\textwidth]{AConvNet_Temperature_Scaled_RD_MSTAR.png}
    \includegraphics[width=0.19\textwidth]{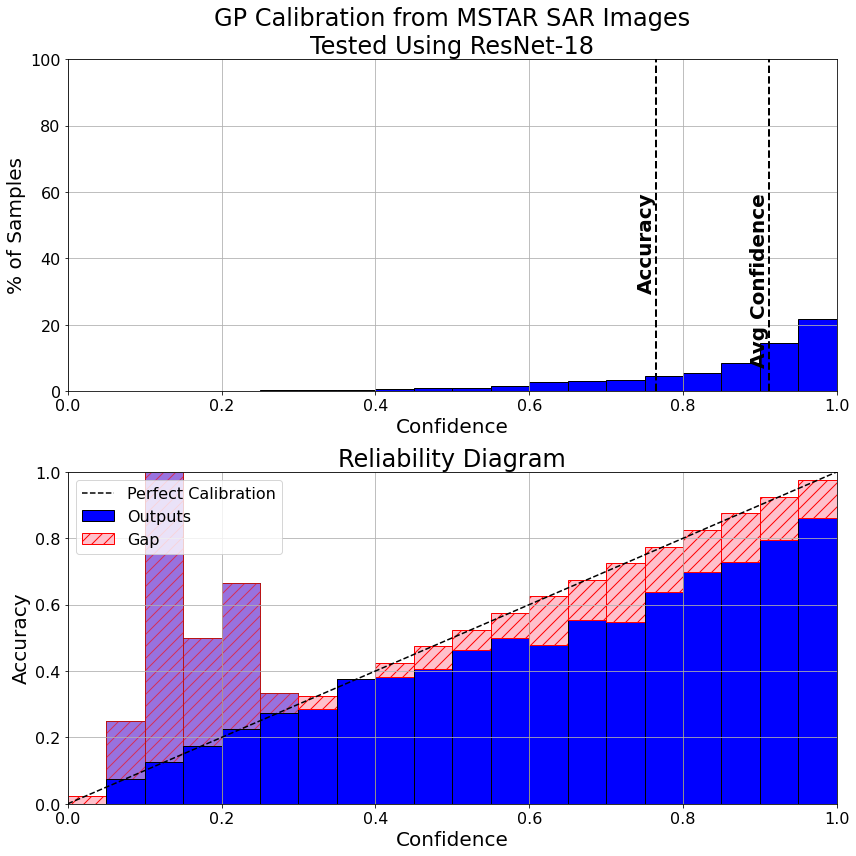}
    \includegraphics[width=0.19\textwidth]{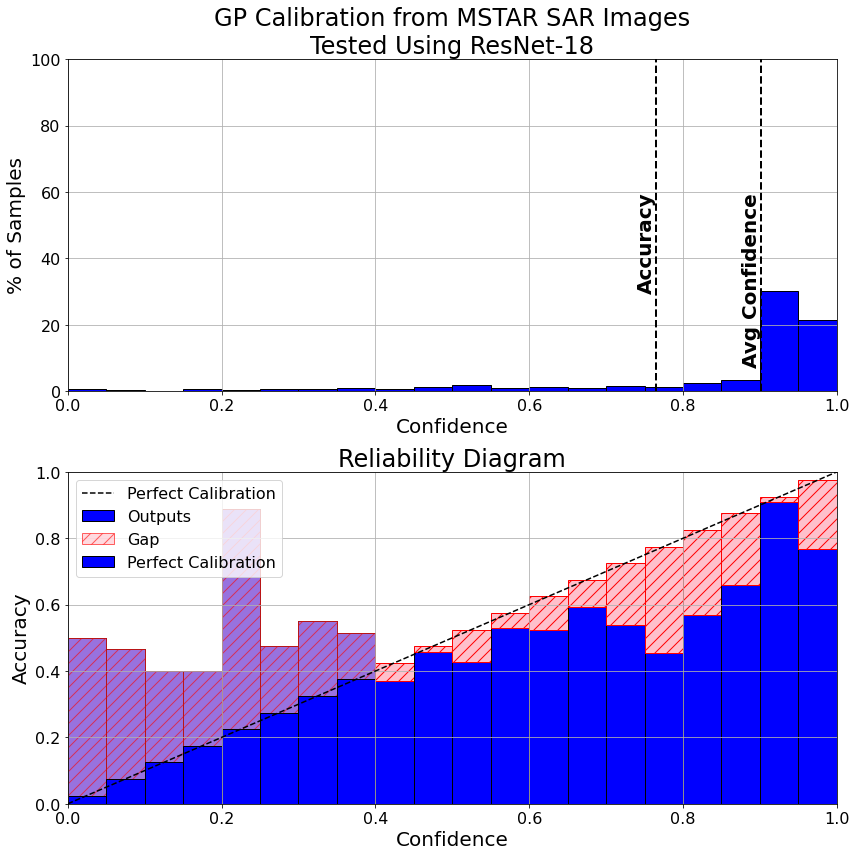}
    \includegraphics[width=0.19\textwidth]{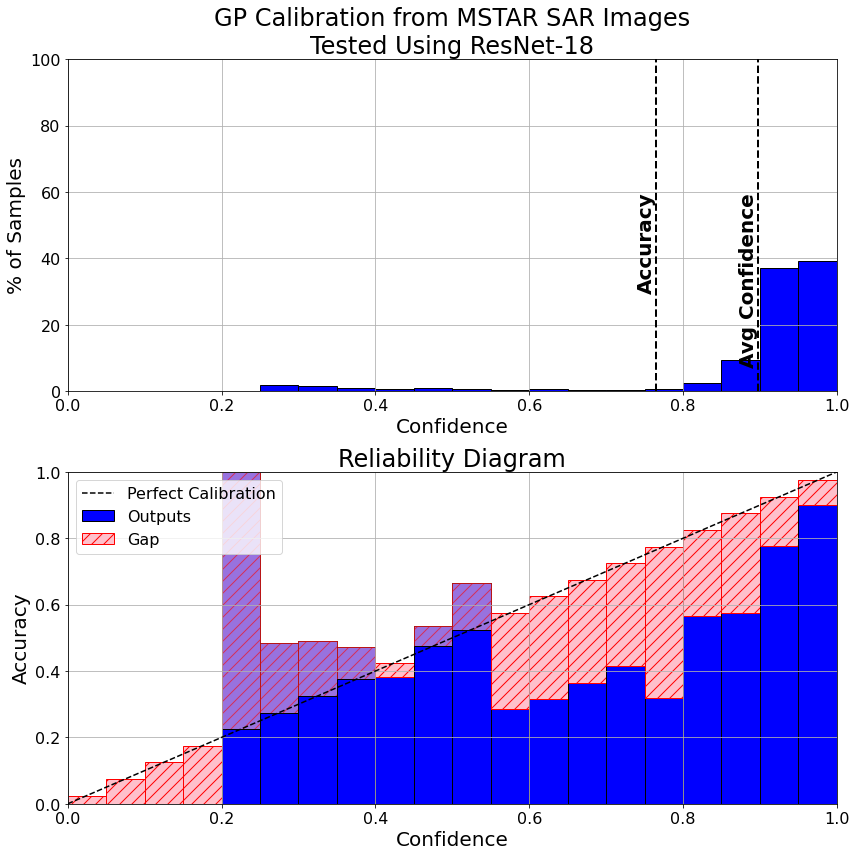} \\

    \begin{minipage}{0.19\textwidth}\centering Uncalibrated\end{minipage}
    \begin{minipage}{0.19\textwidth}\centering Temp. Scaled\end{minipage}
    \begin{minipage}{0.19\textwidth}\centering Single GP (Best Layer)\end{minipage}
    \begin{minipage}{0.19\textwidth}\centering SAL-GP (ML)\end{minipage}
    \begin{minipage}{0.19\textwidth}\centering SAL-GP (HL)\end{minipage}

    \caption{Calibration residual fit plots (top row) and reliability diagrams (bottom row) for each calibration approach on the all convolutional neural network (AConvNet), evaluated on the MSTAR dataset with average pooling. The top row displays only the residual fit plots for GP-based methods. For uncalibrated and temperature scaled columns, no residual fit is shown, since these methods do not provide predictive uncertainty estimates. Each column corresponds to a different calibration method: uncalibrated, temperature scaled, single-layer GP (best-performing layer), SAL-GP (multi-layer), and proposed SAL-GP (hierarchical layer kernel).}
    \label{fig:calibration_comparison_AConvNet_mstarWithAvgPooling}
\end{figure*}

However, it is important to note the high variability in performance across layers and pooling methods in AConvNet: ECEs for different single-layer GPs (max pooling) range from 0.11130 to 0.14431; for average pooling, from 0.08800 to 0.12613. This variability is substantially greater than in the standard ConvNet case, directly reflecting the increased sensitivity of AConvNet’s calibration performance to architectural choices. Such broad dispersion in ECE and MCE values, quantitatively supported in Fig.\ref{fig:calibration_comparison_AConvNet_mstarWithMaxPooling} and Fig.\ref{fig:calibration_comparison_AConvNet_mstarWithAvgPooling} and visually evident from the irregular reliability diagrams, underscores a prior selection key challenge.

The proposed SAL-GP(ML) achieves further improvements in calibration compared to all single-layer and baseline methods. Compared all layers and pooling types of single layer GP, the SAL-GP(ML) calibration's ECE remains consistently improved with 0.08532 of ECE for average pooling), and the key advantage of SAL-GP lies in its enhanced consistency and robustness as a unified calibrator, in contrast to the greater variability observed in the single-layer case. Although the SAL-GP(ML) shows a lesser degree of improvement in MCE, as well as increased NLL and predictive variance, particularly for global predictions, it nonetheless provides a clear advantage in risk management and calibration stability over the single-GP framework.

While SAL-GP(HL) also improves calibration compared to the uncalibrated baseline, most results fall between 0.13351 and 0.15946 for average pooling and max pooling, respectively, indicating underperformance relative to selecting a random single-layer GP. Regardless of the robustness of local GP calibration across configurations, the application of SAL-GP(HL) is still limited in this AConvNet experimental setup as well.

For AConvNet, average pooling consistently outperforms max pooling across all models in terms of ECE, MCE, and Brier score. This is more pronounced than in the standard ConvNet case and suggests that spatial aggregation in this architecture is essential for regularizing overconfident predictions.

The residual fit plots in Fig.\ref{fig:calibration_comparison_AConvNet_mstarWithMaxPooling} and Fig.\ref{fig:calibration_comparison_AConvNet_mstarWithAvgPooling} provide further insights into the distinct behaviors of each calibration model. Both SAL-GP variant models exhibit markedly different kernel adaptation compared to the single-layer GP, particularly in their ability to model both positive residuals (samples where the network is underconfident) and negative residuals (samples where the network is overconfident).

The best single-layer GP tends to focus on moderate residuals across most test samples, with only occasional capture of large negative residuals, and generally fails to model many high-magnitude residuals, especially under average pooling. This limited flexibility results in a calibration that is more locally smooth but can miss systematic patterns of miscalibration, particularly in regions of strong under- or overconfidence.

In contrast, the SAL-GP(ML) is more responsive to negative residuals, more effectively capturing overconfident test samples where the predicted confidence exceeds actual correctness. This behavior is visible as the SAL-GP(ML) mean closely tracks the ground truth residuals, especially for negative outliers, and is reflected in the uncertainty bounds that more consistently represent the spread of the residuals. This capacity to adapt to overconfidence directly addresses a common shortcoming in standard post-hoc calibration.

The SAL-GP(HL) model, while less sensitive to residuals near zero compared to the single-layer GP, is better able to model substantial negative residuals, capturing overconfident cases more robustly than the single-layer approach. However, it demonstrates less responsiveness to positive residuals, indicating potential underfitting for samples where the network is underconfident.

These qualitative differences are visually evident in the residual plots (top row of each figure) and correspond directly to the patterns in the reliability diagrams (bottom row). Both hierarchical models exhibit calibration behaviors indicative of overconfidence, as shown by negative values in the mean residual prediction, which could potentially result in underconfidence for some test samples, as illustrated in the reliability diagrams. These observations highlight the need for further careful kernel design to better capture intra- and inter-network latent space relationships.

\begin{table*}[htbp]
\centering
\caption{\textsc{Main Calibration and Accuracy Metrics: All-Convolutional Network (AConvNet)}}
\label{tab:calibration_metrics_aconvnet}  
\begin{threeparttable}
\begin{tabular}{ll ccccc}
\hline\hline
\multicolumn{7}{c}{\textbf{Calibration Performance Metrics: AConvNet}} \\ \hline
\textbf{Pooling} & \textbf{Method} & \textbf{ECE$\downarrow$} & \textbf{MCE$\downarrow$} & \textbf{NLL$\downarrow$} & \textbf{Brier$\downarrow$} & \textbf{Var.} \\ \hline
- & Uncalibrated (Baseline)      & 0.18257 & 0.72248 & 1.24730 & 0.40602 & - \\
- & Temperature Scaled (Baseline) & 0.16855 & 0.74472 & 1.04467 & 0.39331 & - \\
Max & Single GP (Layer 1)          & 0.11130 & 0.36280 & 1.25671 & 0.40550 & 0.00439 \\
Avg & Single GP (Layer 1)          & 0.09335 & 0.84504 & 1.27182 & 0.40428 & 0.00493 \\
Max & Single GP (Layer 2)          & 0.13613 & 0.71763 & 1.25186 & 0.40522 & 0.00668 \\
Avg & Single GP (Layer 2)          & 0.08800 & 0.86030 & 1.27235 & 0.40409 & 0.01429 \\
Max & Single GP (Layer 3)          & 0.12996 & 0.37020 & 1.25334 & 0.40528 & 0.00688 \\
Avg & Single GP (Layer 3)          & 0.11462 & 0.31248 & 1.26665 & 0.40490 & 0.00464 \\
Max & Single GP (Layer 4)          & 0.12780 & 0.37834 & 1.25310 & 0.40624 & 0.00524 \\
Avg & Single GP (Layer 4)          & 0.10864 & 0.89850 & 1.26317 & 0.40530 & 0.00447 \\
Max & Single GP (Layer 5)          & 0.14431 & 0.66318 & 1.25117 & 0.40543 & 0.00643 \\
Avg & Single GP (Layer 5)          & 0.12613 & 0.33629 & 1.26435 & 0.40483 & 0.00932 \\
Max & SAL-GP (ML) G        & 0.12202 & 0.73160 & 1.46030 & 0.41613 & 0.26033 \\
Max & SAL-GP (ML) L1        & 0.11390 & 0.74721 & 1.46146 & 0.41628 & 0.26534 \\
Max & SAL-GP (ML) L2        & 0.12499 & 0.74396 & 1.46729 & 0.41621 & 0.26473 \\
Max & SAL-GP (ML) L3        & 0.13495 & 0.71782 & 1.46779 & 0.41617 & 0.26500 \\
Max & SAL-GP (ML) L4        & 0.12926 & 0.59363 & 1.46035 & 0.41590 & 0.26422 \\
Max & SAL-GP (ML) L5        & 0.13526 & 0.81549 & 1.46580 & 0.41576 & 0.26517 \\
Avg & SAL-GP (ML) G         & 0.08532 & 0.65580 & 1.40060 & 0.40908 & 0.61551 \\
Avg & SAL-GP (ML) L1        & 0.10238 & 0.62260 & 1.40114 & 0.40925 & 0.65645 \\
Avg & SAL-GP (ML) L2        & 0.10670 & 0.52043 & 1.40149 & 0.40915 & 0.65403 \\
Avg & SAL-GP (ML) L3        & 0.09869 & 0.53037 & 1.39439 & 0.40927 & 0.65185 \\
Avg & SAL-GP (ML) L4        & 0.10016 & 0.60485 & 1.39471 & 0.40935 & 0.65068 \\
Avg & SAL-GP (ML) L5        & 0.10892 & 0.53619 & 1.40120 & 0.40919 & 0.65610 \\
Max & SAL-GP (HL) G  & 0.15946 & 0.90763 & 1.26343 & 0.40547 & 0.00548 \\
Max & SAL-GP (HL) L1  & 0.12605 & 0.74262 & 1.25040 & 0.40460 & 0.00649 \\
Max & SAL-GP (HL) L2  & 0.13456 & 0.84715 & 1.25102 & 0.40461 & 0.00601 \\
Max & SAL-GP (HL) L3  & 0.14607 & 0.82543 & 1.25064 & 0.40456 & 0.00625 \\
Max & SAL-GP (HL) L4  & 0.13885 & 0.85803 & 1.25120 & 0.40452 & 0.00521 \\
Max & SAL-GP (HL) L5  & 0.12992 & 0.80629 & 1.24956 & 0.40452 & 0.00633 \\
Avg & SAL-GP (HL) G  & 0.13351 & 0.75457 & 1.28032 & 0.40662 & 0.01284 \\
Avg & SAL-GP (HL) L1  & 0.13999 & 0.35599 & 1.25147 & 0.40666 & 0.01965 \\
Avg & SAL-GP (HL) L2  & 0.15174 & 0.41876 & 1.25178 & 0.40766 & 0.01279 \\
Avg & SAL-GP (HL) L3  & 0.14273 & 0.71520 & 1.25643 & 0.40833 & 0.00694 \\
Avg & SAL-GP (HL) L4  & 0.14621 & 0.5740 & 1.25479 & 0.40787 & 0.00414 \\
Avg & SAL-GP (HL) L5  & 0.14172 & 0.76029 & 1.25249 & 0.40806 & 0.01640 \\ \hline
\textbf{} & \textbf{Accuracy(Train, Validation, Test)}      & \multicolumn{5}{c}{98.37\%, 98.64\%, 76.44\%} \\ \hline
\end{tabular}
\begin{tablenotes}
\footnotesize
\item Calibration metrics: ECE (Expected Calibration Error), MCE (Maximum Calibration Error), NLL (Negative Log-Likelihood), Brier score, and mean predictive variance (Var.).
\item Accuracy is reported for the original neural network and is unchanged across calibration methods.
\item For "Single GP," each row corresponds to a GP trained using only the $l$-th layer's feature map plus softmax as input.
\item "SAL-GP (ML)" uses all layers as input with the structured multi-layer index kernel.
\item "SAL-GP (HL)" uses all layers as input with the hierarchical layer kernel.
\item Optimized T = 1.2278

\end{tablenotes}
\end{threeparttable}
\end{table*}


\subsubsection{ResNet18 Network}
Table~\ref{tab:calibration_metrics_resnet18} presents the main calibration and accuracy metrics for the ResNet-18 network, including predictive variance and the results of ablation studies.
ResNet-18, unlike standard ConvNet and AConvNet, exhibits moderate miscalibration out of the box, with uncalibrated ECE of 0.03447 and a Brier score of 0.10860, ironically higher than the ConvNet baseline. This indicates a greater gap between predicted confidence and empirical correctness, necessitating careful post-hoc calibration for trustworthy deployment.

\begin{figure*}[t]
    \centering
    \hspace*{0.19\textwidth}
    \hspace*{0.19\textwidth}
    \includegraphics[width=0.19\textwidth]{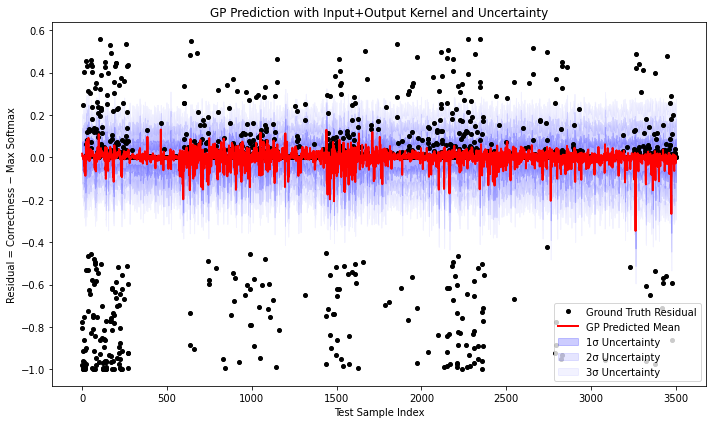}
    \includegraphics[width=0.19\textwidth]{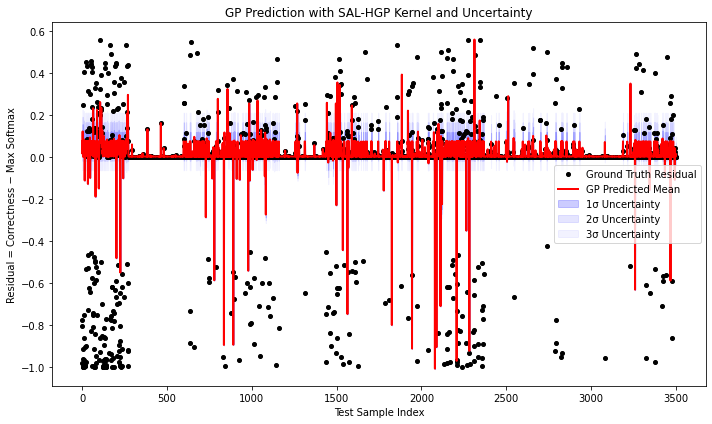}
    \includegraphics[width=0.19\textwidth]{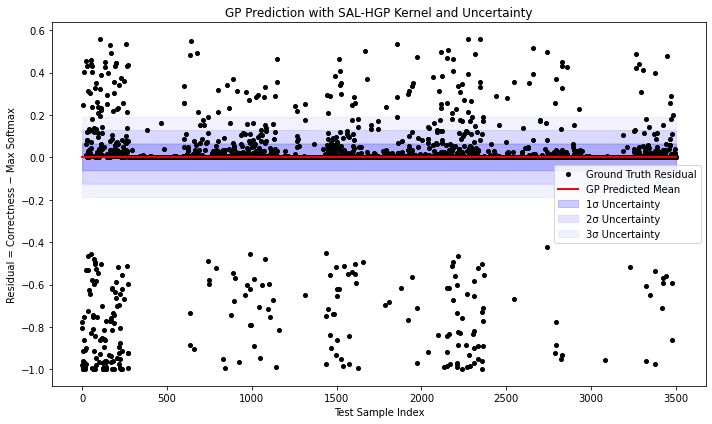} \\[2pt]
    \includegraphics[width=0.19\textwidth]{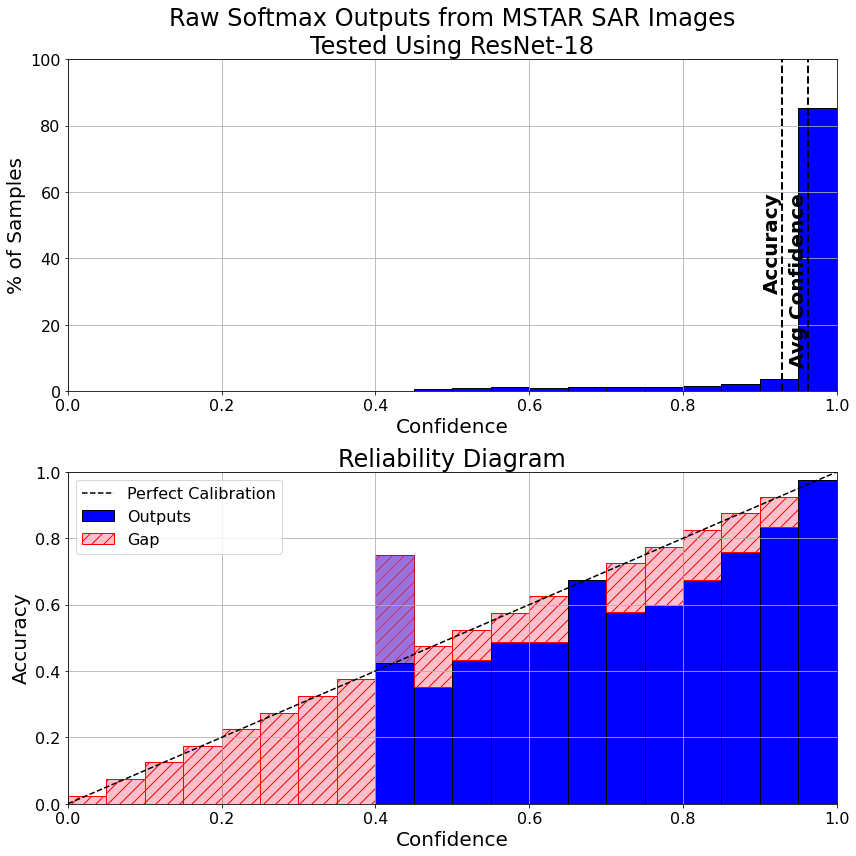}
    \includegraphics[width=0.19\textwidth]{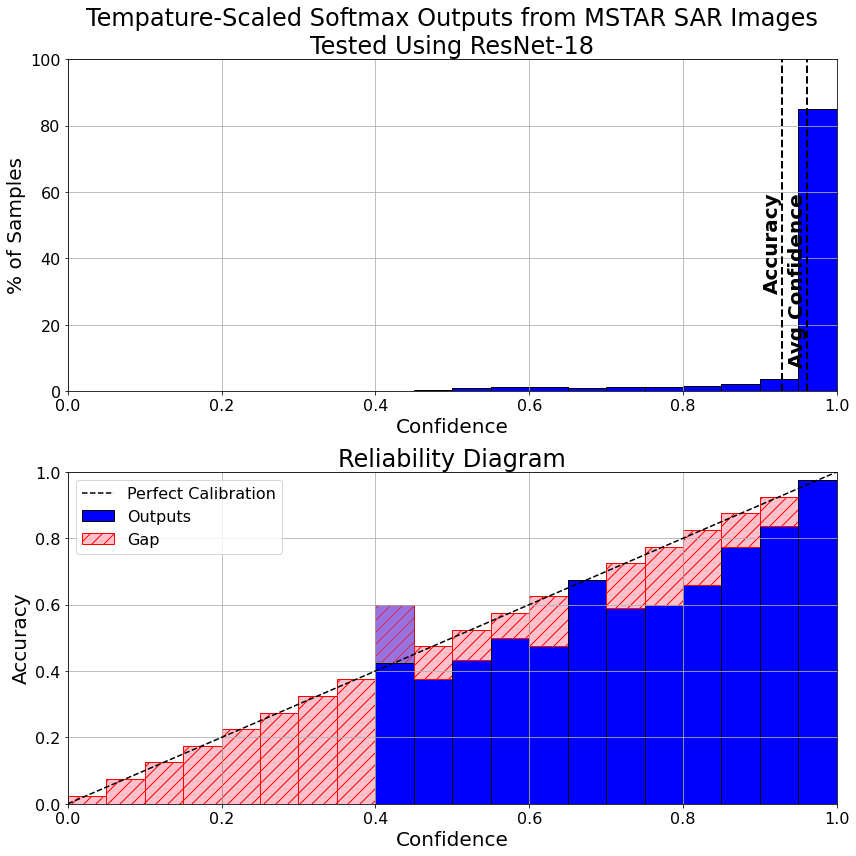}
    \includegraphics[width=0.19\textwidth]{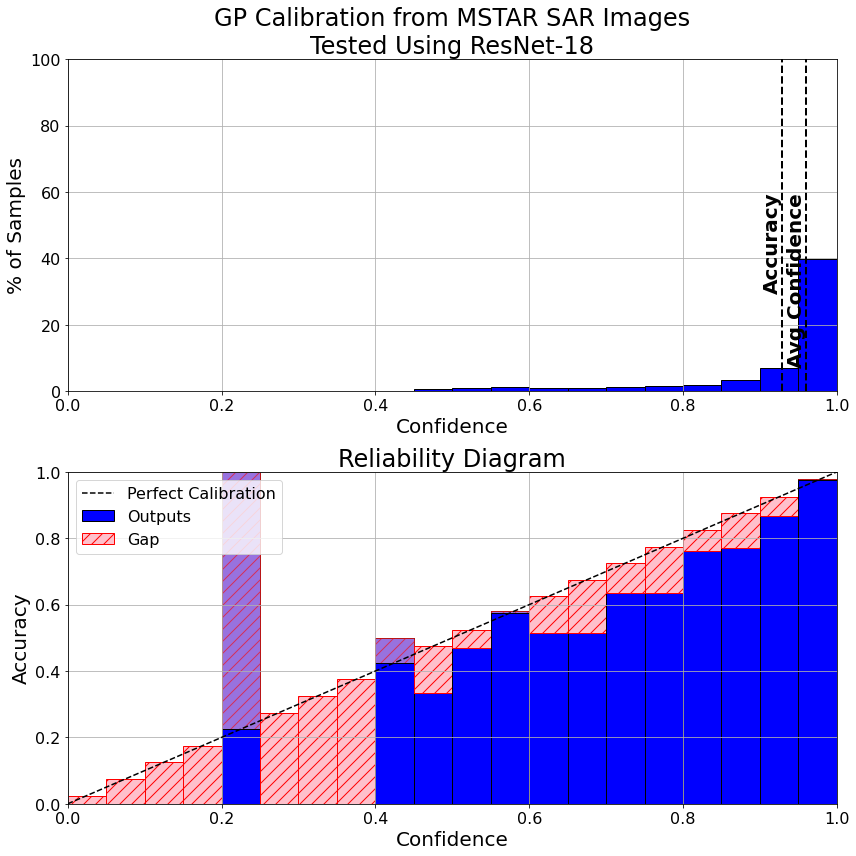}
    \includegraphics[width=0.19\textwidth]{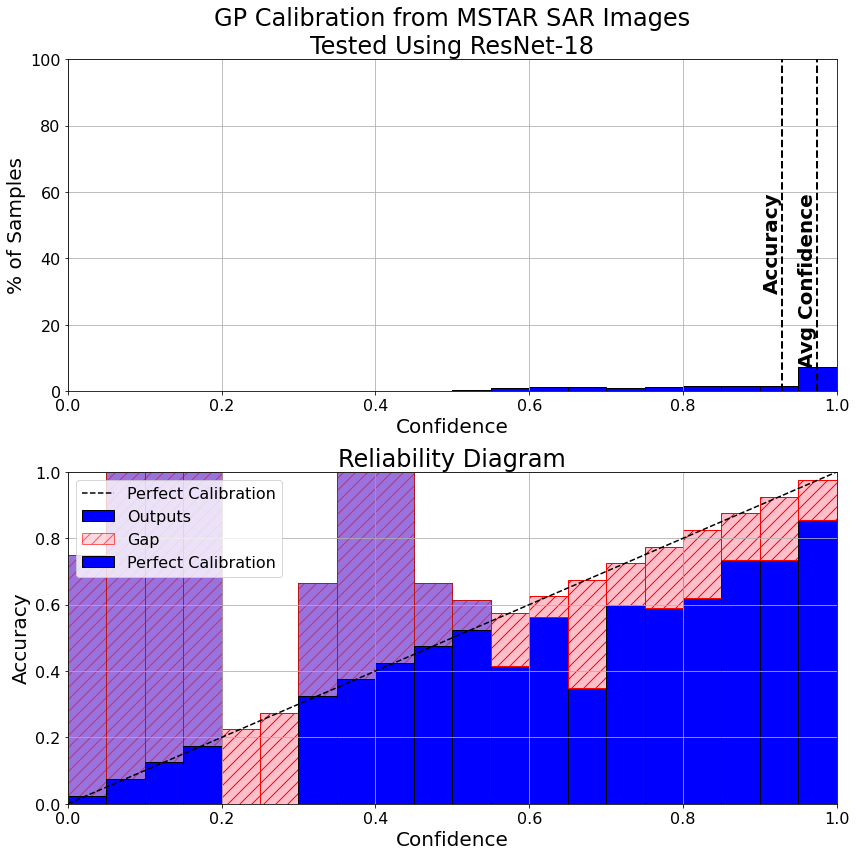}
    \includegraphics[width=0.19\textwidth]{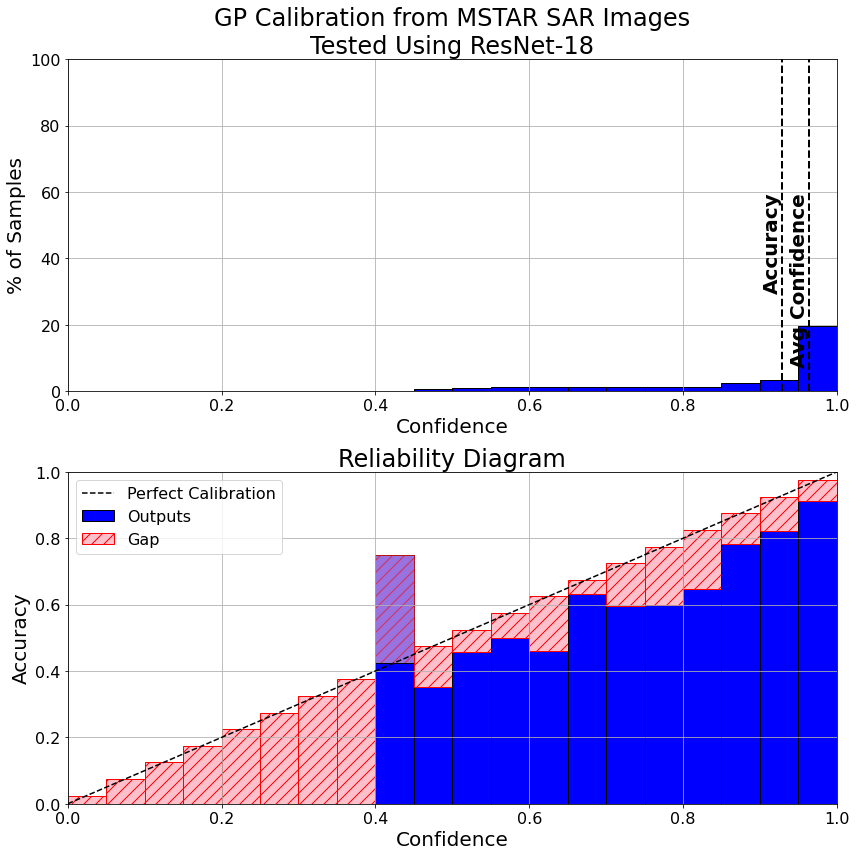} \\

    \begin{minipage}{0.19\textwidth}\centering Uncalibrated\end{minipage}
    \begin{minipage}{0.19\textwidth}\centering Temp. Scaled\end{minipage}
    \begin{minipage}{0.19\textwidth}\centering Single GP (Best Layer)\end{minipage}
    \begin{minipage}{0.19\textwidth}\centering SAL-GP (ML)\end{minipage}
    \begin{minipage}{0.19\textwidth}\centering SAL-GP (HL)\end{minipage}

    \caption{Calibration residual fit plots (top row) and reliability diagrams (bottom row) for each calibration approach on the ResNet18, evaluated on the MSTAR dataset with max pooling. The top row displays only the residual fit plots for GP-based methods. For uncalibrated and temperature scaled columns, no residual fit is shown, since these methods do not provide predictive uncertainty estimates. Each column corresponds to a different calibration method: uncalibrated, temperature scaled, single-layer GP (best-performing layer), SAL-GP (multi-layer), and proposed SAL-GP (hierarchical layer kernel).}
    \label{fig:calibration_comparison_ResNet18_mstarWithMaxPooling}
\end{figure*}

\begin{figure*}[t]
    \centering
    \hspace*{0.19\textwidth}
    \hspace*{0.19\textwidth}
    \includegraphics[width=0.19\textwidth]{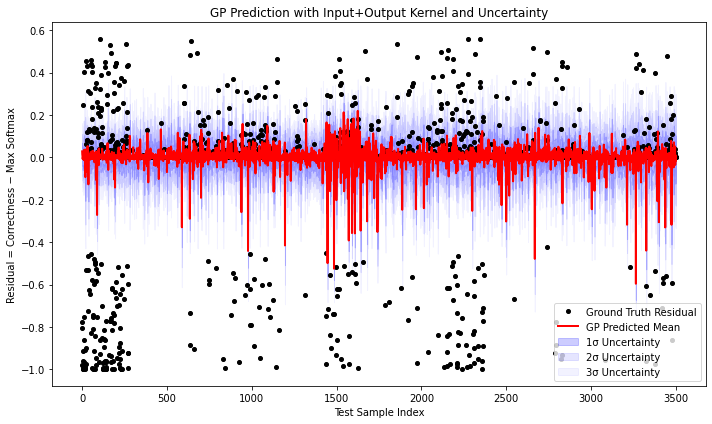}
    \includegraphics[width=0.19\textwidth]{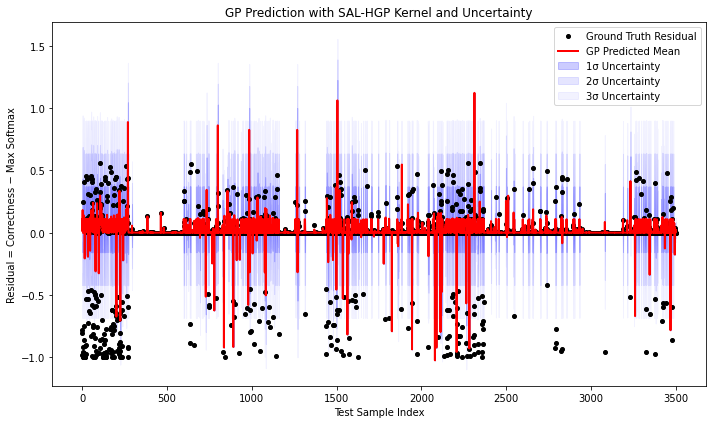}
    \includegraphics[width=0.19\textwidth]{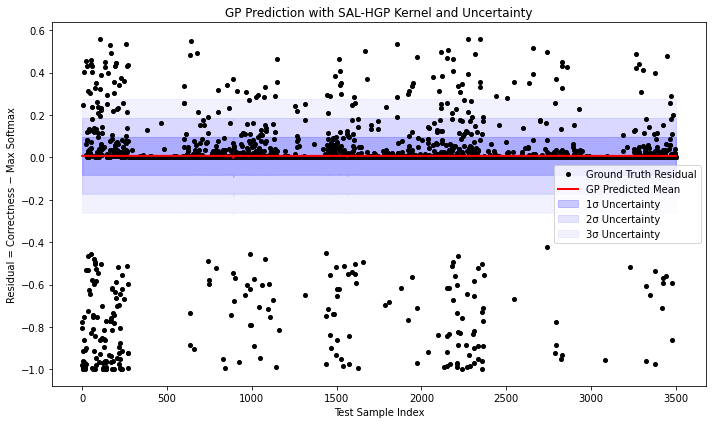} \\[2pt]
    \includegraphics[width=0.19\textwidth]{ResNet18_Uncalibrated_RD_MSTAR.png}
    \includegraphics[width=0.19\textwidth]{ResNet18_Temperature_Scaled_RD_MSTAR.png}
    \includegraphics[width=0.19\textwidth]{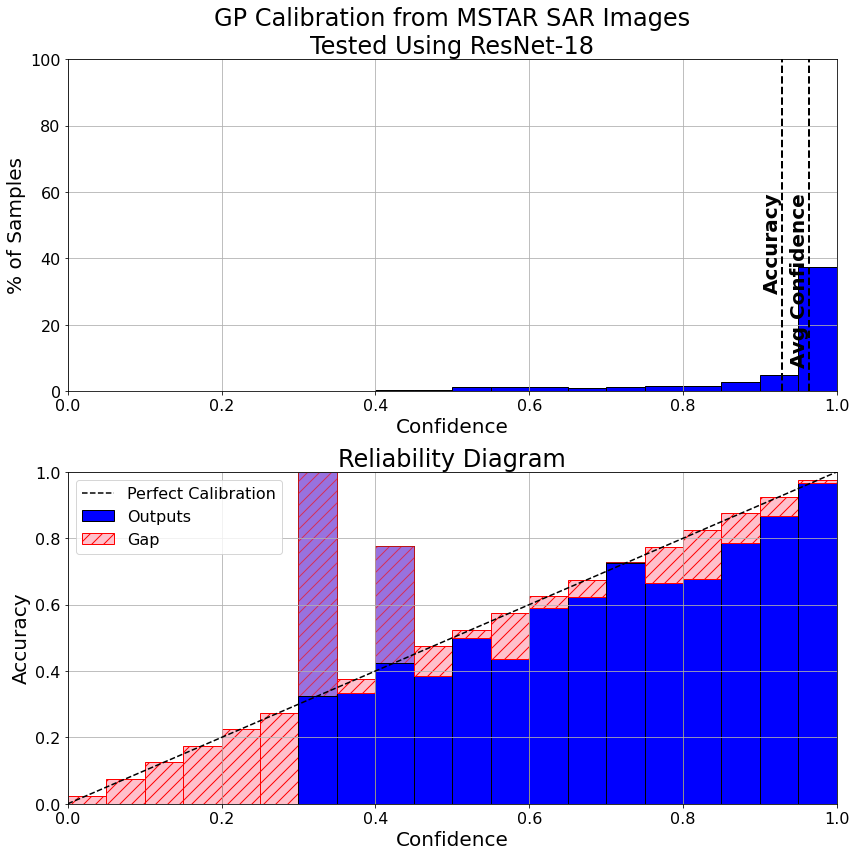}
    \includegraphics[width=0.19\textwidth]{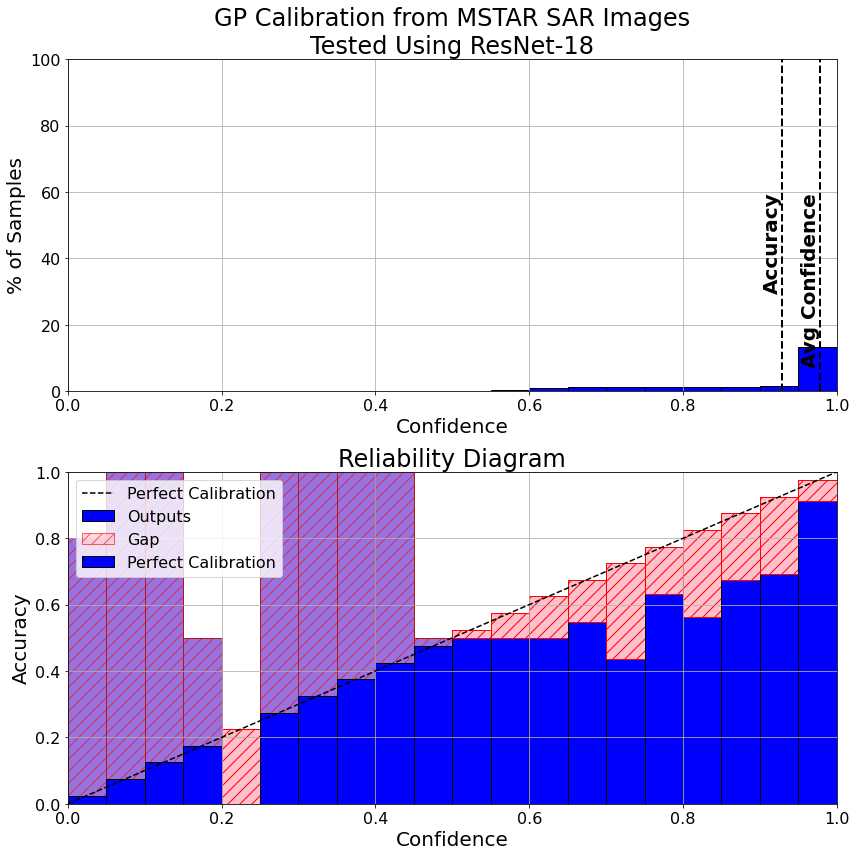}
    \includegraphics[width=0.19\textwidth]{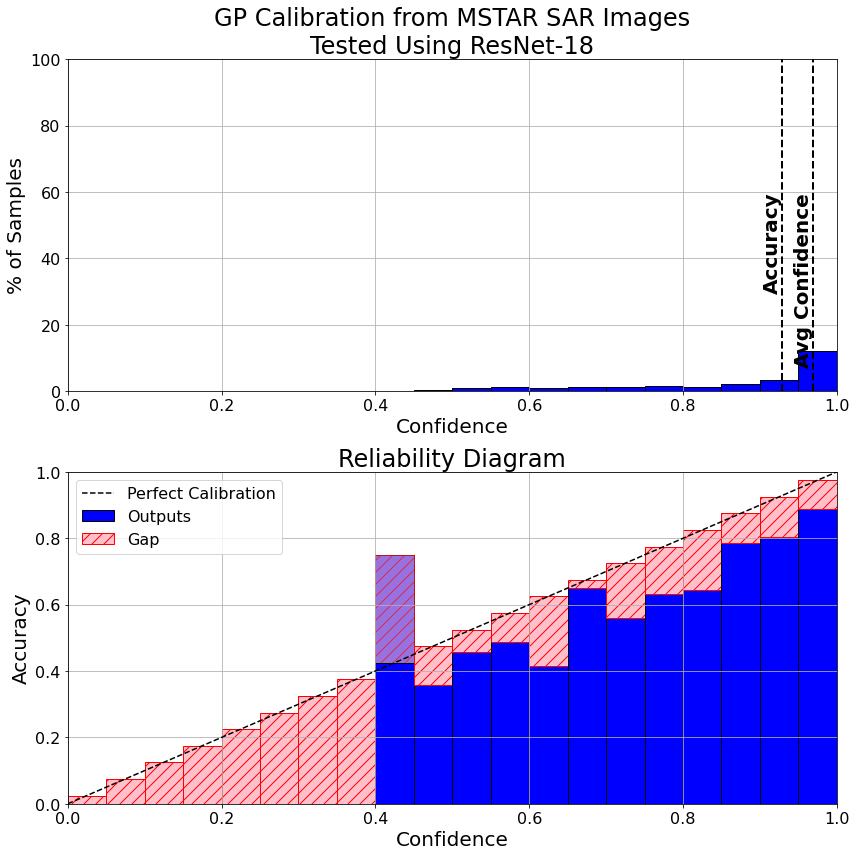} \\

    \begin{minipage}{0.19\textwidth}\centering Uncalibrated\end{minipage}
    \begin{minipage}{0.19\textwidth}\centering Temp. Scaled\end{minipage}
    \begin{minipage}{0.19\textwidth}\centering Single GP (Best Layer)\end{minipage}
    \begin{minipage}{0.19\textwidth}\centering SAL-GP (ML)\end{minipage}
    \begin{minipage}{0.19\textwidth}\centering SAL-GP (HL)\end{minipage}

    \caption{Calibration residual fit plots (top row) and reliability diagrams (bottom row) for each calibration approach on the ResNet18, evaluated on the MSTAR dataset with average pooling. The top row displays only the residual fit plots for GP-based methods. For uncalibrated and temperature scaled columns, no residual fit is shown, since these methods do not provide predictive uncertainty estimates. Each column corresponds to a different calibration method: uncalibrated, temperature scaled, single-layer GP (best-performing layer), SAL-GP (multi-layer), and proposed SAL-GP (hierarchical layer kernel).}
    \label{fig:calibration_comparison_ResNet18_mstarWithAvgPooling}
\end{figure*}

TS provides only a negligible benefit (ECE: 0.03397, Brier: 0.10853), and, as seen in Fig.\ref{fig:calibration_comparison_ResNet18_mstarWithMaxPooling} and Fig.\ref{fig:calibration_comparison_ResNet18_mstarWithAvgPooling}, leaves the characteristic overconfident peak near 1.0 almost unchanged. This reflects the well-known limitation that temperature scaling cannot correct for miscalibration modes other than simple global overconfidence, particularly when the underlying logits are already highly separable or the calibration data lacks sufficient uncertainty diversity, similar to ConvNet scenario.

Single-layer GP calibration (best: Layer 4, max pooling, ECE: 0.01972; Layer 3, average pooling, ECE: 0.02205) delivers a tangible but non-uniform improvement over the baseline. However, as in the previous other experiments, the optimal layer for calibration depends on the pooling type and cannot be identified without access to the test set, highlighting a fundamental limitation. Furthermore, unlike the previous two cases, the superiority of average pooling for single-layer GP calibration is not observed here. The best layer for single GP is with Max pooling method, unlike other two different network backbone cases.

Both SAL-GP variants yield a reasonable improvement in ECE but are less effective across other metrics. The SAL-GP(HL) achieves better calibration with average pooling, while the SAL-GP(ML) performs best with max pooling. Notably, the SAL-GP(HL) does not increase MCE, whereas the SAL-GP(ML) significantly elevates MCE due to assigning low confidence to samples that are actually correct. A similar effect is observed for the best single-layer GP (Layers 3 and 4), which, despite achieving the lowest ECE, also results in a substantially higher MCE (0.76980). This suggests that these specific layers may contribute to the increased MCE seen in the global predictions of the SAL-GP(ML).

The comparative plots in Fig.\ref{fig:calibration_comparison_ResNet18_mstarWithMaxPooling} and Fig.\ref{fig:calibration_comparison_ResNet18_mstarWithAvgPooling} illustrate distinct calibration behaviors. The best single-layer GP shows moderate residual fitting, capturing central residuals but struggling with large overconfident errors, as reflected by persistent gaps in the reliability diagram at high confidence. The SAL-GP(ML) is more responsive to negative residuals, effectively capturing overconfident predictions and expressing greater uncertainty for diverse residual magnitudes, resulting in reduced calibration gaps across a broader confidence range. In contrast, the SAL-GP(HL) model produces smoother residual fits, concentrating on residuals near zero but showing limited ability to detect large residual errors, particularly in overconfident scenarios. The reliability diagrams for SAL-GP(HL) and SAL-GP(ML) reveal discrepancies in their ability to detect residual components at both low and high confidence levels. This suggests that the hierarchical structure of SAL-GP(HL) restricts the model’s capacity to predict extreme corrections, resulting in differing adaptability compared to SAL-GP(ML).

\begin{table*}[htbp]
\centering
\caption{\textsc{Main Calibration and Accuracy Metrics: ResNet-18}}
\label{tab:calibration_metrics_resnet18}  
\begin{threeparttable}
\begin{tabular}{ll ccccc}
\hline\hline
\multicolumn{7}{c}{\textbf{Calibration Performance Metrics: ResNet-18}} \\ \hline
\textbf{Pooling} & \textbf{Method} & \textbf{ECE$\downarrow$} & \textbf{MCE$\downarrow$} & \textbf{NLL$\downarrow$} & \textbf{Brier$\downarrow$} & \textbf{Var.} \\ \hline
- & Uncalibrated (Baseline)      & 0.03447 & 0.31247 & 0.22809 & 0.10860 & - \\
- & Temperature Scaled (Baseline) & 0.03397 & 0.17125 & 0.22752 & 0.10853 & - \\
Max & Single GP (Layer 1)          & 0.03157 & 0.31062 & 0.22743 & 0.10863 & 0.00371 \\
Avg & Single GP (Layer 1)          & 0.02555 & 0.77225 & 0.23297 & 0.10831 & 0.00394 \\
Max & Single GP (Layer 2)          & 0.03186 & 0.31078 & 0.22743 & 0.10863 & 0.00373 \\
Avg & Single GP (Layer 2)          & 0.02462 & 0.27719 & 0.23199 & 0.10843 & 0.00374 \\
Max & Single GP (Layer 3)          & 0.03208 & 0.31074 & 0.22748 & 0.10863 & 0.00373 \\
Avg & Single GP (Layer 3)          & 0.02205 & 0.66595 & 0.23558 & 0.10830 & 0.00311 \\
Max & Single GP (Layer 4)          & 0.01972 & 0.76980 & 0.23552 & 0.10858 & 0.00476 \\
Avg & Single GP (Layer 4)          & 0.02502 & 0.37908 & 0.23382 & 0.10824 & 0.00134 \\
Max & Single GP (Layer 5)          & 0.02828 & 0.28134 & 0.22963 & 0.10886 & 0.00134 \\
Avg & Single GP (Layer 5)          & 0.02664 & 0.30788 & 0.23053 & 0.10880 & 0.00052 \\
Max & SAL-GP (ML) G        & 0.02983 & 0.92416 & 0.23640 & 0.11095 & 0.00026 \\
Max & SAL-GP (ML) L1        & 0.03313 & 0.92501 & 0.23649 & 0.11095 & 0.01487 \\
Max & SAL-GP (ML) L2        & 0.03007 & 0.92456 & 0.23641 & 0.11094 & 0.01487 \\
Max & SAL-GP (ML) L3        & 0.03286 & 0.92493 & 0.23647 & 0.11095 & 0.01487 \\
Max & SAL-GP (ML)) L4        & 0.03147 & 0.92497 & 0.23644 & 0.11094 & 0.01476 \\
Max & SAL-GP (ML) L5        & 0.03065 & 0.92443 & 0.23642 & 0.11095 & 0.01463 \\
Avg & SAL-GP (ML) G         & 0.03243 & 0.93509 & 0.23813 & 0.11190 & 0.00976 \\
Avg & SAL-GP (ML) L1        & 0.03737 & 0.93599 & 0.23834 & 0.11192 & 0.01764 \\
Avg & SAL-GP (ML) L2        & 0.02886 & 0.93440 & 0.23802 & 0.11189 & 0.01764 \\
Avg & SAL-GP (ML) L3        & 0.04114 & 0.93739 & 0.23929 & 0.11194 & 0.01744 \\
Avg & SAL-GP (ML) L4        & 0.02843 & 0.93342 & 0.23789 & 0.11186 & 0.01741 \\
Avg & SAL-GP (ML) L5        & 0.03712 & 0.93614 & 0.23830 & 0.11192 & 0.01740 \\
Max & SAL-GP (HL) G  & 0.03134 & 0.31027 & 0.22741 & 0.10863 & 0.00401 \\
Max & SAL-GP (HL) L1  & 0.03142 & 0.31027 & 0.22732 & 0.10863 & 0.00793 \\
Max & SAL-GP (HL) L2  & 0.03140 & 0.31027 & 0.22734 & 0.10863 & 0.00793 \\
Max & SAL-GP (HL) L3  & 0.03142 & 0.31027 & 0.22732 & 0.10863 & 0.00793 \\
Max & SAL-GP (HL) L4  & 0.03143 & 0.31027 & 0.22732 & 0.10863 & 0.00793 \\
Max & SAL-GP (HL) L5  & 0.03110 & 0.73119 & 0.22841 & 0.10866 & 0.00321 \\
Avg & SAL-GP (HL) G  & 0.02874 & 0.30550 & 0.22651 & 0.10870 & 0.00798 \\
Avg & SAL-GP (HL) L1  & 0.0281 & 0.19250 & 0.22645 & 0.10870 & 0.00924 \\
Avg & SAL-GP (HL) L2  & 0.03537 & 0.44653 & 0.22656 & 0.10925 & 0.00592 \\
Avg & SAL-GP (HL) L3  & 0.03669 & 0.87221 & 0.22857 & 0.10946 & 0.00314 \\
Avg & SAL-GP (HL) L4  & 0.03839 & 0.87450 & 0.23010 & 0.10968 & 0.00270 \\
Avg & SAL-GP (HL) L5  & 0.03761 & 0.96199 & 0.22974 & 0.10957 & 0.00262 \\ \hline
\textbf{} & \textbf{Accuracy(Train, Validation, Test)}      & \multicolumn{5}{c}{99.86\%, 99.32\%, 92.83\%} \\ \hline
\end{tabular}
\begin{tablenotes}
\footnotesize
\item Calibration metrics: ECE (Expected Calibration Error), MCE (Maximum Calibration Error), NLL (Negative Log-Likelihood), Brier score, and mean predictive variance (Var.).
\item Accuracy is reported for the original neural network and is unchanged across calibration methods.
\item For "Single GP," each row corresponds to a GP trained using only the $l$-th layer's feature map plus softmax as input.
\item "SAL-GP (ML)" uses all layers as input with the structured multi-layer index kernel.
\item "SAL-GP (HL)" uses all layers as input with the hierarchical layer kernel.
\item Optimized T = 1.00453
\end{tablenotes}
\end{threeparttable}
\end{table*}

\subsection{Calibration and Accuracy with Ablation Study: PLAsTiCC Data}

Unlike standard image-based classification with CNNs, PLAsTiCC classification addresses time-series multi-band photometry with associated metadata. RNNs are particularly well-suited for this temporal structure, but the task is complicated by pronounced domain shift between training and test sets—unlike datasets such as MSTAR. The PLAsTiCC training data is highly limited and imbalanced (comprising less than 1\% of the test set) and does not capture the full diversity of the test distribution; several “unknown” classes appear in the test set but are absent from training.

These challenges closely reflect real-world conditions in astronomical survey pipelines, where severe class imbalance and non-overlapping distributions between training and test sets are common. In time-domain astronomy, labeled data is often unrepresentative, and new or rare classes frequently appear during inference.

Our experimental protocol addresses these issues by:
(a) introducing substantial train/test distribution shifts to assess model robustness under non-i.i.d. conditions;
(b) ensuring some classes are under- or unrepresented in the training or test sets to reveal generalization limits for poorly sampled categories; and
(c) including unknown or novel classes only in the test set to enable direct evaluation of OOD and novelty detection.

These design choices directly affect model performance and calibration. While models may perform well on majority classes, accuracy typically declines for minority, shifted, or previously unseen categories. Moderate validation accuracy can obscure this generalization gap, underscoring the need for robust calibration and adaptation strategies.

By incorporating distribution shift and class imbalance, our protocol closely replicates the operational challenges of astronomical survey classification. The observed reduction in test accuracy highlights these real-world complexities and emphasizes the need for rigorous evaluation and adaptive methods in practical applications.

\begin{figure}[H]  
    \centering
    \includegraphics[width=0.95\linewidth]{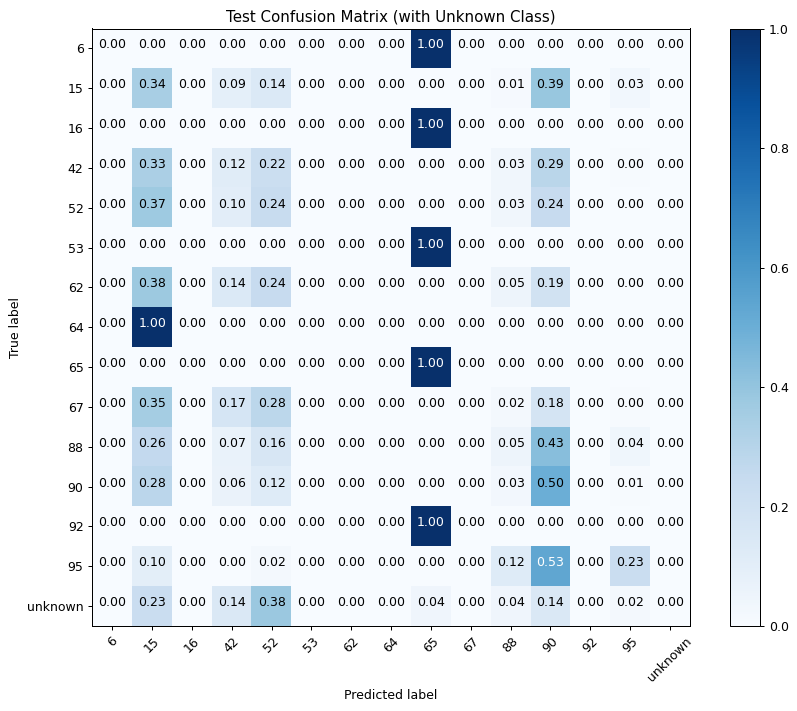}
    \caption{Test phase confusion matrix for PLAsTiCC data including unknown class detection. The matrix shows normalized classification results across known classes and the 'unknown' category.}
    \label{fig:test_confusion_matrix}
\end{figure}

\subsubsection{Recurrent Neural Networks (RNNs)}

The application of RNNs for the classification of astronomical transients and variables in the PLAsTiCC dataset results in substantially reduced accuracy across training, validation, and especially test phases. This is primarily due to the significant domain shift and class imbalance described above. Specifically, the training accuracy reaches 70.58\%, while the test accuracy drops markedly to 30.40\%, with validation accuracy falling in between. As illustrated in Fig.~\ref{fig:test_confusion_matrix}, the confusion matrix and class-wise performance show that well-represented and well-trained classes achieve reasonable accuracy, whereas under-represented or under-trained classes perform very poorly and remain highly overconfident. The presence of unknown classes, which are not seen during training but appear in the test set, further degrades accuracy and highlights the need for dedicated OOD or anomaly detection methods.

Table~\ref{tab:calibration_metrics_rnn} demonstrates that the uncalibrated RNN is severely miscalibrated, exhibiting strong overconfidence and a high MCE, largely due to OOD and anomalous samples absent from the training set. TS provides some reduction in ECE and improvement in NLL and Brier score, but overall calibration remains inadequate. Single-layer GP calibration, regardless of the selected layer, does not produce meaningful reductions in miscalibration, with both ECE and MCE remaining high across all configurations.

In contrast, the proposed SAL-GP(ML) achieves a substantial reduction in ECE, clearly outperforming both TS and single-layer GP calibration. Importantly, this improvement is only realized through the joint, multi-layer kernel structure; individual single-layer GPs do not yield similar effects. Nevertheless, MCE remains elevated, which is a direct result of persistent misclassification for OOD classes and systematic overconfidence in under-represented categories. This observation suggests that further algorithmic developments, such as OOD sample filtering, are needed to fully resolve calibration challenges in this setting \cite{b79}.

Layer-wise ablation for the SAL-GP(ML) shows that the specific combination of layers moderately influences calibration outcomes: using all layers (G(1–5)), only early layers (G(1–3)), or only late layers (G(3–5)) each yields somewhat different ECE values, suggesting that selective layer inclusion and automated kernel learning could further enhance calibration performance.

The SAL-GP(HL) does not offer significant improvements over TS or single-layer GP calibration. The ECE, MCE, and other calibration metrics remain similarly high, which is consistent with previous findings for this kernel.

\begin{figure*}[t]
    \centering
    \hspace*{0.19\textwidth}
    \hspace*{0.19\textwidth}
    \includegraphics[width=0.19\textwidth]{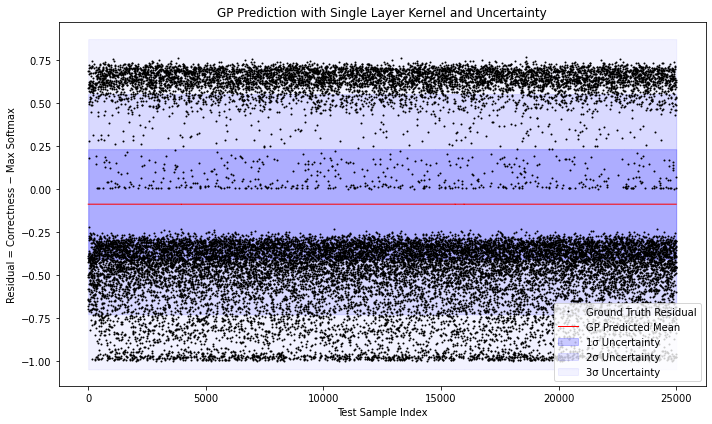}
    \includegraphics[width=0.19\textwidth]{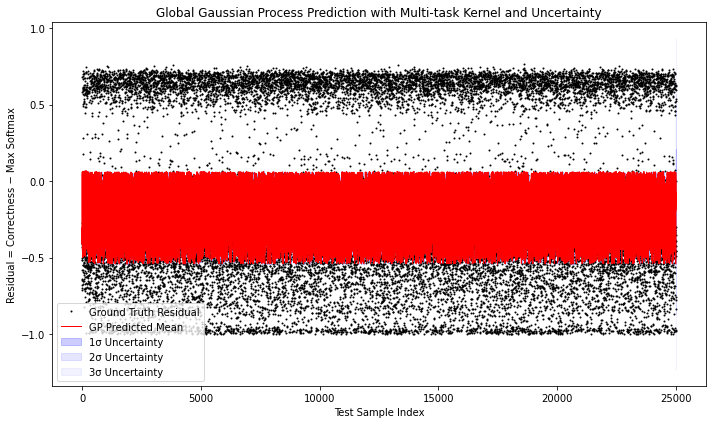}
    \includegraphics[width=0.19\textwidth]{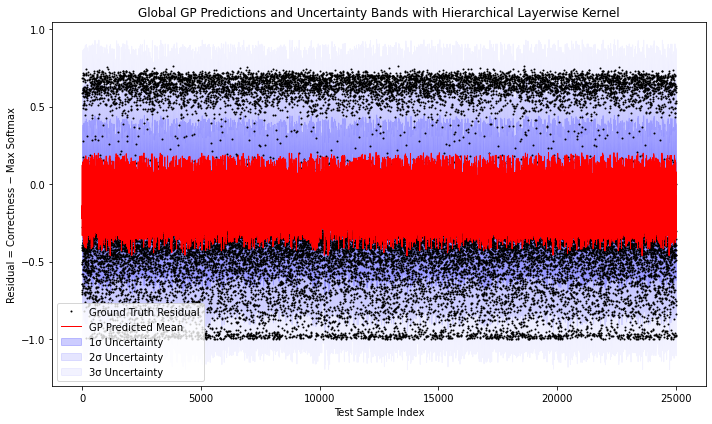} \\[2pt]
    \includegraphics[width=0.19\textwidth]{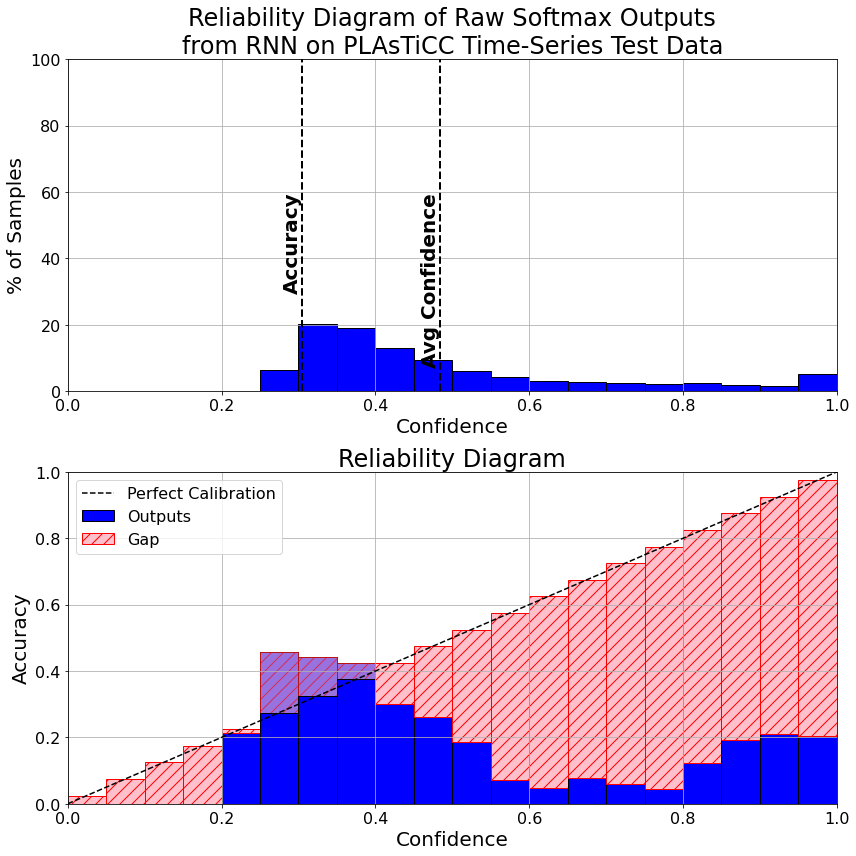}
    \includegraphics[width=0.19\textwidth]{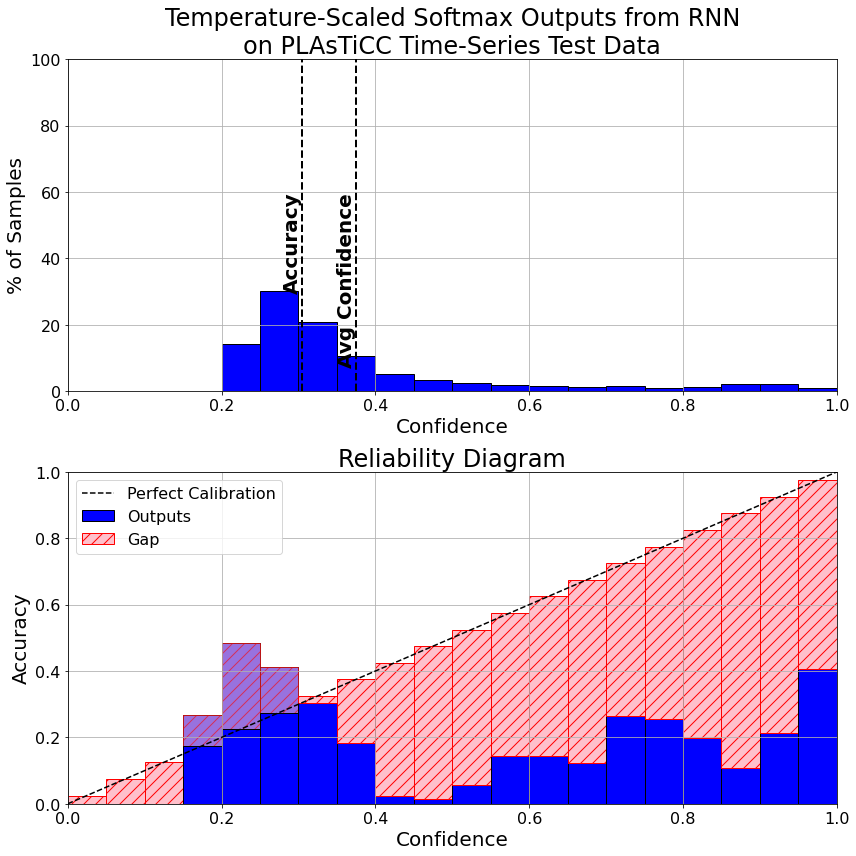}
    \includegraphics[width=0.19\textwidth]{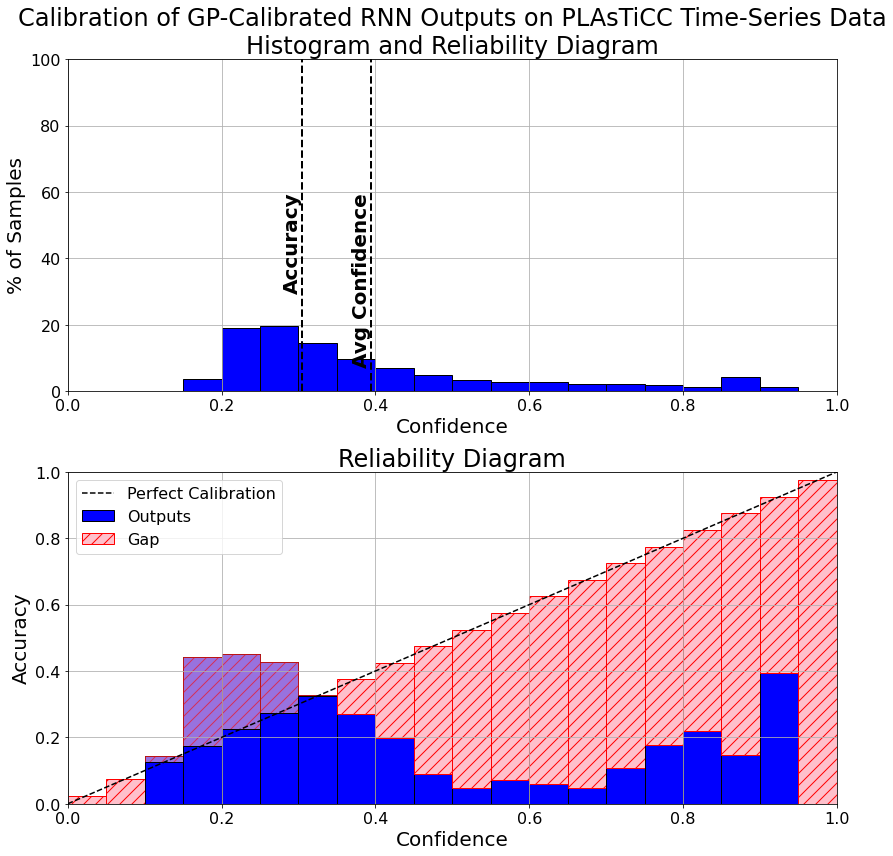}
    \includegraphics[width=0.19\textwidth]{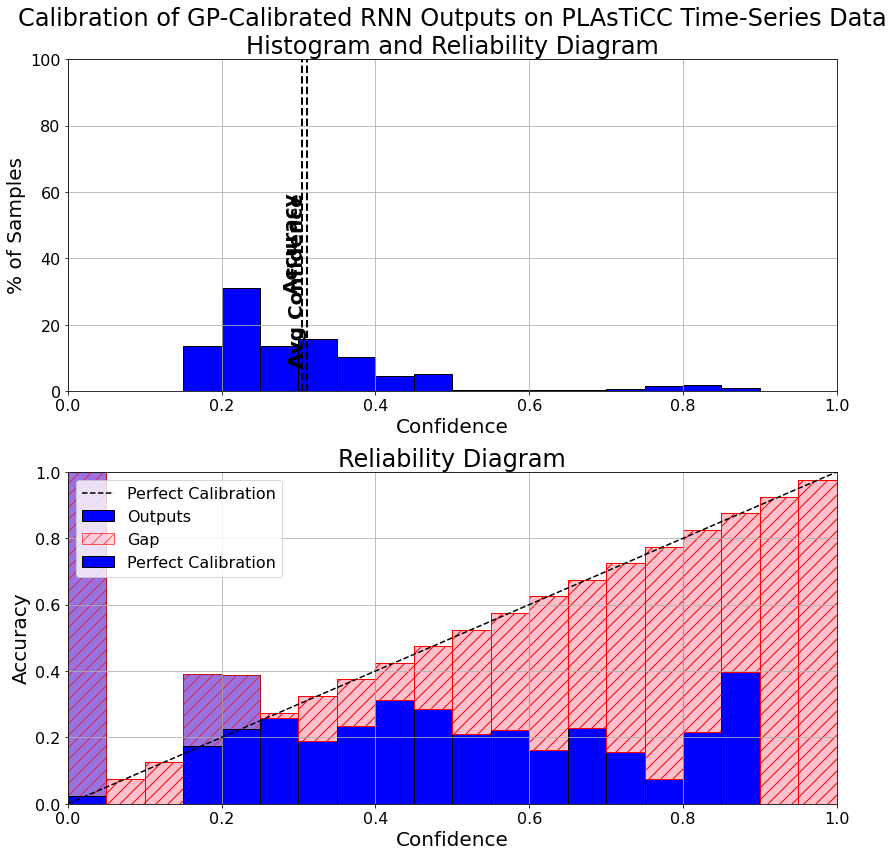}
    \includegraphics[width=0.19\textwidth]{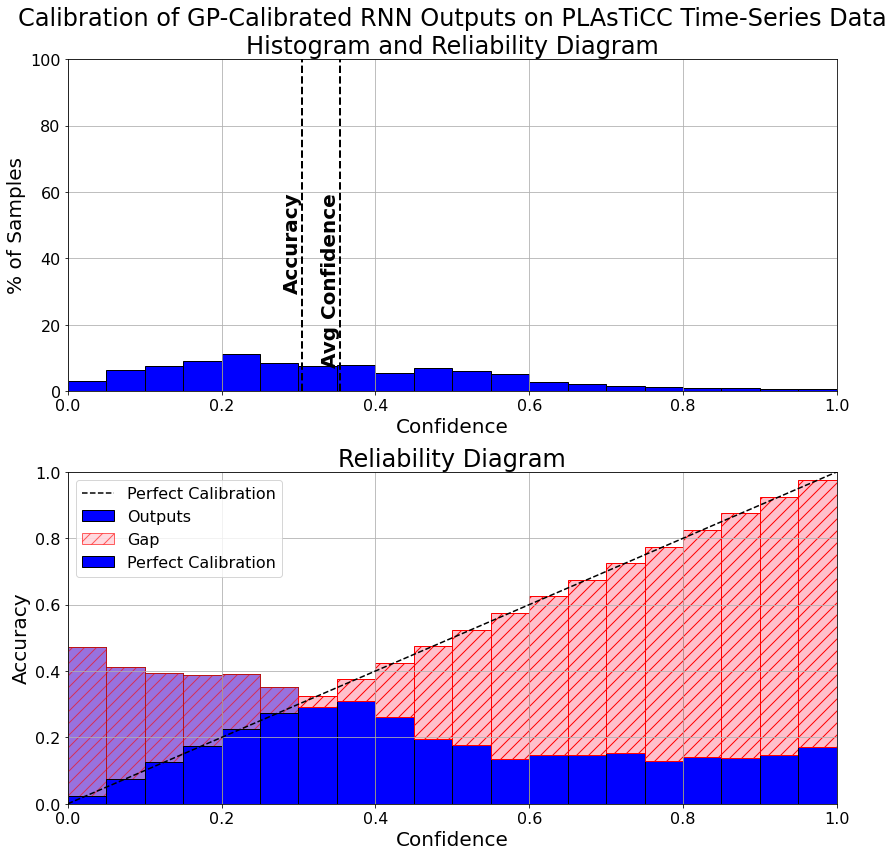} \\

    \begin{minipage}{0.19\textwidth}\centering Uncalibrated\end{minipage}
    \begin{minipage}{0.19\textwidth}\centering Temp. Scaled\end{minipage}
    \begin{minipage}{0.19\textwidth}\centering Single GP (Best Layer)\end{minipage}
    \begin{minipage}{0.19\textwidth}\centering SAL-GP (ML)\end{minipage}
    \begin{minipage}{0.19\textwidth}\centering SAL-GP (HL)\end{minipage}

    \caption{Calibration residual fit plots (top row) and reliability diagrams (bottom row) for each calibration approach on the RNN, evaluated on the PLAsTiCC time-series dataset. The top row displays only the residual fit plots for GP-based methods. For uncalibrated and temperature scaled columns, no residual fit is shown, since these methods do not provide predictive uncertainty estimates. Each column corresponds to a different calibration method: uncalibrated, temperature scaled, single-layer GP (best-performing layer), SAL-GP (multi-layer), and proposed SAL-GP (hierarchical layer kernel).}
    \label{fig:calibration_comparison_rnn_mstar}
\end{figure*}

The residual fit in the GP predictions and the corresponding diagrams confirm that only the SAL-GP(ML) and SAL-GP(HL) variants exhibit significant signals for overconfidence correction in the negative residual, as shown in Fig.~\ref{fig:calibration_comparison_rnn_mstar}. The inadequate quantification of predictive uncertainty observed for the SAL-GP(ML) variants is attributed to the choice of kernel hyperparameters.

When compared to results from image domain datasets, where the training data is more balanced and domain shift is less pronounced, the benefit of multi-layer kernel calibration is much more substantial for PLAsTiCC. This underscores the particular importance of SAL-GP(ML) calibration in time-domain astronomy, where class imbalance and train/test distribution shift are inherent. Future work on selective kernel design and OOD filtering will be essential for achieving robust calibration in practical astronomical survey pipelines \cite{b79}.

\begin{table*}[htbp]
\centering
\caption[
    Main Calibration and Accuracy Metrics: RNN with PLAsTiCC dataset
]{
    \textsc{Main Calibration and Accuracy Metrics: RNN}\\
    \textsc{with PLAsTiCC dataset}
}
\label{tab:calibration_metrics_rnn}  
\begin{threeparttable}
\begin{tabular}{l ccccc}
\hline\hline
\multicolumn{6}{c}{\textbf{Calibration Performance Metrics: RNN (PLAsTiCC Dataset)}} \\ \hline
\textbf{Method} & \textbf{ECE$\downarrow$} & \textbf{MCE$\downarrow$} & \textbf{NLL$\downarrow$} & \textbf{Brier$\downarrow$} & \textbf{Var.} \\ \hline
Uncalibrated (Baseline)        & 0.26768 & 0.77582 & 2.18086 & 0.90021 & - \\
Temperature Scaled (Baseline)  & 0.22461 & 0.77035 & 1.96450 & 0.85088 & - \\
Single GP (Layer 1)            & 0.26159 & 0.71018 & 2.19158 & 0.89532 & 0.09502 \\
Single GP (Layer 2)            & 0.25624 & 0.73339 & 2.19802 & 0.89444 & 0.10224 \\
Single GP (Layer 3)            & 0.26328 & 0.66357 & 2.22017 & 0.89503 & 0.10040 \\
Single GP (Layer 4)            & 0.25805 & 0.74805 & 2.19609 & 0.89470 & 0.09931 \\
Single GP (Layer 5)            & 0.27174 & 0.72540 & 2.20595 & 0.89604 & 0.10152 \\
SAL-GP (ML) G(1$\sim$5)   & 0.16533 & 0.98859 & 2.13469 & 0.85407 & 0.00001 \\
SAL-GP (ML) G(1$\sim$3)   & 0.15402 & 0.98832 & 2.11529 & 0.85976 & 0.11702 \\
SAL-GP (ML) G(3$\sim$5)   & 0.17804 & 0.79560 & 2.11565 & 0.86249 & 0.04352 \\
SAL-GP (ML) L1            & 0.17628 & 0.85311 & 2.11831 & 0.87247 & 0.12524 \\
SAL-GP (ML) L2            & 0.16961 & 0.84085 & 2.11814 & 0.86963 & 0.12372 \\
SAL-GP (ML) L3            & 0.16732 & 0.80338 & 2.11827 & 0.86804 & 0.12288 \\
SAL-GP (ML) L4            & 0.18074 & 0.79450 & 2.11861 & 0.86605 & 0.12281 \\
SAL-GP (ML) L5            & 0.17969 & 0.79558 & 2.11869 & 0.86557 & 0.12279 \\
SAL-GP (HL) G (1$\sim$5)  & 0.24537 & 0.80326 & 2.50549 & 0.90535 & 0.05843 \\
SAL-GP (HL) G (1$\sim$3)  & 0.25869 & 0.78893 & 2.22042 & 0.89561 & 0.05715 \\
SAL-GP (HL) G (3$\sim$5)  & 0.25702 & 0.79242 & 2.22636 & 0.89674 & 0.06096 \\
SAL-GP (HL) L1                     & 0.26533 & 0.70895 & 2.21802 & 0.89502 & 0.11417 \\
SAL-GP (HL) L2                     & 0.25851 & 0.72198 & 2.21370 & 0.89440 & 0.11402 \\
SAL-GP (HL) L3                     & 0.24871 & 0.73069 & 2.19964 & 0.89128 & 0.11394 \\
SAL-GP (HL) L4                     & 0.24759 & 0.70811 & 2.19674 & 0.88986 & 0.11405 \\
SAL-GP (HL) L5                     & 0.24997 & 0.70175 & 2.19814 & 0.89052 & 0.11413 \\ \hline
\textbf{Accuracy(Train, Validation, Test)} & \multicolumn{5}{c}{70.58\%, 58.60\%, 30.40\%} \\ \hline
\end{tabular}
\begin{tablenotes}
\footnotesize
\item Calibration metrics: ECE (Expected Calibration Error), MCE (Maximum Calibration Error), NLL (Negative Log-Likelihood), Brier score, and mean predictive variance (Var.).
\item Accuracy is reported for the original neural network and is unchanged across calibration methods.
\item For "Single GP," each row corresponds to a GP trained using only the $l$-th layer's latent representation plus softmax as input.
\item "SAL-GP (ML)" uses all layers as input with the structured multi-layer index kernel.
\item "SAL-GP (HL)" uses all layers as input with the hierarchical layer kernel.
\item Optimized T = 1.5315
\end{tablenotes}
\end{threeparttable}
\end{table*}

\section{Conclusion}

Our comprehensive evaluation across multiple datasets and neural network architectures demonstrates the effectiveness of the SAL-GP calibration framework, particularly in scenarios prone to miscalibration and overfitting.

For well-calibrated models, such as the standard ConvNet, all calibration methods—including SAL-GP—yield only modest improvements, reflecting limited potential for further gains. However, SAL-GP(ML) still provides robust, architecture-insensitive calibration, showing consistently superior performance across datasets and architectures. This stability directly addresses the unpredictability and sensitivity inherent in single-layer GP calibration, where optimal results often depend on ad hoc layer selection.

In cases where the baseline network exhibits significant miscalibration and overconfidence, as observed in AConvNet with the MSTAR dataset and in RNNs with the PLAsTiCC dataset, SAL-GP(ML) achieves substantial reductions in ECE, outperforming both single-layer GP and TS. These improvements are possible only through the joint, multi-layer kernel design, as no single-layer GP configuration yields comparable benefits. The SAL-GP(ML) consistently enhances calibration stability as well, while the SAL-GP(HL) provides less consistent gains and can underperform relative to simpler methods.

Ablation studies indicate that while average pooling can offer moderate additional benefits, particularly in severely overfitted architectures, the principal advantage of the SAL-GP framework is its resilience to both layer and pooling choices. This practical robustness is essential for real-world deployment, where the optimal calibration layer cannot be identified in advance.

For the PLAsTiCC dataset, which is marked by substantial domain shift, class imbalance, and the presence of OOD classes, the benefits of SAL-GP(ML) calibration are particularly pronounced. SAL-GP(ML) significantly reduces ECE, but the persistently elevated maximum calibration error (MCE) highlights ongoing challenges posed by OOD and underrepresented class samples, underscoring the need for further advances in hybrid calibration and OOD detection strategies.

Overall, the SAL-GP(ML) framework provides a principled and robust approach to post-hoc neural network calibration, delivering consistent and interpretable uncertainty quantification across diverse architectures and datasets. The process of aggregating information from each layer’s latent representation is analogous to a cognitive reasoning chain that integrate sequential reasoning and inference stages, Each layer contributes to distinct logical flow, and identifying which layers require calibration can provide insights into misalignments between training and test distributions. For this study, the additive variant of the multi-layer kernel was selected for its scalability and computational efficiency. However, more advanced calibration and deeper analysis of neural network decision-making could benefit from modeling intra- and inter-layer relations beyond additive kernels, potentially incorporating multiplicative forms for increased expressiveness. The benefits of SAL-GP are most evident in cases of severe miscalibration, architectural variability, and domain shift. Continued development of kernel structures and integration with OOD detection remain important directions for advancing calibration methods and ensuring robust deployment in challenging real-world scenarios.


%

\newpage

\comment{
\section*{\textbf{Supplementary Materials}}
\renewcommand{\thesection}{S\arabic{section}} 
\setcounter{section}{0} 
}

\section*{Acknowledgment}
The authors thank the Pohang University of Science and Technology for its generous support.

\ifCLASSOPTIONcaptionsoff
  \newpage
\fi

\end{document}